\documentclass{article}

\newif\ifdevscianonymise
\devscianonymisefalse

\newif\ifshowkorean
\showkoreanfalse  % Show English

\ifshowkorean
{

}
\else
\fi

\usepackage[utf8]{inputenc} % allow utf-8 input
\usepackage[T1]{fontenc}    % use 8-bit T1 fonts
\usepackage{hyperref}       % hyperlinks
\usepackage{url}            % simple URL typesetting
\usepackage{booktabs}       % professional-quality tables
\usepackage{amsfonts}       % blackboard math symbols
\usepackage{nicefrac}       % compact symbols for 1/2, etc.
\usepackage{microtype}      % microtypography
\usepackage{xcolor}         % colors
\usepackage{mathtools}      % package for \mathclap = more space for underbrace text

\usepackage[pdftex]{graphicx}

\ifshowkorean
    \usepackage[nonfrench,finemath,strictcharcheck]{kotex}
    \usepackage{dhucs-nanumfont}
    \newcommand{\devscismall}{\normalsize}  % 원본 크기
    \newcommand{\devsciscriptsize}{\small}  % 한국어 문서용 글자 작게
\else
    \newcommand{\devscismall}{\normalsize}  % 원본 크기
    \newcommand{\devsciscriptsize}{\small}  % 한국어 문서용 글자 작게
\fi

\usepackage{amsmath}  % Math
\usepackage{amssymb}  % Therefore symbol
\usepackage{subcaption}  % For subfigure
\usepackage{makecell}
\DeclareMathOperator*{\argmin}{\arg\min}

\usepackage[normalem]{ulem}
\usepackage{soul}
\definecolor{revisedcolor}{RGB}{200,220,255}  % Light blue
\sethlcolor{revisedcolor}
\newcommand{\rev}[1]{#1}  % Highlight OFF (no change)

\usepackage[preprint,nonatbib,mylinenumbers]{style_devsci}
\makeatletter
\renewcommand{\uline}[1]{#1}
\makeatother
\usepackage[round,longnamesfirst,sort]{natbib}
\newcommand\blfootnote[1]{%
    \bgroup
    \renewcommand\thefootnote{\fnsymbol{footnote}}%
    \renewcommand\thempfootnote{\fnsymbol{mpfootnote}}%
    \footnotetext[0]{#1}%
    \egroup
}

\title{Emergence of Goal-Directed Behaviors \\ via Active Inference with Self-Prior}

\ifdevscianonymise
    \author{
        Anonymous Author(s) \\
        Anonymous Affiliation(s)
    }
\else
    \author{%
  Dongmin~Kim \quad Hoshinori~Kanazawa$^*$ \quad Yasuo~Kuniyoshi$^*$ \\
  Graduate School of Information Science and Technology \\
  The University of Tokyo \\
  Tokyo, Japan \\
  \texttt{\{d-kim,kanazawa,kuniyosh\}@isi.imi.i.u-tokyo.ac.jp} \\
  \And
  Naoto~Yoshida \\
  Graduate School of Informatics \\
  Kyoto University \\
  Kyoto, Japan \\
  \texttt{yoshida.naoto.8x@kyoto-u.ac.jp}
}
\fi

\begin{document}

\maketitle

%TC:endignore

%TC:ignore
\begin{abstract}

    \ifshowkorean
      {\devsciscriptsize 아기는 외부에서 보상의 기준이 제공되지 않더라도, 감각 자극을 향해 리칭하는 등의 목표 지향적 행동을 종종 보인다. 이러한 내발적으로 동기 부여된 행동은, 초기 발달 단계에 있어서 신체와 환경을 자발적으로 탐색하고 학습하는 데 도움을 준다. 계산론적 모델링은 이러한 행동의 기전에 대한 통찰을 줄 수 있지만, 내발적 동기를 사용하는 많은 연구들은 탐색이 외부 보상을 얼마나 획득하는지에 대한 조사가 주를 이룬다. 본 논문에서 우리는 자신의 멀티모달 감각 경험에 대한 새로운 밀도 모델인 "self-prior"를 제안하여, 자발적으로 목표 지향적 행동을 유도할 수 있는지 조사한다. 자유 에너지 원리에 기반한 능동적 추론 프레임워크에 통합된 self-prior는, 지금까지 경험했던 평균적인 감각과 실제 감각 사이의 불일치를 최소화하는 순수한 내발적 과정으로부터 행동의 기준을 생성한다. 이는 환경과의 지속적인 상호작용에 의한 body schema의 획득 및 활용과도 유사하다. 우리는 이 접근법을 시뮬레이션 환경에서 시험하여(examine), 에이전트가 촉각 자극을 향해 손을 자발적으로 리칭함을 확인한다(confirm). 우리의 연구는 발달 초기 단계에서의 의도적인 행동의 자발적인 창발을 설명하는, 에이전트 스스로의 감각 경험으로부터 형성된 내발적으로 동기 부여된 행동을 구현한다.
      }
    \else
      %TC:endignore
      Infants often exhibit goal-directed behaviors, such as reaching for a sensory stimulus, even when no external reward criterion is provided. These intrinsically motivated behaviors facilitate spontaneous exploration and learning of the body and environment during early developmental stages. Although computational modeling can offer insight into the mechanisms underlying such behaviors, many existing studies on intrinsic motivation focus primarily on how exploration contributes to acquiring external rewards. In this paper, we propose a novel density model for an agent’s own multimodal sensory experiences, called the “self-prior,” and investigate whether it can autonomously induce goal-directed behavior. Integrated within an active inference framework based on the free energy principle, the self-prior generates behavioral references purely from an intrinsic process that minimizes mismatches between average past sensory experiences and current observations. This mechanism is also analogous to the acquisition and utilization of a body schema through continuous interaction with the environment. We examine this approach in a simulated environment and confirm that the agent spontaneously reaches toward a tactile stimulus. Our study implements intrinsically motivated behavior shaped by the agent's own sensory experiences, demonstrating the spontaneous emergence of intentional behavior during early development.
      \ifdevscianonymise
      \else
            \blfootnote{$*$ Co-corresponding authors: Hoshinori Kanazawa, Yasuo Kuniyoshi}
\blfootnote{Code is available at \url{https://github.com/kim135797531/self-prior}.}
      \fi
      % This is the pre-peer reviewed version of the following article: [FULL CITE], which has been published in final form at [Link to final article using the DOI]. This article may be used for non-commercial purposes in accordance with Wiley Terms and Conditions for Use of Self-Archived Versions.
      % This is the peer reviewed version of the following article: [FULL CITE], which has been published in final form at [Link to final article using the DOI]. This article may be used for non-commercial purposes in accordance with Wiley Terms and Conditions for Use of Self-Archived Versions. This article may not be enhanced, enriched or otherwise transformed into a derivative work, without express permission from Wiley or by statutory rights under applicable legislation. Copyright notices must not be removed, obscured or modified. The article must be linked to Wiley’s version of record on Wiley Online Library and any embedding, framing or otherwise making available the article or pages thereof by third parties from platforms, services and websites other than Wiley Online Library must be prohibited.
      %TC:ignore
    \fi
    \end{abstract}
    
    \section*{Research Highlights}
    \ifshowkorean
    {\devsciscriptsize
    -- 자유 에너지 원리 틀에서 신체 스키마의 형성과 목표 지향적 행동을 통합하여, 발달 초기의 의도적 행동에 대한 계산 모델을 제안합니다.
    
    -- 외부 보상 없이 목표 지향적 행동의 출현을 주도하는 내부 밀도 모델인 self-prior를 소개합니다.
    
    -- 관찰된 다감각 입력과 경험적으로 획득한 self-prior 간의 불일치를 최소화하여 스티커에 대한 자발적인 손 뻗기를 보여줍니다.
    }
    \else
    %TC:endignore
    -- Suggests a computational model for early intentional behavior, integrating body-schema formation and goal-directed actions under the free energy principle.
    
    -- Introduces a self-prior as an internal density model that drives the emergence of goal-directed behaviors without external rewards.
    
    -- Demonstrates spontaneous reaching for a sticker by minimizing mismatches between observed multimodal sensory inputs and the empirically acquired self-prior.
    %TC:ignore
    \fi
    
    \ifdevscianonymise
        \section*{Keywords}
        %TC:endignore
        Intrinsic Motivation, Free Energy Principle, Active Inference, Emergent Intentionality, Reaching, Body Schema
        %TC:ignore
    \else
        
    \fi

%TC:endignore

\newpage

%TC:ignore
\section{Introduction}
\ifshowkorean
{\devscismall
유아는 감각 자극을 향해 손을 뻗거나 주변 환경을 적극적으로 탐색하는 등 자발적인 목표 지향적 행동을 보이며, 이는 생존에 필수적인 외부 보상이 없는 상황에서도 발생한다. 이러한 행동은 내부적 만족을 위해 높은 동기를 가지고 추구되며, 이를 내발적으로 동기 부여된 행동이라고 한다~\citep{czikszentmihalyi_flow_1990, ryan_intrinsic_2000}. 이러한 행동은 유아가 외부의 명시적인 보상이나 처벌 없이도 비지도(unsupervised) 방식으로 학습을 촉진하는데, 이러한 비지도 학습은 초기 발달 단계에서 여러 가지 이점을 제공하는 것으로 알려져 있다~\citep{white_motivation_1959, gopnik_philosophical_2009, zaadnoordijk_lessons_2022, kanazawa_open-ended_2023}. 예를 들면, 유아의 자발적인 움직임에 의해 발생한 self-touch와 같은 감각 경험은, 그 자체만으로도 body representation의 획득과 같이 자신의 신체에 대한 학습을 촉진할 뿐만 아니라~\citep{hoffmann_development_2017}, early sense of self의 형성에도 기여하는 것으로 알려져 있다~\citep{rochat_self-perception_1998}.
}
\else
%TC:endignore
Infants spontaneously exhibit goal-directed behaviors, such as reaching for a sensory stimulus or actively exploring their surroundings, even in the absence of external rewards essential for survival. These behaviors are highly motivated by internal satisfaction and are referred to as intrinsically motivated behaviors \citep{czikszentmihalyi_flow_1990, ryan_intrinsic_2000}. Such behaviors facilitate unsupervised learning without explicit external rewards or punishments, which is known to offer various advantages in early developmental stages \citep{white_motivation_1959, gopnik_philosophical_2009, zaadnoordijk_lessons_2022, kanazawa_open-ended_2023}. For example, sensory experiences arising from an infant’s spontaneous movements, such as self-touch, not only promote learning about one’s own body (\emph{e.g.}, acquiring body representations) \citep{hoffmann_development_2017}, but also contribute to the formation of an early sense of self \citep{rochat_self-perception_1998}.
%TC:ignore
\fi

\ifshowkorean
{\devscismall
이러한 내발적으로 동기 부여된 행동 발달의 기전을 조사하기 위한 방법으로는 전통적인 관찰 연구나, 신경과학적 방법이 도입되고 있으며~\citep{di_domenico_emerging_2017}, 특히 행동 발달을 정량적으로 해석하고 예측할 수 있게 하는 계산론적 모델링에 의해서도 조사되고 있다~\citep{oudeyer_what_2009,shultz_computational_2013}. 이러한 컴퓨터 시뮬레이션을 사용한 실험은 특정한 변수를 자유롭게 조작·제어하여 다양한 시나리오를 조사할 수 있으므로, 전통적 행동 실험만으로는 확인하기 어려운 잠재적 효과나 상호작용을 발견하는 데 유리하다.
}
\else
%TC:endignore
Traditional observational studies and neuroscience methodologies have been employed to investigate the mechanisms underlying intrinsically motivated behavioral development \citep{di_domenico_emerging_2017}. In particular, computational modeling has also been utilized to quantitatively interpret and predict behavioral development \citep{oudeyer_what_2009,shultz_computational_2013}. Experiments using computer simulations allow researchers to freely manipulate and control specific variables to investigate various scenarios, making them advantageous for uncovering latent effects and interactions that may be difficult to observe through conventional behavioral studies alone.
%TC:ignore
\fi

\ifshowkorean
{\devscismall
계산론적 모델링이 유아의 행동 발달을 연구하는데 유용한 방법임에도 불구하고, 현대 로봇공학 및 머신 러닝 분야에서 제안된 많은 모델들은 내발적 동기를 주로 ‘희소한 외부 보상 환경에서 보상 획득 효율을 높이는 탐색 기제’로 다루는 경향이 크다~\citep{aubret_information-theoretic_2023}. 실제로 이러한 접근법들을 활용하면, 강화학습 에이전트가 비디오 게임 점수나 명시적인 과업 성능을 빠르게 높이는 데 도움을 줄 수 있다. 그러나 명시적인 외부 보상 목표가 설정되지 않았을 때, 즉 성능 측정 기준이나 정보 이득이 주어지지 않는 상황에서 완전히 새로운 행동이 어떻게 창발하는지에 대해서는 충분히 다루지 않는다는 한계가 존재한다.
}
\else
%TC:endignore
Although computational modeling is a useful approach for studying infant behavioral development, many models proposed in modern robotics and machine learning tend to treat intrinsic motivation primarily as an exploratory mechanism for improving reward acquisition efficiency in environments with sparse external rewards \citep{aubret_information-theoretic_2023}. Indeed, such approaches can enhance reinforcement learning agents’ ability to improve explicit task performance, such as maximizing video game scores. However, they do not sufficiently address how entirely new behaviors emerge when no explicit external reward objective is set, meaning that no performance criterion or information gain is provided.
%TC:ignore
\fi

\begin{figure}[t]
    \centering
    % \fbox{\rule[-.5cm]{0cm}{4cm} \rule[-.5cm]{4cm}{0cm}}
    \includegraphics[width=1.0\linewidth]{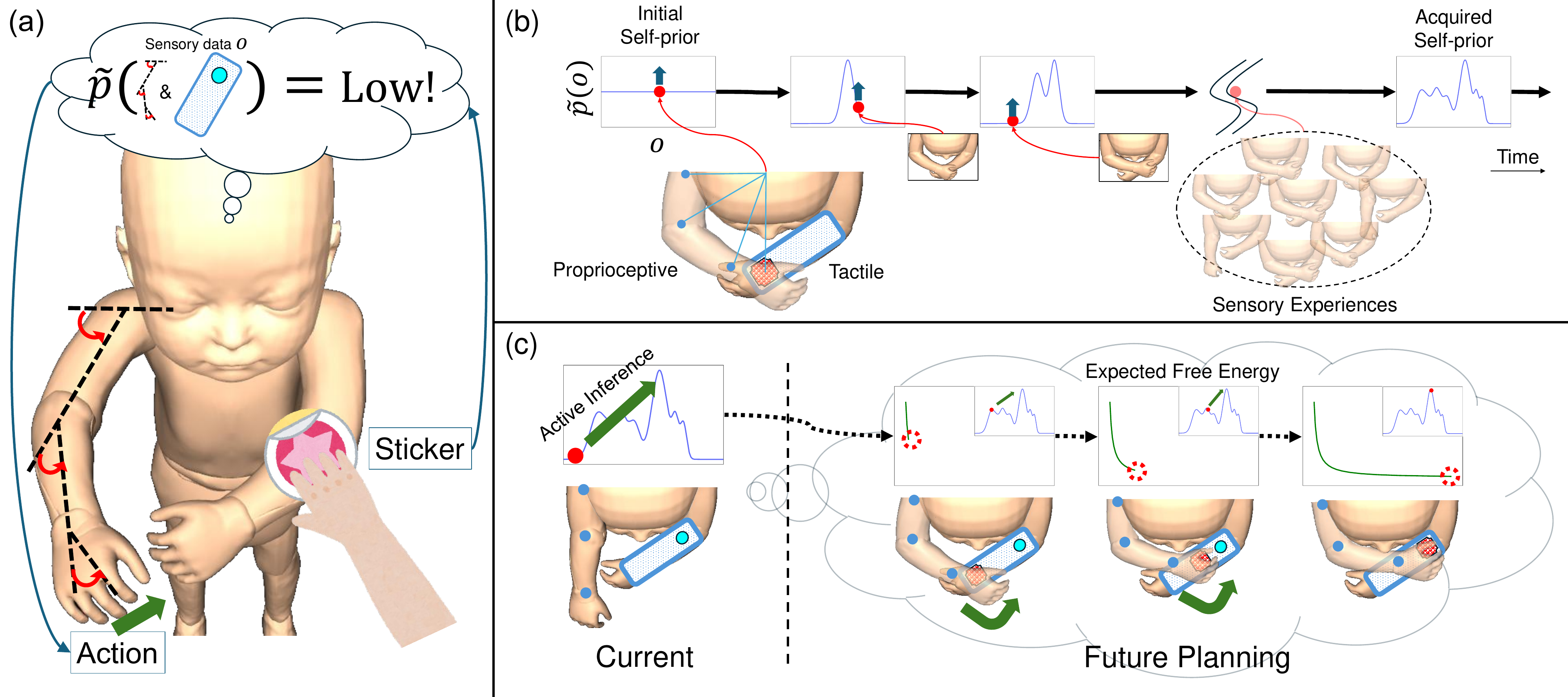}
    \caption{
      \ifshowkorean
      {\devsciscriptsize Self-prior와 능동적 추론에 의한 리칭 행위의 창발. (a) 시뮬레이션 된 에이전트의 왼팔에 스티커가 배치되면, 스티커가 없었던 지금까지의 경험과 불일치를 느끼고, 불일치를 최소화하기 위해 자신의 오른손을 스티커를 향해 뻗는다. (b) 경험을 통한 self-prior의 발달: 감각 경험을 수집하며 점차 감각 패턴의 확률 분포가 발달한다. (c) 에이전트가 감각 입력과 학습된 self-prior를 정렬하는 것으로 기대 자유 에너지를 최소화하기 위해 어떻게 미래 행동을 계획하는지 보이는 능동적 추론 프로세스. 그 결과 에이전트는 sticker를 향한 리칭 행위를 수행한다. 이해를 돕기 위해 아기의 전신 모델 그래픽 일러스트를 사용하였으며, 실제 실험은 의사 3d 환경에서 실행되었다.}
      \else
  %TC:endignore
      Emergence of reaching behavior via the self-prior and active inference. (a) When a sticker is placed on the left arm of the simulated agent, it detects a mismatch with its prior experience of not having a sticker, and reaches toward the sticker with its right hand to minimize the discrepancy. (b) Development of the self-prior through experience: as sensory experiences are collected, the probability distribution over sensory patterns gradually develops. (c) The active inference process in which the agent plans future actions to minimize expected free energy by aligning sensory inputs with the learned self-prior. As a result, the agent performs a reaching action toward the sticker. A full-body infant illustration is used for clarity, the actual experiment was conducted in a pseudo-3D environment.
  %TC:ignore
      \fi
    }
    \label{fig:fig1}
  \end{figure}

\ifshowkorean
{\devscismall
이를 보완하고자, 우리는 기대 자유 에너지~\citep{friston_free-energy_2010,friston_active_2016}에 기반한 새로운 내발적 동기의 계산 모델을 제안한다 (Figure~\ref{fig:fig1}). 우리가 도입한 \emph{self-prior}라고 명명한 내부 밀도 모델은, 에이전트가 들어오는 멀티모달 감각 신호에 대한 통계적 표상을 학습하도록 한다. 학습된 확률 모델과 실제 관찰 간의 불일치가 감지되면, 명시적인 보상 기준 없이도 목표 지향적인 행위가 창발한다.

% rev: 차별점 재정리
\uline{우리의 접근법은 기존의 발달 행동에 대한 계산 모델들에서 통합되지 않았던 여러 핵심 특징들을 결합한다: (\textit{i}) 행동의 기준점이 외부에서 미리 정의되지 않고 에이전트 자신의 경험으로부터 자율적으로 형성되며; (\textit{ii}) 리칭과 같은 특정한 목표 지향적 행동이 단순히 보상 탐색을 위한 보조 메커니즘이 아니라 독립적으로 창발하고; (\textit{iii}) 정보 이득이 없는 상황에서도 익숙한 상태와의 불일치가 있는 한 행동이 지속되며; (\textit{iv}) 이 메커니즘이 능동적 추론 프레임워크 내에서 기존의 외발적 및 내발적 동기와 이론적으로 통합 가능하다.}
우리는 self-prior 매커니즘에 의해 팔에 붙은 스티커와 같은 자극을 인지하고 이를 자발적으로 만지려는 행동이 창발하는 시뮬레이션된 에이전트를 구현한다. 논문의 후반부에서는 제안한 메커니즘이 발달과학에 기여할 수 있는 방식에 대해 논의하는 것으로, 유아의 초기 의도적인 행동 발달에 관한 계산론적 연구로서 자리매김하는 방식을 설명한다.
}
\else
%TC:endignore
To address this limitation, we propose a novel computational model of intrinsic motivation based on the expected free energy \citep{friston_free-energy_2010,friston_active_2016} (Figure~\ref{fig:fig1}).
We introduce an internal density model, termed the \emph{self-prior,} which enables an agent to learn a statistical representation of incoming multimodal sensory signals.
When a mismatch between the learned probabilistic model and actual observations is detected, goal-directed behavior emerges without any explicit reward criterion.

\rev{Our approach combines several key features that, to our knowledge, have not been integrated in previous computational models of developmental behavior: (\textit{i}) the behavioral setpoint is autonomously formed from the agent's own experience rather than externally predefined; (\textit{ii}) specific goal-directed behaviors such as reaching emerge independently, rather than serving merely as auxiliary mechanisms for reward exploration; (\textit{iii}) behaviors persist even without information gain, as long as there is a mismatch with familiar states; and (\textit{iv}) the mechanism is theoretically integrable with existing extrinsic and intrinsic motivations within the active inference framework.}
We implement a simulated agent that perceives a stimulus (such as a sticker) on its arm and spontaneously exhibits reaching behavior toward it through the self-prior mechanism.
In the latter part of this paper, we discuss how the proposed mechanism may contribute to developmental science, positioning this study as a computational investigation into the early development of intentional behaviors in infants.
%TC:ignore
\fi

\subsection{Free Energy Principle}

\ifshowkorean
    {\devscismall
\textit{자유 에너지 원리}는 에이전트가 감각 입력으로 인해 발생하는 "놀라움"을 최소화함으로써 지각하고 행동하는 법을 학습할 수 있다는 개념을 제시한다~\citep{friston_free-energy_2010}. 이 놀라움이란 에이전트가 예측하지 못한 감각 관찰을 경험할 때 발생하는 예측 오류를 의미하며, 이를 줄이는 것이 에이전트의 주요 목표가 된다.

각 시간 단계 $t$에서, 에이전트는 감각 기관을 통해 세계로부터 정보를 받아들인다. 우리는 이 감각 입력을 $o_t$라고 표기하며, 이를 관찰이라고 부른다. 에이전트가 직접 관찰할 수 있는 것은 $o_t$뿐이지만, 실제로는 이 관찰을 생성하는 세계의 근본적인 요인이 존재할 수 있다. 이러한 근본적인 요인을 숨겨진 상태라 하며, 시간 $t$에서의 숨겨진 상태를 $s_t$로 표기한다. 예를 들어, 적절한 학습이 이루어진 모델을 가진 에이전트는 팔에 가해지는 촉각 패턴(관찰 $o_t$)을 바탕으로, 자신의 팔에 스티커가 붙어 있다는 숨겨진 상태 $s_t$를 추론할 수 있다.
}
\else
%TC:endignore
The \textit{free energy principle} posits that an agent can learn to perceive and act by minimizing the "surprise" induced by sensory inputs \citep{friston_free-energy_2010}.
This surprise refers to the prediction error that arises when the agent encounters unexpected sensory observations, and reducing this surprise becomes the agent's primary objective.

At each time step $t$, the agent receives information from the world through its sensory organs.
We denote this sensory input by $o_t$, referring to it as the observation.
Although $o_t$ is all the agent can directly observe, there may exist underlying factors in the world that generate these observations.
Such factors are called hidden states, denoted by $s_t$ at time $t$.
For example, given a properly learned model, an agent can infer from a tactile pattern on its arm (observation $o_t$) that there is a sticker attached to its arm (hidden state $s_t$).
%TC:ignore
\fi

\ifshowkorean
{\devscismall
어떤 에이전트가 생존을 위해 특정한 생물학적 기준을 안정적으로 만족해야 한다고 가정하자. 이를 위해서는 그 에이전트가 환경을 정확하게 이해해야 하며, 이는 $p(o_t)$를 추론하는 것으로 귀결된다. 자유 에너지 원리에 따르면, 에이전트는 숨겨진 상태에 대한 자체적인 확률 모델을 유지하며, 이를 통해 $p(o_t)$를 근사하여 세계에 대한 추론을 수행할 수 있다. 실제 계산에서는 종종 $\log p(o_t)$를 최대화하는데, 이를 모델 증거 최대화라고 한다. 반대 부호의 $-\log p(o_t)$는 놀라움이라고 하며, 이를 최소화하는 것은 모델 증거를 최대화하는 것과 동치이다.
}
\else
%TC:endignore
Consider an agent that must satisfy certain biological criteria for survival in a stable manner. To do so, the agent must accurately understand its environment, which amounts to inferring $p(o_t)$. According to the free energy principle, an agent maintains its own probabilistic model of these hidden states because doing so enables it to approximate $p(o_t)$ and thus make inferences about the world. In practice, we often maximize the log of $p(o_t)$ (i.e., $\log p(o_t)$) for computational convenience, which is referred to as maximizing the model evidence. Minimizing the negative term, $-\log p(o_t)$, known as surprise, is therefore equivalent to maximizing this model evidence.
%TC:ignore
\fi

\ifshowkorean
{\devscismall
현실적으로, $p(o_t)$를 계산하려면 숨겨진 상태에 대해 주변화해야 하는데 (즉, $p(o_t, s_t)$), 이는 일반적으로 해석적으로 풀기 어렵다. 변분 자유 에너지 원리는 베이즈 추론의 변분 근사를 도입하여 이 문제를 해결하며, 이를 통해 에이전트는 놀라움의 변분 상한인 변분 자유 에너지 $\mathcal{F}$를 최소화함으로써 간접적으로 놀라움을 줄인다. 수식적으로, $\mathcal{F}$는 다음과 같이 정의된다 (자세한 근사 및 유도 과정은~\citep{parr_active_2022}를 참고하라):
}
\else
%TC:endignore
In reality, calculating $p(o_t)$ requires marginalization over hidden states (i.e., $p(o_t, s_t)$), which is often intractable. The variational free energy principle tackles this by introducing a variational approximation of Bayesian inference, wherein the agent indirectly reduces surprise by minimizing the variational free energy $\mathcal{F}$, an upper bound on $-\log p(o_t)$. Formally, $\mathcal{F}$ is defined as follows (see \citep{parr_active_2022} for details):
%TC:ignore
\fi

\begin{equation}
    \begin{aligned}
        -\log p(o_t) \le \mathcal{F} 
        &= \mathbb{E}_{q(s_t)}[\log q(s_t) - \log p(o_t, s_t)]\\
        &= \underbrace{D_{\mathrm{KL}}[q(s_t)\|p(s_t)]}_{\text{complexity}} 
        -
        \underbrace{\mathbb{E}_{q(s_t)}[\log p(o_t \mid s_t)]}_{\text{accuracy}} \label{equation_3}
    \end{aligned}
\end{equation}

\ifshowkorean
{\devscismall
여기서 $D_{\mathrm{KL}}(\cdot \| \cdot)$은 Kullback--Leibler 발산을 의미하며, $q(\cdot)$는 변분 근사된 분포를 의미한다. 즉, $q(s_t)$는 숨겨진 상태에 대한 사후 분포를 근사하는 변분 분포이다. 따라서, 시간 $t$에서 놀라움을 줄이려면, 에이전트는 사후 분포와 사전 분포가 가깝도록 유지하는 동시에(complexity 감소), 관찰의 가능도를 최대화하여야 한다(accuracy 향상).
}
\else
%TC:endignore
where $D_{\mathrm{KL}}(\cdot \| \cdot)$ denotes the Kullback--Leibler divergence, and $q(\cdot)$ represents the variationally approximated distribution. That is, $q(s_t)$ is the variational distribution that approximates the posterior distribution over hidden states. Therefore, to reduce surprise at time $t$, the agent must keep the posterior and prior distributions close (reducing complexity) while maximizing the likelihood of observations (improving accuracy).
%TC:ignore
\fi

\subsection{Active Inference and Expected Free Energy}
\ifshowkorean
{\devscismall
자유 에너지 원리에서는 현재의 감각 관찰을 기반으로 놀라움을 줄이기 위해 자신의 모델을 갱신하지만, 에이전트는 관찰이 자신의 모델과 더 잘 맞도록 직접 행동 $a_t$을 통해 관찰을 바꿀 수도 있다. 이것이 \textit{능동적 추론}의 본질이며, 구체적으로는 근사된 사후 분포 $q(s_t)$가 행동을 포함하도록 확장되어 $q(s_t, a_t) = q(a_t \mid s_t)\,q(s_t)$로 표현된다.

현재 및 미래의 최적 행동 $a_t$를 얻기 위해, 미래 시점의 자유 에너지에 대한 기대값을 고려하여 행동을 결정할 수 있도록 \textit{기대 자유 에너지}를 계산하는 방법이 제안되어 있다~\citep{friston_active_2016}. 즉, 에이전트는 여러 가능한 미래 상태를 예측한 후, 자유 에너지를 가장 효과적으로 최소화할 것으로 예상되는 경로를 선택한다.
}
\else
%TC:endignore
While the free energy principle describes how an agent updates its model to reduce surprise based on current sensory observations, the agent can also directly change its observations through actions $a_t$ to make them align better with its internal model. This is the essence of \textit{active inference}, and more concretely, the approximate posterior $q(s_t)$ is extended to include actions as $q(s_t, a_t) = q(a_t \mid s_t)\,q(s_t)$.

To obtain optimal actions $a_t$ for both present and future steps, a method has been proposed that computes \textit{expected free energy} by considering the expected free energy at future time points~\citep{friston_active_2016}. That is, the agent predicts multiple possible future states and selects the trajectory that is expected to minimize free energy most effectively.
%TC:ignore
\fi

\ifshowkorean
{\devscismall
그러나 미래의 관찰은 실제로 그 시점이 도래하기 전까지 얻을 수 없기 때문에, 에이전트는 미래 상태의 자유 에너지를 직접 계산할 수 없다. 따라서, 기대 자유 에너지는 에이전트가 선호하는 관찰을 포함하는 \textit{선호하는 사전 분포} $\tilde{p}(o_t)$를 가지고 있다고 가정하며, 이를 기준점(setpoint)으로 삼는다. 결과적으로, $p(o_t, s_t)$는 행동을 포함하고 선호하는 관찰로 편향되도록 확장된 \uline{$\tilde{p}(o_t, s_t, a_t) = p(a_t)\,p(s_t \mid o_t)\,\tilde{p}(o_t)$로 표현된다.} 따라서 기대 자유 에너지 $\mathcal{G}$는 다음과 같이 정의된다 (자세한 근사 및 유도 과정은~\citep{millidge_relationship_2020, mazzaglia_contrastive_2021}를 참고하라):

% 여기서 $q(s_t \mid o_t) \approx p(s_t \mid o_t)$를 가정하고, 에이전트가 행동에 대해 균일한 사전 분포 (uniform prior) $p(a_t)$를 갖는다고 가정하면,
}
\else
%TC:endignore
However, since future observations cannot be obtained before the corresponding time actually arrives, the agent cannot directly compute the free energy of future states.
Therefore, the expected free energy assumes that the agent possesses a \textit{preferred prior} $\tilde{p}(o_t)$ over desired observations, which acts as a setpoint.
As a result, $p(o_t, s_t)$ is extended to include actions and is biased toward preferred observations, \rev{expressed as $\tilde{p}(o_t, s_t, a_t) = p(a_t)\,p(s_t \mid o_t)\,\tilde{p}(o_t)$.}
Thus, the expected free energy $\mathcal{G}$ is defined as follows (see \citep{millidge_relationship_2020, mazzaglia_contrastive_2021} for details):
%TC:ignore
\fi

\begin{equation}
    \begin{aligned}
    \mathcal{G} &= \mathbb{E}_{q(o_t, s_t, a_t)}[\log q(s_t, a_t) - \log \tilde{p}(o_t, s_t, a_t)] \\
    &\approx
    -\underbrace{\mathbb{E}_{q(o_t)}[\log \tilde{p}(o_t)]}_{\text{extrinsic value}}
    -\underbrace{\mathbb{E}_{q(o_t)}[D_{\mathrm{KL}}[q(s_t \mid o_t) \| q(s_t)]]}_{\text{intrinsic value}}
    -\underbrace{\mathbb{E}_{q(s_t)}[\mathcal{H}(q(a_t \mid s_t))]}_{\text{action entropy}}
    \end{aligned}
    \label{eq:efe}
\end{equation}

\ifshowkorean
{\devscismall
여기서 $\mathcal{H}$는 정보 엔트로피를 의미한다. 즉, 미래 시점 $t$에서 놀라움을 줄이려면, (\textit{i}) 선호하는 관찰의 확률을 증가시키고, (\textit{ii}) 해당 관찰로부터 높은 정보 이득을 얻으며, (\textit{iii}) 다양한 행동을 유지해야 한다. 위와 같이 기대 자유 에너지는 pragmatic (외발적) 가치와 epistemic (내발적) 가치를 하나의 수식으로 통합하므로, 탐색-활용의 딜레마를 자연스럽게 해결할 수 있다.

% 즉, 외발적 동기 항은 생존과 관련된 외부 보상이나 목표를 달성하려는 동기와 유사한 반면, 내발적 동기 항은 새로운 정보를 탐색하거나 불확실성을 줄이려는 내적인 욕구와 유사하다. 두 동기는 동일한 정보 단위(nats)로 측정되므로, 하나의 체계 안에서 자동적으로 균형을 이루게 된다.
}
\else
%TC:endignore
where $\mathcal{H}$ denotes the information entropy. That is, to reduce surprise at a future time $t$, the agent must (\textit{i}) increase the probability of preferred observations, (\textit{ii}) obtain high information gain from those observations, and (\textit{iii}) maintain diverse actions.
As shown above, expected free energy integrates both pragmatic (extrinsic) and epistemic (intrinsic) value into a single expression, thereby naturally resolving the exploration--exploitation dilemma.

% That is, the extrinsic motivation term is similar to a drive for achieving goals or obtaining rewards necessary for survival, whereas the intrinsic motivation term corresponds to an internal desire to explore new information or reduce uncertainty. Since both are measured in the same information unit (nats), they are automatically balanced within a unified system.
%TC:ignore
\fi
%TC:endignore

%TC:ignore
\section{Self-Prior: Acquired Preference to Induce Goal-Directed Behavior}
\ifshowkorean
{\devscismall
기대 자유 에너지에 의해 움직이는 에이전트는 환경으로부터의 스칼라 보상을 정의해 줄 필요가 없다. 왜냐면 에이전트는 스스로가 선호하는 관찰에 대한 정보를 preferred prior 분포로서 갖고 있으며, 이것이 행동 계획의 기준점(setpoint)으로 작용하기 때문이다.
}
\else
%TC:endignore
Under the expected free energy, agents do not require scalar rewards to be explicitly defined by the environment.
This is because the agent is assumed to possess information about preferred observations as a preferred prior distribution, which serves as the setpoint for action planning.
%TC:ignore
\fi

\ifshowkorean
{\devscismall
그러나 일반적으로 preferred prior는 고정된 상태로 주어지기 때문에, 외부 보상 개념이 사라졌다고 하더라도, 이 에이전트가 목표 지향적인 행위를 \textit{스스로} 만들어 냈다고는 할 수 없다. 따라서, 본 연구에서는 에이전트가 행동 계획의 기준점을 스스로 결정할 수 있도록, 우리가 \emph{self-prior}라고 부를(refer to) 밀도 모델을 추가적으로 학습하도록 설정했다.

% rev: 설명 추가 보충
\uline{Self-prior는 "자신이 경험한 감각의 관측 빈도에 대한 확률밀도"를 나타내어, 에이전트가 익숙한 상태를 유지하거나 재경험하려는 행동 경향을 유도하는 가변적 preferred prior로 정의된다. 이는 고정된 preferred prior와 달리, 동일한 구조를 가진 모델이라도 다른 경험에 따라 다른 prior가 발달하여, 다른 행위가 창발할 수 있음을 의미한다.}
Self-prior의 작동 과정을 직관적으로 이해하기 위해, 본 논문에서는 자신의 몸에 붙은 스티커로의 자발적 리칭 및 제거를 예시로 보인다. 이는 \citep{bigelow_development_1986}에서 blind인 아기의 몸에 소리가 나지 않는 장난감을 배치하여 reaching 행동을 조사한 연구의 작업 설정과 비슷하다.
% rev: bigelow의 self-prior에서의 해석 추가 보충
\uline{여기서 우리는 아기 에이전트가 자신의 몸에 붙은 스티커를 떼는 이유를, "스티커가 없는 자신"이라는 prior와 "관측한 현재의 자신"이 불일치하기 때문이라고 해석한다. 그리고 그 prior는, 에이전트가 (스티커가 붙어 있지 않은) 일상적인 경험을 통해 스스로, 그리고 서서히 발달시켜 온 것이다. 이 self-prior에는 스티커를 제거하라는 구체적인 지시는 포함되어 있지 않으므로, 고정된 setpoint를 제공하여 reaching을 직접 명령하는 방식과는 구별된다.}
}
\else
%TC:endignore 
However, since the preferred prior is typically given in a fixed form, the removal of external reward signals does not imply that the agent has \textit{autonomously} generated its goal-directed behavior.
Therefore, in this study, we allow the agent to autonomously determine the setpoint for its action planning by additionally learning a density model, which we refer to as the \emph{self-prior}.

\rev{The self-prior represents "the probability density over the frequency of observed sensations that the agent has experienced," and is defined as a variable preferred prior that induces behavioral tendencies for the agent to maintain or re-experience familiar states.
Unlike fixed preferred priors, this means that models with the same structure can develop different priors depending on different experiences, leading to the emergence of different behaviors.}
To provide an intuitive understanding of how the self-prior operates, this paper presents the example of spontaneous reaching toward and removal of a sticker attached to one's own body.
This is similar to the experimental setup of \citep{bigelow_development_1986}, where a silent toy was placed on the body of a blind infant to examine reaching behavior.
\rev{Here, we interpret the reason why the infant agent removes a sticker attached to its body as being due to a mismatch between the prior of "oneself without a sticker" and "the currently observed self."
Moreover, this prior has been autonomously and gradually developed by the agent through everyday experiences (in which no sticker was attached).
Since this self-prior does not contain specific instructions to remove the sticker, it is distinguished from approaches that directly command reaching by providing fixed setpoints.}
%TC:ignore
\fi

\ifshowkorean
{\devscismall
능동적 추론 프레임워크에서의 self-prior 구현은, 기존의 기대 자유 에너지 공식의 preferred prior 항으로부터 self-prior 항을 분리해 내는 단순한 변경으로 도출되었다. 구체적으로는, 각 선호 관찰 $o$를 다음 두 구성 요소로 재정의함으로서 이루어진다: (\textit{i}) 생존 요구를 위해 반드시 유지되어야 하는 \emph{외발적} 관찰 $o^E$, (\textit{ii}) self-prior와 관련된 \emph{내발적} 관찰 $o^I$. 즉, $o^I$는 팔에 가해지는 촉각 패턴과 같이, 에이전트의 기본적인 기능을 위협하지 않으면서 변동될 수 있는 신체 상태나 감각을 나타낸다. 이에 따라, preferred prior $\tilde{p}(o)$를 $\tilde{p}(o^E, o^I)$로 분리하며, 본 연구에서는 독립성을 가정하여 $\tilde{p}(o^E, o^I) \approx \tilde{p}(o^E)\,\tilde{p}(o^I)$로 취급한다:
}
\else
%TC:endignore
The implementation of the self-prior in the active inference framework is derived from a simple modification to the standard expected free energy formulation, separating the self-prior term from the original preferred prior term.
Specifically, this is achieved by redefining each preferred observation $o$ in terms of two components: (\textit{i}) an \emph{extrinsic} observation $o^E$ that must be maintained to satisfy survival requirements, and (\textit{ii}) an \emph{intrinsic} observation $o^I$ related to the self-prior.
Here, $o^I$ refers to bodily states or sensations such as a tactile pattern on the arm, that can vary without threatening the agent's essential functions.
Consequently, we decompose the preferred prior $\tilde{p}(o)$ into $\tilde{p}(o^E, o^I)$, and in this study, we assume independence so that $\tilde{p}(o^E, o^I) \approx \tilde{p}(o^E)\,\tilde{p}(o^I)$:
%TC:ignore
\fi

\begin{equation}
  \begin{aligned}
    \text{Preferred prior:}~~~~~&\tilde{p}(o) = \tilde{p}(o^E, o^I) \approx \tilde{p}(o^E)\,\underbrace{\tilde{p}(o^I)}_{\mathclap{\textbf{Self-Prior}}}~~~
  \end{aligned}
  \label{eq:preferred_prior_decomposition}
\end{equation}

\ifshowkorean
\else
%TC:endignore
\rev{In our approach, we choose to define the self-prior over observations $o$ rather than internal states $s$.
This is because, as mentioned earlier, the self-prior can be derived through simple manipulation of the standard expected free energy formulation, enabling natural integration of fixed extrinsic preferences and experience based intrinsic preferences within the same framework.}
The fact that the self-prior is related to intrinsic motivation yet can emerge from external sensory data may seem counterintuitive.
This confusion stems from inconsistent use of "external-extrinsic" and "internal-intrinsic" terminology.
To resolve this confusion and rigorously classify intrinsic motivation, we adopt the computational classification of intrinsic motivation proposed by \citep{oudeyer_what_2009}:

\begin{itemize}
\item \textbf{External vs.\ Internal}: Referring to whether the reward \emph{computation} occurs in the environment or within the agent.
\item \textbf{Extrinsic vs.\ Intrinsic}: Referring to whether the reward \emph{criterion} is determined externally or internally.
% \item \emph{Homeostatic vs.\ heterostatic}, referring to whether the motivation aims to maintain a setpoint or to deviate from it.
\end{itemize}

According to these definitions, any reward provided from the external environment, such as a game score, qualifies as extrinsic motivation.
Meanwhile, internal motivation can be either extrinsic or intrinsic.
An internal motivation is considered intrinsic if the agent sets its reward criterion according to its internal model (e.g., measuring information gain based on a learned world model).
By contrast, an internal yet extrinsic motivation is one in which the agent checks its internal state and computes a reward based on an externally fixed criterion (e.g., "maintaining a battery level of 50\%").

Therefore, the key to distinguishing extrinsic and intrinsic motivation is not the source of information, but rather who sets the reward criterion.
In this sense, expected free energy is an internal framework since the agent itself computes both extrinsic and intrinsic components.
Furthermore, the self-prior proposed in this paper qualifies as \emph{internal intrinsic motivation} because the agent's internal model directly calculates and updates the reward criterion.

Additionally, \citep{oudeyer_what_2009} proposed that motivation can also be distinguished as \emph{homeostatic vs.\ heterostatic}.
For example, if there is an intrinsic motivation to pursue new information acquisition, this is heterostatic, whereas if there is an intrinsic motivation to maintain a familiar state, this is defined as homeostatic.

In active inference, the preferred prior has traditionally been interpreted as a fixed setpoint representing observations essential for survival (e.g., blood sugar level, battery level).
These setpoints have been considered extrinsic values since they are externally specified, but we regard this term as strictly representing a homeostatic value characteristic.
\rev{Therefore, the decomposition in Equation}~\eqref{eq:preferred_prior_decomposition}\rev{ separates the homeostatic value term in Equation}~\eqref{eq:efe}\rev{ into extrinsic and intrinsic terms:}
%TC:ignore
\fi

\begin{equation}
    \begin{aligned}
-\underbrace{
\mathbb{E}_{q(o_t)}[\log \tilde{p}(o_t)]
}_{\text{homeostatic value}} 
= 
-\underbrace{
\mathbb{E}_{q(o^E_t)}[\log \tilde{p}(o^E_t)]
}_{\text{extrinsic (homeostatic)}}
-\underbrace{
\mathbb{E}_{q(o^I_t)}[\log \tilde{p}(o^I_t)]
}_{\text{intrinsic (homeostatic)}}
    \end{aligned}
    \label{eq:preferred_prior_decomposition_in_efe}
\end{equation}

\ifshowkorean
{\devscismall
\uline{식~\eqref{eq:efe}에서 $\mathbb{E}_{q(o_t)}[\log \tilde{p}(o_t)]$를 extrinsic value로 표기한 것과 달리, 식~\eqref{eq:preferred_prior_decomposition_in_efe}에서는 homeostatic으로 표기한 것에 주의하라. 따라서 self-prior를 포함하도록 명시된 기대 자유 에너지는 다음과 같다:}
}
\else
%TC:endignore
\rev{Note that unlike Equation}~\eqref{eq:efe}\rev{, where $\mathbb{E}_{q(o_t)}[\log \tilde{p}(o_t)]$ is labeled as extrinsic value, Equation~}\eqref{eq:preferred_prior_decomposition_in_efe}\rev{ labels it as homeostatic.}
Thus, the expected free energy incorporating the self-prior becomes:
%TC:ignore
\fi

\begin{equation}
    \begin{aligned}
\mathcal{G} &\approx
-\underbrace{
\mathbb{E}_{q(o^E_t)}[
\log \tilde{p}(o^E_t)]
}_{\text{extrinsic (homeo-, fixed)}}
-\underbrace{
\mathbb{E}_{q(o^I_t)}[
\log \tilde{p}(o^I_t)]
}_{\text{intrinsic (homeo-, familiarity)}}
-
\underbrace{
\mathbb{E}_{q(o_t)}[D_{\mathrm{KL}}[q(s_t|o_t) \| q(s_t)]]
}_{\text{intrinsic (hetero-, novelty)}}
 - \mathcal{H'}
    \end{aligned}
    \label{eq:efe_with_self_prior}
\end{equation}

\ifshowkorean
{\devscismall
여기서 $\mathcal{H}'$는 행동 분포의 엔트로피를 의미한다.

정리하면, 본 연구의 self-prior는 에이전트가 신체 감각 경험에서 학습한 기준 패턴을 "유지하거나 복원"하는 것을 목표로 하므로, \emph{internal homeostatic intrinsic motivation}에 해당한다. 이는 \citep{oudeyer_what_2009}에서 제안한 내발적 동기의 분류 중에서 distributional familiarity motivation (DFM)과 가장 유사하다. Table~\ref{table:motivation_table}은 동기 분류법에서의 self-prior의 자리매김을 요약한다.

}
\else
%TC:endignore
Here, $\mathcal{H}'$ denotes the entropy of the action distribution.

In summary, the self-prior in this study aims to "maintain or restore a reference pattern" learned from the agent's sensory experiences, thus qualifying as \emph{internal homeostatic intrinsic motivation}.
This most closely resembles distributional familiarity motivation (DFM) among the classifications of intrinsic motivation proposed by \citep{oudeyer_what_2009}.
Table~\ref{table:motivation_table} summarizes the positioning of the self-prior within this motivational taxonomy.

%TC:ignore
\fi

\begin{table}[ht]
    \centering
    
    \caption{
      \ifshowkorean
      {\devsciscriptsize
      동기(Motivation) 분류법을 통한 Self-prior의 자리매김
      }
      \else
      %TC:endignore
      Positioning of the self-prior within a taxonomy of motivations
      %TC:ignore
      \fi
    }
    \label{table:motivation_table}
    
    \begin{tabular}{c | c | c | c}
      \toprule
               & Extrinsic     & \multicolumn{2}{c}{Intrinsic}   \\
      \cmidrule(r){2-4}
               & \multicolumn{2}{c|}{Homeostatic} & Heterostatic  \\
      \midrule
      External & Score in video game   & (N/A)      & (N/A)                  \\
      \hline
      Internal & \makecell{Gap of sensory data \\ from fixed value} & 
                 \makecell{\textbf{Gap of sensory data} \\ \textbf{from self-prior}} &
                 \makecell{Information gain \\ by sensory data} \\
      \bottomrule
    \end{tabular}
  
  \end{table}
  
\ifshowkorean
{\devscismall
% rev: self-prior 구축법의 대안 살짝 추가
\uline{Self-prior 개념을 구현하는 대안적인 방법으로서, 내부 상태에 대한 선호를 구축하는 접근도 존재한다. 이는 }\citep{sajid_exploration_2021}\uline{에서 제안된 방식이지만, 외발적 동기와 내발적 동기를 명시적으로 통합할 수 없었다. 또 다른 방법으로는, 계층적 모델의 상위 계층에서 학습된 내부 표현이 하위 계층의 self-prior로 변환되는 방식도 고려할 수 있다. 이러한 향후의 확장 가능성들에 대해서는 Discussion에서 다시 논한다.}
}
\else
%TC:endignore
\rev{As an alternative approach to implementing the self-prior concept, there also exist approaches that construct preferences over internal states.
This is the approach proposed by }\citep{sajid_exploration_2021}\rev{, but it could not explicitly integrate extrinsic and intrinsic motivations.
Another approach could consider hierarchical models wherein internal representations learned at higher levels are transformed into self-priors for lower levels.
We revisit these potential future extensions in the Discussion section.}
%TC:ignore
\fi
%TC:endignore

%TC:ignore
\section{Computational Model of Self-Prior}
\ifshowkorean
{\devscismall
이 섹션에서 우리는 self-prior를 포함한 능동적 추론의 프레임워크를 컴퓨터 시뮬레이션에서 계산하기 위한 구체적인 모델의 구현을 소개한다. 전체적인 프레임워크는 이산 환경에서는 모델의 거동을 직접 확인할 수 있도록 작은 크기의 행렬로 구성하며, 연속 환경에서는 심층 강화 학습에서 개발된 기법을 사용한 심층 신경망을 사용하도록 구성한다.

% rev: self-prior가 특정 과제에 연관되지 않는다는 설명 추가
\uline{우리는 self-prior로서 감각 데이터의 특정 부분을 필터하여 학습하는 등의 사전 준비가 필요하지 않은 end-to-end 학습 모델을 사용하여, 특정 과제에 직접적으로 연관되지 않은, 보다 일반적인 행동에도 적용 가능한 범용 구조로 구현한다. 구체적으로는 이산 환경을 위한 모델에서 경험적 빈도 정규화를 통한 최대가능도추정(plug-in MLE)를, 연속 환경을 위한 모델에서 Normalizing Flow 기반의 밀도 추정기를 최대가능도로 학습한다~}\citep{durkan_neural_2019}.
}
\else
%TC:endignore
In this section, we introduce the concrete implementation of our model for computational simulations based on the active inference framework incorporating the self-prior.
The overall framework is constructed using small-sized matrices in the discrete environment to allow direct inspection of model behavior, and deep neural networks using techniques developed in deep reinforcement learning in the continuous environment.

\rev{We implement this as a general-purpose structure applicable to a wide range of behaviors beyond specific tasks, using an end-to-end learning model that requires no prior preparation such as filtering specific parts of sensory data for the self-prior.
Specifically, in the discrete environment model, we learn maximum likelihood estimation via empirical frequency normalization (plug-in MLE), while in the continuous environment model, we train Normalizing Flow based density estimators via maximum likelihood}~\citep{durkan_neural_2019}.
%TC:ignore
\fi

\subsection{Model for Discrete Environment}
\ifshowkorean
{\devscismall
이산 환경을 위한 모델은 이산의 확률 변수를 다루는 이산 확률 분포인 카테고리 분포를 생성 모델로서 사용한다:
}
\else
%TC:endignore
The model for the discrete environment uses categorical distributions as the generative model, handling discrete random variables:
%TC:ignore
\fi

\begin{equation}
    \begin{aligned}
\text{Prior:}~~~~~&P(s_t \mid s_{t-1}, a_{t-1}) = \text{Cat}(\mathbf{B}_{a_{t-1}}) \\
\text{Posterior:}~~~~~&Q(s_t) = \text{Cat}(\phi_t) \\
\text{Likelihood:}~~~~~&P(o_t \mid s_t) = \text{Cat}(\mathbf{A})
    \end{aligned}
\end{equation}

\ifshowkorean
{\devscismall
% rev: appendix로 이동
여기서 대문자 $P,\,Q$는 이산 분포를 나타내며, $\mathbf{A},\, \mathbf{B},\, \phi_t$는 카테고리 분포의 파라미터 행렬 및 벡터이다. $\mathbf{B}$는 전이 모델을 구현하며, 각 행동 $a$에 대해 별도의 $\mathbf{B}_a$ 행렬이 존재한다. 이러한 표기법을 토대로, 변분 자유 에너지를 최소화하면 현재 관찰 $o_t$에 대한 최적 사후 분포 파라미터 $\phi_t$를 다음과 같이 Softmax 정규화 함수 $\sigma$를 통해 얻을 수 있다
\uline{(유도 과정은} 부록 \ref{subsubsec:variational_free_energy_derivation}\uline{ 참조)}:
}
\else
%TC:endignore
Here, uppercase letters $P,\,Q$ denote discrete distributions, and $\mathbf{A},\, \mathbf{B},\, \phi_t$ are parameter matrices and vectors of categorical distributions.
$\mathbf{B}$ implements the transition model, with separate $\mathbf{B}_a$ matrices for each action $a$.
Using this notation, minimizing variational free energy yields the optimal posterior distribution parameter $\phi_t$ for the current observation $o_t$ through the Softmax normalization function $\sigma$ as follows
\rev{(derivation process see Appendix} \ref{subsubsec:variational_free_energy_derivation}\rev{)}:
%TC:ignore
\fi

\begin{equation}
    \begin{aligned}
\phi_t &\approx \sigma(\log{(\mathbf{A} \cdot o_t)}  + \log{(\mathbf{B}_{a_{t-1}}\phi_{t-1})})
    \end{aligned}
    \label{eq:fe_discrete_post}
\end{equation}

\ifshowkorean
{\devscismall
우리는 가능도 분포 $\text{Cat}(\mathbf{A})$와 사전 분포 $\text{Cat}(\mathbf{B})$가 이미 정확하게 학습된 상태를 가정하고 논의를 전개한다. 즉, 파라미터 $\mathbf{A}$는 존재 가능한 모든 숨겨진 상태 $s_t$에 대해 각각 어떤 관찰 $o_t$을 갖는지 완벽하게 알고 있고, $\mathbf{B}$는 각 상태에 대해 존재 가능한 모든 행동 $a_t$의 실행 결과를 완벽하게 알고 있다. 이러한 단순화 가정은, self-prior의 변화에 따른 행동의 변화를 조사하는데 있어, 생성 모델의 다른 부분에서 발생할 수 있는 부작용을 배제하기 위한 것이다 (후술할 연속 환경을 위한 모델에는 가능도 분포와 사전 분포의 학습도 포함되어 있다).
}
\else
%TC:endignore
We assume that the likelihood distribution $\text{Cat}(\mathbf{A})$ and the prior distribution $\text{Cat}(\mathbf{B})$ have already been learned accurately.
That is, the parameter $\mathbf{A}$ perfectly knows which observation $o_t$ corresponds to each possible hidden state $s_t$, and $\mathbf{B}$ perfectly knows the outcome of each possible action $a_t$ for each state.
This simplifying assumption is intended to exclude side effects that may arise from other parts of the generative model when investigating changes in behavior due to changes in the self-prior (the learning of both likelihood and prior distributions is included in the model for the continuous environment discussed later).
%TC:ignore
\fi

\ifshowkorean
{\devscismall
미래의 관찰에 대한 기준점이 되는 분포 $\tilde{P}(o_t)$는 카테고리 분포 $\text{Cat}(\mathbf{C})$ 형태로 주어진다. 이 때 $\mathbf{C}$는 존재 가능한 $o_t$의 경우의 수만큼의 행을 갖는 열벡터이다.
% rev: self-prior의 전체를 C로 사용하는 설명 조금 더 상세히
Eq.~\eqref{eq:preferred_prior_decomposition}\uline{의 정의를 따르면 preferred prior $\tilde{p}(o_t)$는 외발적 성분 $\tilde{p}(o^E_t)$와 내발적 성분 $\tilde{p}(o^I_t)$로 분해되어야 하지만, 본 연구에서는 self-prior에 의해 유도된 순수한 내발적 동기 행동을 조사하기 위해 외발적 성분을 배제하였다.}
따라서 관찰 벡터 $o_t$의 모든 성분이 내발적 관찰 $o^I_t$에 해당하며, $\text{Cat}(\mathbf{C})$ 전체가 self-prior $\tilde{P}(o^I_t)$로 정의된다:
}
\else
%TC:endignore
The distribution $\tilde{P}(o_t)$ that serves as the setpoint for future observations is given in the form of a categorical distribution $\text{Cat}(\mathbf{C})$, where $\mathbf{C}$ is a column vector with a row for each possible $o_t$.
\rev{According to the definition in Eq.~}\eqref{eq:preferred_prior_decomposition}\rev{, the preferred prior $\tilde{p}(o_t)$ should decompose into an extrinsic component $\tilde{p}(o^E_t)$ and an intrinsic component $\tilde{p}(o^I_t)$, but in this study, we excluded the extrinsic component in order to investigate purely intrinsic motivational behavior driven by the self-prior.}
Therefore, all components of the observation vector $o_t$ correspond to intrinsic observations $o^I_t$, and the entire $\text{Cat}(\mathbf{C})$ is defined as the self-prior $\tilde{P}(o^I_t)$:
%TC:ignore
\fi

\begin{equation}
    \begin{aligned}
\text{Self-prior:}~~~~~&\tilde{P}(o^I_t) = \text{Cat}(\mathbf{C})
    \end{aligned}
    \label{eq:discrete_self_prior}
\end{equation}

\ifshowkorean
{\devscismall
% rev: appendix로 이동
에이전트는 지금까지 경험한 각 관찰의 횟수를 기록하고, 이에 따른 관찰 확률을 파라미터 $\mathbf{C}$에 저장하는 방식으로, self-prior를 스스로 학습한다. $\mathbf{C}$의 초기값은 모든 관찰에 대한 확률이 동일하도록 설정했다. 이를 바탕으로 최적 행동 $a_t$를 결정하기 위해, 각 행동에 대한 기대 자유 에너지를 계산한다
\uline{(유도 과정은 부록} \ref{subsubsec:expected_free_energy_derivation}\uline{ 참조)}:
}
\else
%TC:endignore
The agent autonomously learns the self-prior by recording the frequency of each observation experienced so far and storing the corresponding observation probabilities in the parameter $\mathbf{C}$.
The initial value of $\mathbf{C}$ is set so that all observations have equal probability.
To determine the optimal action $a_t$ based on this, we compute the expected free energy for each action
\rev{(derivation process see Appendix} \ref{subsubsec:expected_free_energy_derivation}\rev{)}:
%TC:ignore
\fi

\begin{equation}
    \begin{aligned}
    \mathcal{G} \approx (\mathbf{B}_{a_{t}}\phi_{t}) \cdot \mathcal{H}[\mathbf{A}] + D_{\mathrm{KL}}[\mathbf{A} \mathbf{B}_{a_{t}}\phi_{t} \| \mathbf{C}]
    \end{aligned}
    \label{eq:efe_discrete}
\end{equation}

\ifshowkorean
{\devscismall
% rev: 미래 정책에 사용되는 신념 설명
\uline{미래 정책을 평가하기 위해, 전이 모델 $\mathbf{B}$를 통해 현재 신념 $\phi_t$로부터 미래 신념 $\phi_{t+1:t+N}$을 "상상된 시간"으로 전파한다: $\phi_{t+\tau} = \mathbf{B}_{a_{t+\tau-1}}\phi_{t+\tau-1}$.} 각 후보 정책 $\pi = \{a_t, ..., a_{t+N}\}$에 대한 기대 자유 에너지를 합산하여 최적 정책을 선택한다. 정책은 카테고리 분포로 표현되며, 행동은 이로부터 샘플링된다:
}
\else
%TC:endignore
\rev{To evaluate future policies, we propagate the current belief $\phi_t$ through the transition model $\mathbf{B}$ into future beliefs $\phi_{t+1:t+N}$ along "imagined time": $\phi_{t+\tau} = \mathbf{B}_{a_{t+\tau-1}}\phi_{t+\tau-1}$.} The optimal policy is selected by summing expected free energies for each candidate policy $\pi = \{a_t, ..., a_{t+N}\}$. The policy is expressed as a categorical distribution, from which actions are sampled:
%TC:ignore
\fi
\begin{equation}
    \begin{aligned}
\text{Policy:}~~~~~&P(\pi) = \sigma(-\mathcal{G})
    \end{aligned}
    \label{eq:discrete_policy}
\end{equation}

\subsection{Model for Continuous Environment}

\ifshowkorean
{\devscismall
% rev: 리뷰어 제안 전통적 심층 추론 알고리즘 언급
\uline{자유 에너지 원리를 (고차원의) 연속 환경에서 적용하려는 선행 연구들에서, 많은 구현들은 관절각을 직접 내부 상태로 사용하여 해석적으로 유도된 backpropagation을 통해 행동을 생성한다 (예: } \citep{sancaktar_end_2020, priorelli_deep_2023}\uline{). 이러한 접근법들은 해석적인 계산이 가능하다는 장점이 있지만, 관절각으로 표현이 제한되어 복잡한 감각운동 정보를 고차원 잠재 공간에 암묵적으로 인코딩하기 어렵고, 장기 계획이 필요한 복잡한 과제에 적용하기 어렵다는 한계가 있다.}

\uline{반면, 심층 신경망을 활용하는 접근법은 생물학적 세부 사항을 직접 모방하지는 않지만, 능동적 추론의 핵심 계산 원리를 고차원 문제로 확장할 수 있는 실용적인 방법을 제공한다 }\citep{millidge_deep_2020}\uline{. 본 연구의 초점은 self-prior라는 개념적 메커니즘을 제안하고 증명하는 것이며, 이는 특정 신경 구현에 구속되지 않는다. 따라서 우리는 심층 신경망 기반 접근법을 따라, 복잡한 감각운동 정보를 고차원 잠재 공간에 표현하고 미래 계획을 수행할 수 있도록 한다.}
% rev: RSSM이 VAE 확장임을 명시
구체적으로, 본 연구에서는 PlaNet \citep{hafner_planet_2019}의 Recurrent State-Space Model (RSSM)을 변분 자유 에너지를 최적화하는 과정으로 해석하고 \citep{catal_learning_2020}, 정책 그래디언트 방법을 기대 자유 에너지를 최소화하는 과정으로 해석하는 선행 연구의 기법을 따른다 \citep{millidge_deep_2020}.
\uline{RSSM은 본질적으로 Variational Autoencoder (VAE, }\citep{kingma2013auto}\uline{) 구조를 시계열 데이터로 확장한 것으로, 변분 자유 에너지 최소화의 원리를 직접 구현한다.}
}
\else
%TC:endignore
\rev{In prior research attempting to apply the free energy principle to (high-dimensional) continuous environments, many implementations use joint angles directly as internal states and generate actions through analytically derived backpropagation (e.g., }\citep{sancaktar_end_2020, priorelli_deep_2023}\rev{).
While these approaches offer the advantage of analytical computation, they are limited in that the representation is constrained to joint angles, making it difficult to implicitly encode complex sensorimotor information in high-dimensional latent spaces and challenging to apply to complex tasks requiring long-term planning.}

\rev{In contrast, approaches utilizing deep neural networks do not directly mimic biological details, but they provide a practical method to extend the core computational principles of active inference to high-dimensional problems }\citep{millidge_deep_2020}\rev{.
The focus of this study is to propose and demonstrate the conceptual mechanism of the self-prior, which is not bound to any specific neural implementation.
Therefore, we follow the deep neural network based approach to enable representation of complex sensorimotor information in high-dimensional latent spaces and to perform future planning.}
Specifically, in this study, we follow the approach of prior research that interprets the Recurrent State-Space Model (RSSM) in PlaNet \citep{hafner_planet_2019} as optimizing variational free energy \citep{catal_learning_2020}, and interprets policy gradient methods as minimizing expected free energy \citep{millidge_deep_2020}.
\rev{The RSSM is essentially an extension of the Variational Autoencoder (VAE, }\citep{kingma2013auto}\rev{) architecture to sequential data, directly implementing the principle of variational free energy minimization.}
%TC:ignore
\fi

\ifshowkorean
{\devscismall
이산 환경을 위한 모델에서는 변분 자유 에너지에서의 가능도 분포와 사전 분포의 파라미터가 고정되어 있었지만, 연속 환경을 위한 모델은 표준 RSSM 구조를 따라 심층 신경망을 사용하여 파라미터 $\phi$에 대한 학습을 진행한다.
% rev: RSSM의 변분 분해 명시
\uline{RSSM은 결정론적 잠재 상태 $h_t$와 확률론적 잠재 상태 $s_t$를 분리하여 시계열 분해를 가능하게 한다:}
}
\else
%TC:endignore
Whereas the discrete model fixed the parameters of the likelihood and prior distributions in variational free energy, the continuous model follows the standard RSSM structure and trains the parameters $\phi$ using deep neural networks. \rev{The RSSM separates the deterministic latent state $h_t$ and the stochastic latent state $s_t$, enabling proper sequential decomposition:}
%TC:ignore
\fi

\begin{equation}
    \begin{aligned}
\text{Deterministic state:}~~~~~& h_t = f_{\phi}(h_{t-1}, s_{t-1}, a_{t-1}) \\
\text{Stochastic prior:}~~~~~& p_{\phi}(\hat{s}_t \mid h_t) \\
\text{Stochastic posterior:}~~~~~& q_{\phi}(s_t \mid h_t, o_t) \\
\text{Likelihood:}~~~~~& p_{\phi}(o_t \mid h_t, s_t)
    \end{aligned}
    \label{eq:rssm}
\end{equation}

\ifshowkorean
{\devscismall
% rev: 변분 분해 언급, 부록 추가
\uline{결정론적 상태는 GRU cell을 통해 시계열 의존성을 포착하고, 확률적 사전과 사후는 각각 결정론적 상태 (및 관찰)로부터 가우시안 분포 파라미터를 생성한다. 이러한 분리는 각 시간 단계에서 적절한 변분 분해를 보장한다. 이러한 RSSM 구조를 구성하는 모든 파라미터 $\phi$는 변분 자유 에너지를 최소화하도록 경사 하강법에 의해 학습된다}
\uline{(구현 세부사항은 부록} \ref{subsubsec:rssm_computation}\uline{ 참조)}.
}
\else
%TC:endignore
\rev{The deterministic state captures temporal dependencies through a GRU cell, while the stochastic prior and posterior each generate Gaussian distribution parameters from the deterministic states (and observations). This separation ensures proper variational decomposition at each time step}
\rev{(implementation details in Appendix} \ref{subsubsec:rssm_computation}\rev{)}.
%TC:ignore
\fi

\begin{figure}[t]
  \centering
    \begin{subfigure}[htbp]{0.49\textwidth}
      \centering
      \includegraphics[width=1.0\linewidth]{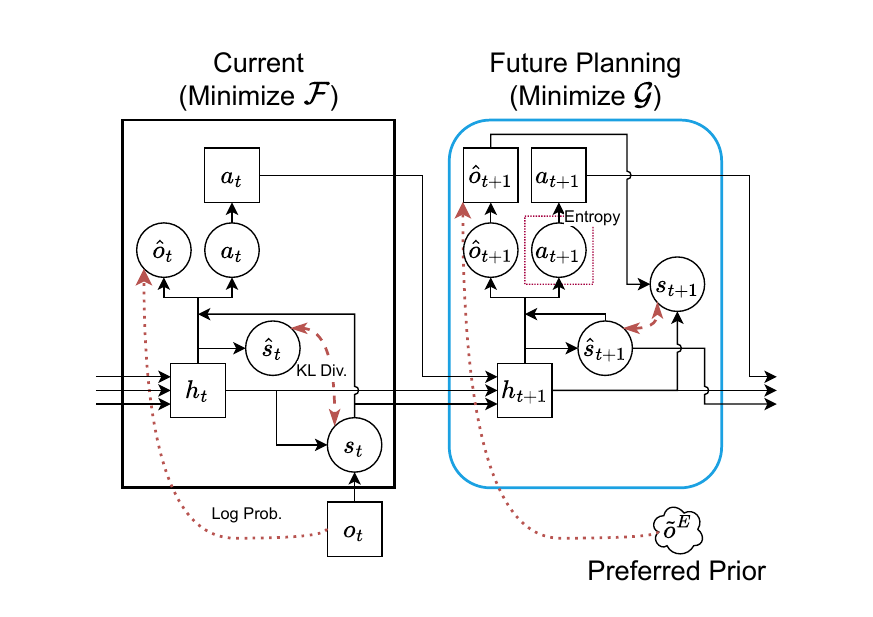}
      \caption{Original active inference}
    \end{subfigure}
    \begin{subfigure}[htbp]{0.49\textwidth}
      \centering
      \includegraphics[width=1.0\linewidth]{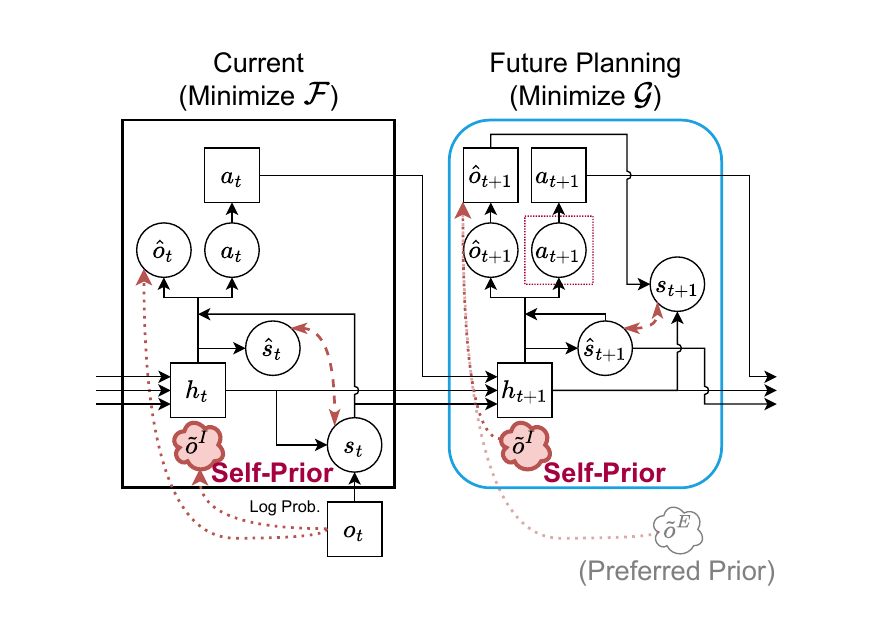}
      \caption{Active inference with self-prior}
    \end{subfigure}
  \caption
  {
    \ifshowkorean
    {\devsciscriptsize 심층 신경망을 사용하여 변분 자유 에너지 $\mathcal{F}$와 기대 자유 에너지 $\mathcal{G}$를 최소화하는 능동적 추론의 그래프(graph) 모델. Self-prior $\tilde{p}(o^I_t)$는 관찰 $o^I_t$의 로그 가능도를 최대화하도록 학습된다. 하늘색으로 표시된 기대 자유 에너지 계산에서, 학습된 기준점인 self-prior는 고정된 기준점인 preferred prior와 함께 행동의 기준점을 제공한다. 이론상 Preferred prior와 self-prior는 동시에 적용할 수 있지만, 본 연구에서는 논의의 간결함을 위해 self-prior만 활용하였으므로, 그림에서 preferred prior는 흐리게 표시되었다.
    }
    \else
    %TC:endignore
    Graphical model of active inference using deep neural networks that minimize variational free energy $\mathcal{F}$ and expected free energy $\mathcal{G}$. The self-prior $\tilde{p}(o^I_t)$ is trained to maximize the log-likelihood of observations $o^I_t$. In the expected free energy calculation (highlighted in blue), the learned self-prior serves as the behavioral setpoint alongside the fixed preferred prior. Although the preferred prior and self-prior can theoretically be applied simultaneously, we use only the self-prior in this study for clarity of exposition; thus, the preferred prior is shown faded in the figure.
    %TC:ignore
    \fi
  }
  \label{fig:fig2}
\end{figure}

\ifshowkorean
{\devscismall
연속 환경을 위한 모델도 self-prior에 의해 유도된 순수한 내발적 동기 행동을 조사하므로 $\tilde{p}(o_t) = \tilde{p}(o^I_t)$로 설정했다. Self-prior는 Neural Spline Flows (NSF, \citep{durkan_neural_2019})를 사용하여 관찰의 로그 가능도를 최대화하도록 학습된다. 이는 최근에 경험한 궤적을 저장하고 있는 Replay buffer $\mathcal{D}$에서 관찰 $o$를 샘플링하여 이루어진다:
}
\else
%TC:endignore
The continuous model also investigates purely intrinsic motivational behavior, thus $\tilde{p}(o_t) = \tilde{p}(o^I_t)$.
The self-prior is learned using Neural Spline Flows (NSF, \citep{durkan_neural_2019}) to maximize observation log-likelihood.
This is done by sampling observations $o$ from a replay buffer $\mathcal{D}$ that stores recently experienced trajectories:
%TC:ignore
\fi

\begin{equation}
    \begin{aligned}
\text{Self-prior:}&~~~~~\tilde{p}_{\xi}(o_t) \\
\argmin_\xi \mathcal{L_{\text{self}}} 
&= \argmin_\xi \mathbb{E}_{o\sim\mathcal{D}}[
  - \log{ \tilde{p}_{\xi}(o)}
]
    \end{aligned}
\end{equation}

\ifshowkorean
{\devscismall
연속 환경에서는 행동 후보를 모두 열거할 수 없으므로, 정책 경사법을 사용한다. \citep{millidge_deep_2020}를 따라 상태로부터 행동을 예측하는 Policy network와 무한 지평선에 이르는 기대 자유 에너지를 추정하는 Value network를 학습한다:
}
\else
%TC:endignore
Since action candidates cannot be enumerated in continuous environments, we use policy gradient methods.
Following \citep{millidge_deep_2020}, we train a policy network that predicts actions from states and a value network that estimates expected free energy over an infinite horizon:
%TC:ignore
\fi

\begin{equation}
    \begin{aligned}
\text{Policy:}~~~~~&q_\theta(a_t \mid s_t) \\
\text{Expected utility:}~~~~~& g_\psi(s_t)
    \end{aligned}
\end{equation}

\ifshowkorean
{\devscismall
% rev: 부록 추가
이는 Soft Actor-Critic \citep{haarnoja_sac_2018}과 유사한 접근이며, GAE($\lambda$) 추정 \citep{schulman_gae_2015}을 통해 기대 효용을 근사한다.
\uline{학습의 세부사항과 하이퍼파라미터는 부록} \ref{subsubsec:policy_and_value_network_training}\uline{ 및 }\citep{mazzaglia_contrastive_2021}\uline{을 참조하라.}
}
\else
%TC:endignore
This is a similar approach to Soft Actor-Critic \citep{haarnoja_sac_2018}, and approximates expected utility through GAE($\lambda$) estimation \citep{schulman_gae_2015}.
\rev{Training details and hyperparameters are provided in Appendix} \ref{subsubsec:policy_and_value_network_training}\rev{ and }\citep{mazzaglia_contrastive_2021}.
%TC:ignore
\fi
%TC:endignore

%TC:ignore
\section{Experiments and Results}
\ifshowkorean
{\devscismall
이번 장(section)에서 우리는 multimodal 감각 경험들을 학습한 self-prior에 의한 목표 지향적인 행위의 창발을 컴퓨터 시뮬레이션을 통해 확인한다. 앞서 간략하게 설명했듯이, 우리의 실험은 (\textit{i}) 보통 자신의 몸에 스티커가 붙어있지 않은 경우가 많은 아기는 "내 몸에는 스티커가 없다"고 하는 self-prior를 획득하게 되고, (\textit{ii}) 이 아기에게 스티커를 붙여주면 감각 관찰과 self-prior 사이의 불일치에 의해 자유 에너지가 증가하므로, (\textit{iii}) 자유 에너지를 감소시키기 위해서 자발적으로 스티커로의 리칭을 목표로 하는 행동을 생성한다는 가정 하에 이루어졌다.
%rev: self-prior는 사전 설정이 없다는 것을 재강조
\uline{중요한 점은, 우리는 스티커를 제거하거나 특정 자세를 취하는 것을 선호 상태로 사전에 설정하지 않았으며, 모든 목표 지향적 행동은 에이전트가 모터 버블링을 통해 경험한 감각으로부터 자율적으로 형성된 self-prior에서 창발한다.}
}
\else
%TC:endignore
In this section, we validate the emergence of goal-directed behavior driven by the self-prior, which is learned from multimodal sensory experiences, through computer simulations.
As briefly mentioned earlier, our experiment is based on the following assumptions: (\textit{i}) since an infant typically does not have stickers attached to its body, it acquires a self-prior that "there are no stickers on my body"; (\textit{ii}) when a sticker is attached, the mismatch between sensory observations and the self-prior increases free energy; and (\textit{iii}) to reduce free energy, the agent spontaneously generates reaching behavior aimed at the sticker.
\rev{Importantly, we did not preset any preferred states such as removing stickers or adopting specific postures; all goal-directed behaviors emerge from the self-prior autonomously formed from the sensations the agent experiences through motor babbling.}
%TC:ignore
\fi

\ifshowkorean
{\devscismall
이러한 실험을 통해, 우리는 (\textit{i}) own preference를 나타내는 self-prior가 스스로 형성되는 과정을 검증하고, (\textit{ii}) self-prior를 포함한 자유 에너지 최소화 과정에서 reaching 행동이 창발하는 것을 확인한다. 더욱이, 이산 환경에서의 실험에서는 (\textit{iii}) self-prior의 변화에 따라 동일한 자극에 대해서도 행동이 달라질 수 있음도 추가적으로 확인한다.
}
\else
%TC:endignore
Through these experiments, we (\textit{i}) verify the process by which the self-prior, representing the agent's own preference, is autonomously formed, (\textit{ii}) confirm the emergence of reaching behavior within the free energy minimization framework incorporating the self-prior, and (\textit{iii}) in experiments in the discrete environment, additionally show that behavior can vary for the same stimulus depending on changes in the self-prior.
%TC:ignore
\fi

\subsection{Simulation in Discrete Environment}

\ifshowkorean
{\devscismall
이산 환경에서 에이전트는 이산적인 고유 감각과 촉각으로 이루어진 멀티모달 관찰을 받는다 (Figure~\ref{fig:fig3}). 에이전트의 오른손은 $0, 1, 2, 3, 4$ 중 하나의 위치를 가질 수 있으며 (5개의 경우의 수), 이 값이 고유 감각 상태를 나타낸다. 왼팔에는 3개의 촉각 센서가 있으며, 이는 오른손의 좌표계에서 위치 $1, 2, 3$에 해당하고, 각각의 촉각은 동시에 활성화 될 수 있다 ($2^3=8$개의 경우의 수). 따라서 이 환경에서 관찰 변수 $o_t$는 $5\times8=40$개의 경우의 수를 가질 수 있으며, 40행의 one-hot 열벡터로 표현된다~(Table \ref{table:discrete_obs_table}).
}
\else
%TC:endignore
In the discrete environment, the agent receives multimodal observations composed of discrete proprioceptive and tactile inputs (Figure~\ref{fig:fig3}). The agent's right hand can occupy one of five positions, labeled $0, 1, 2, 3,$ or $4$, representing its proprioceptive state. The left arm has three tactile sensors, located at positions $1, 2,$ and $3$ in the right hand’s coordinate frame, and each sensor can be activated independently, yielding $2^3=8$ possible tactile states. Consequently, the observation variable $o_t$ can take $5\times8=40$ distinct values and is represented as a 40-dimensional one-hot column vector (Table \ref{table:discrete_obs_table}).
%TC:ignore
\fi

\begin{table*}[t]
\centering
  \begin{minipage}[ht]{0.45\textwidth}
    \centering
    \includegraphics[width=0.9\linewidth]{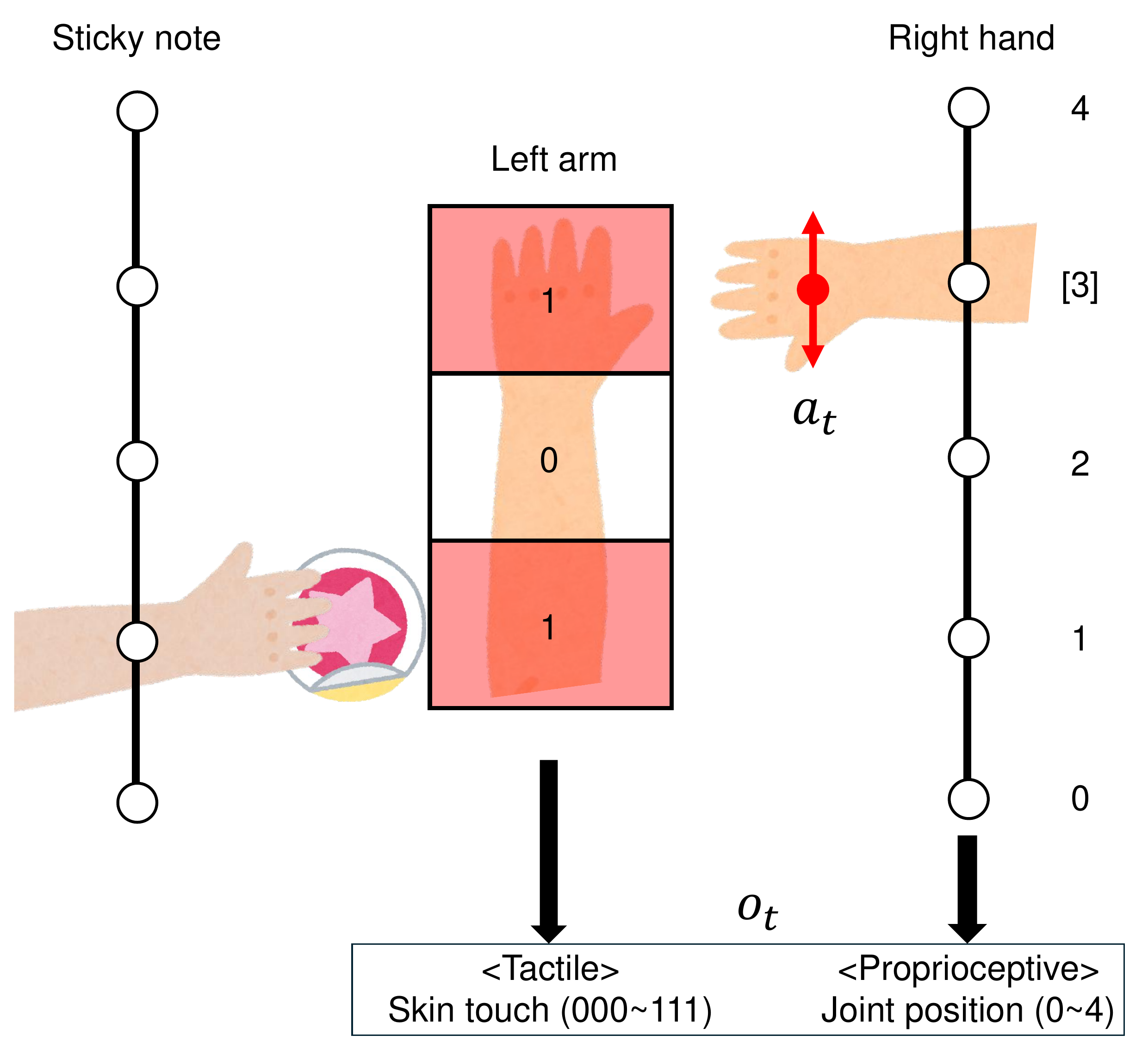}
    \captionof{figure}
    {
    \ifshowkorean
    {\devsciscriptsize 이산 환경의 개요. 오른손은 왼팔 위 또는 왼팔 밖을 좌우로 움직일 수 있고, 오른손이 있는 곳이나 스티커가 있는 곳에서 촉각이 발생한다.
    }
    \else
    %TC:endignore
     Overview of the discrete environment. The right hand can move left or right either above the left arm or outside of it, and tactile input occurs either where the right hand is located or where the sticker is attached.
    %TC:ignore
    \fi
    }
    \label{fig:fig3}
  \end{minipage}
\hspace{0.05\textwidth}
  \begin{minipage}[ht]{0.45\textwidth}
  \centering
  % Table content extracted from chapter/04result.tex
% This table can be used both in the main document (within minipage) 
% and in a separate tables-only document (as standalone table)

% For use in main document: wrap with minipage
% For use in tables-only document: wrap with table environment

  \captionof{table}{
    \ifshowkorean
    {\devsciscriptsize
     이산 환경에서 존재 가능한 에이전트의 감각 관찰 조합
    }
    \else
    %TC:endignore
    Possible combinations of sensory observations in the discrete environment
    %TC:ignore
    \fi
  }
  \label{table:discrete_obs_table}
  
    \begin{tabular}{c | c | c}
        \toprule
        $o_t$ & Touch & Hand position \\
              & (000 $\sim$ 111) & (0 $\sim$ 4) \\
        \midrule
            0 & 000 & 0 \\
        \hline
            1 & 000 & 1 \\
        \hline
            \multicolumn{1}{c}{$\vdots$} & \multicolumn{1}{c}{$\vdots$} & \multicolumn{1}{c}{$\vdots$} \\
        \hline
            28 & 101 & 3 \\
        \hline
            \multicolumn{1}{c}{$\vdots$} & \multicolumn{1}{c}{$\vdots$} & \multicolumn{1}{c}{$\vdots$} \\
        \hline
            39 & 111 & 4 \\
        \bottomrule
    \end{tabular}

\end{minipage}
\end{table*}

\ifshowkorean
{\devscismall
보육자는 에이전트의 왼팔에 스티커를 부착할 수 있다. 특정 위치(예: 위치 $x$)에 스티커가 붙어 있으면, 오른손의 위치에 관계없이 에이전트는 위치 $x$에서 지속적으로 촉각 신호를 받는다. 만약 오른손이 스티커가 붙어 있는 위치와 정확히 일치하면 단일한 촉각 신호가 감지된다.
}
\else
%TC:endignore
A caregiver can attach a sticker to the agent's left arm. If a sticker is placed at a specific position (e.g., position $x$), the agent continuously receives a tactile signal from position $x$, regardless of the right hand's position. If the right hand exactly matches the sticker's location, the agent perceives a single tactile signal.
%TC:ignore
\fi

\ifshowkorean
{\devscismall
각 시간 단계에서 에이전트는 다음 세 가지 행동 $a_t$ 중 하나를 선택할 수 있다: (\textit{i}) 정지, (\textit{ii}) 한 칸 왼쪽으로 이동, (\textit{iii}) 한 칸 오른쪽으로 이동. 따라서 $a_t$는 3행의 one-hot 열벡터로 표현된다. 이산 환경 실험에서 정책 $\pi$는 4스텝 미래까지의 가능한 행동 시계열인 $3^4=81$개가 존재하며, 각각의 기대 자유 에너지의 합에 따라 식 \ref{eq:discrete_policy}에 의해 확률적으로 결정된다.
}
\else
%TC:endignore
At each time step, the agent can choose one of three actions $a_t$: (\textit{i}) remain stationary, (\textit{ii}) move one step to the left, (\textit{iii}) move one step to the right.
Thus, $a_t$ is represented as a 3-dimensional one-hot column vector.
In the discrete environment experiment, there are $3^4=81$ possible policies $\pi$, which are sequences of actions up to 4 steps into the future, and are probabilistically determined by Equation \ref{eq:discrete_policy} according to the sum of expected free energies for each.
%TC:ignore
\fi

\ifshowkorean
{\devscismall
실험 시작 시에는 self-prior가 모든 가능한 관찰에 대해 균등하게 초기화되어 있다. 이는 모든 정책에 대해 기대 자유 에너지를 동등하게 만들어서, 무작위로 정책을 선택하게 만든다. 즉, 스티커가 붙더라도 특별한 반응이 시작되지 않는다 않는다~(Figure~\ref{fig:fig5a}).
}
\else
%TC:endignore
At the beginning of the experiment, the self-prior was uniformly initialized across all possible observations. This makes the expected free energy equal for all policies, causing random policy selection. That is, no special response occurs even when a sticker is attached~(Figure~\ref{fig:fig5a}).
%TC:ignore
\fi

\begin{figure}[t]
  \centering
  \includegraphics[width=1.0\linewidth]{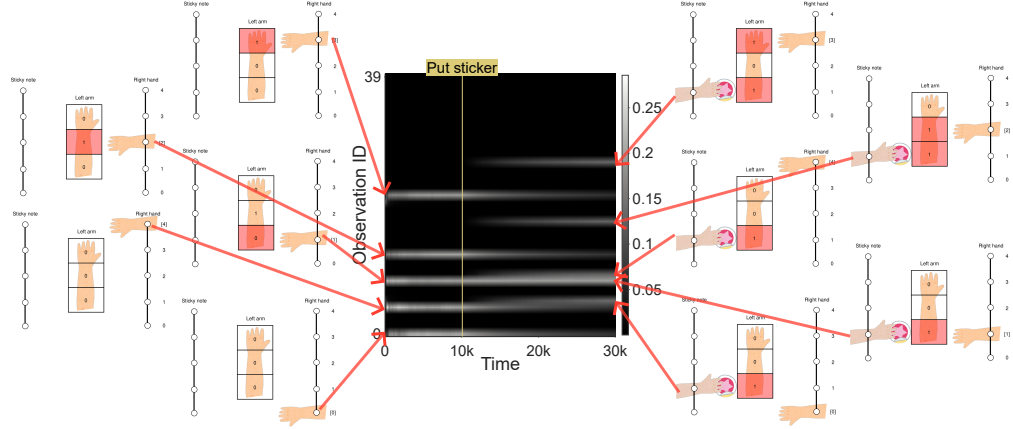}
  \caption{
    \ifshowkorean
    {\devsciscriptsize 시간에 따른 self-prior의 변화. 스티커가 붙기 전까지는 팔 위에 스티커가 없는 상황에 대한 확률이 높아진다 ($t<10,000$). 스티커가 붙고 나서는, 점점 스티커가 붙은 상황에 적응하게 된다 ($t\ge10,000$).}
    \else
    %TC:endignore
    Change in self-prior over time. Before the sticker is attached, the probability increases for situations where no sticker is present on the arm ($t<10,000$). After the sticker is attached, the agent gradually adapts to the new situation where the sticker is present ($t\ge10,000$).
    %TC:ignore
    \fi
  }
  \label{fig:fig4}
\end{figure}

\begin{figure}[t]
  \centering
    \begin{subfigure}[htbp]{0.49\textwidth}
      \centering
      \includegraphics[width=0.8\linewidth]{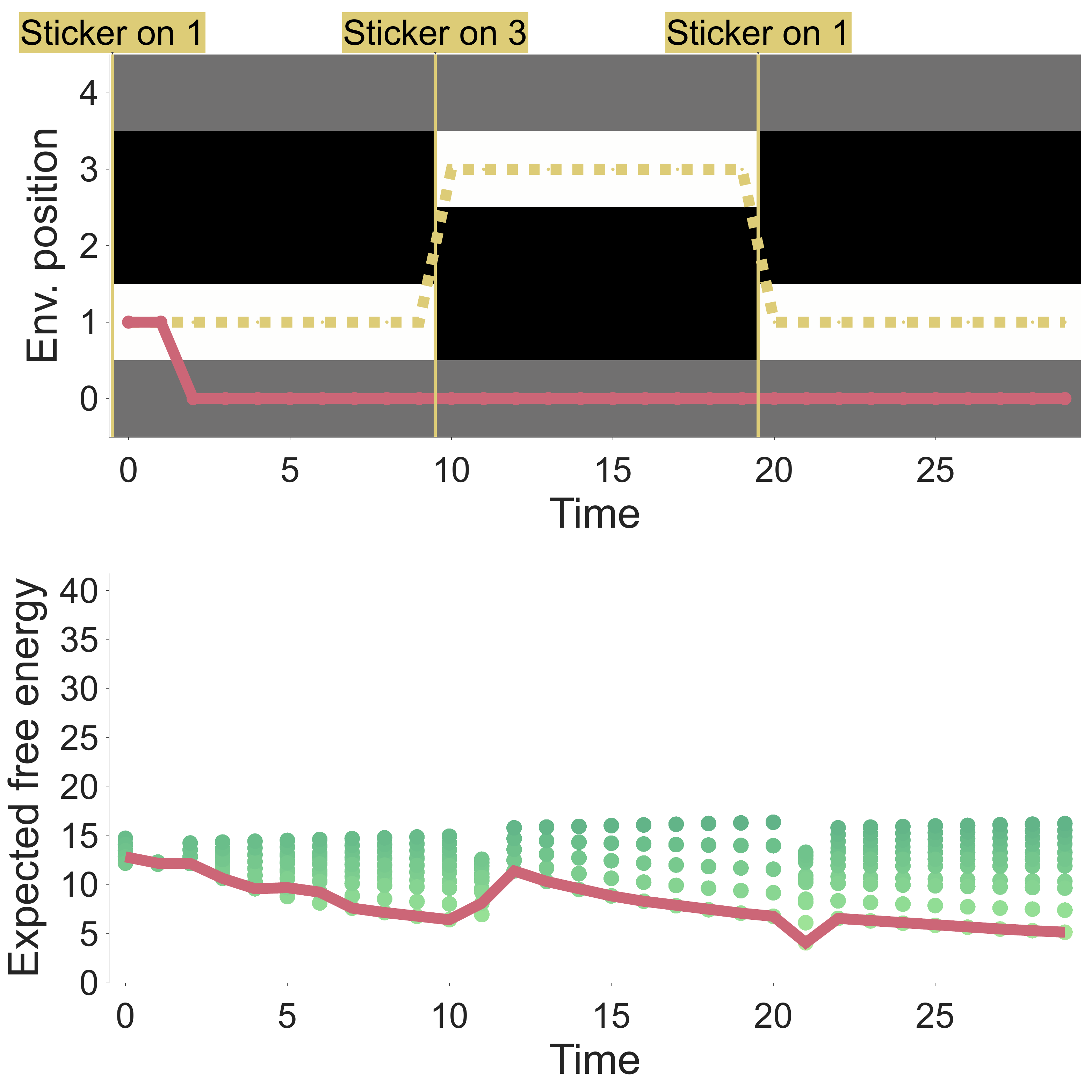}
      \caption{Before acquiring the self-prior ($t$ = 0)}
      \label{fig:fig5a}
    \end{subfigure}
    \begin{subfigure}[htbp]{0.49\textwidth}
      \centering
      \includegraphics[width=0.8\linewidth]{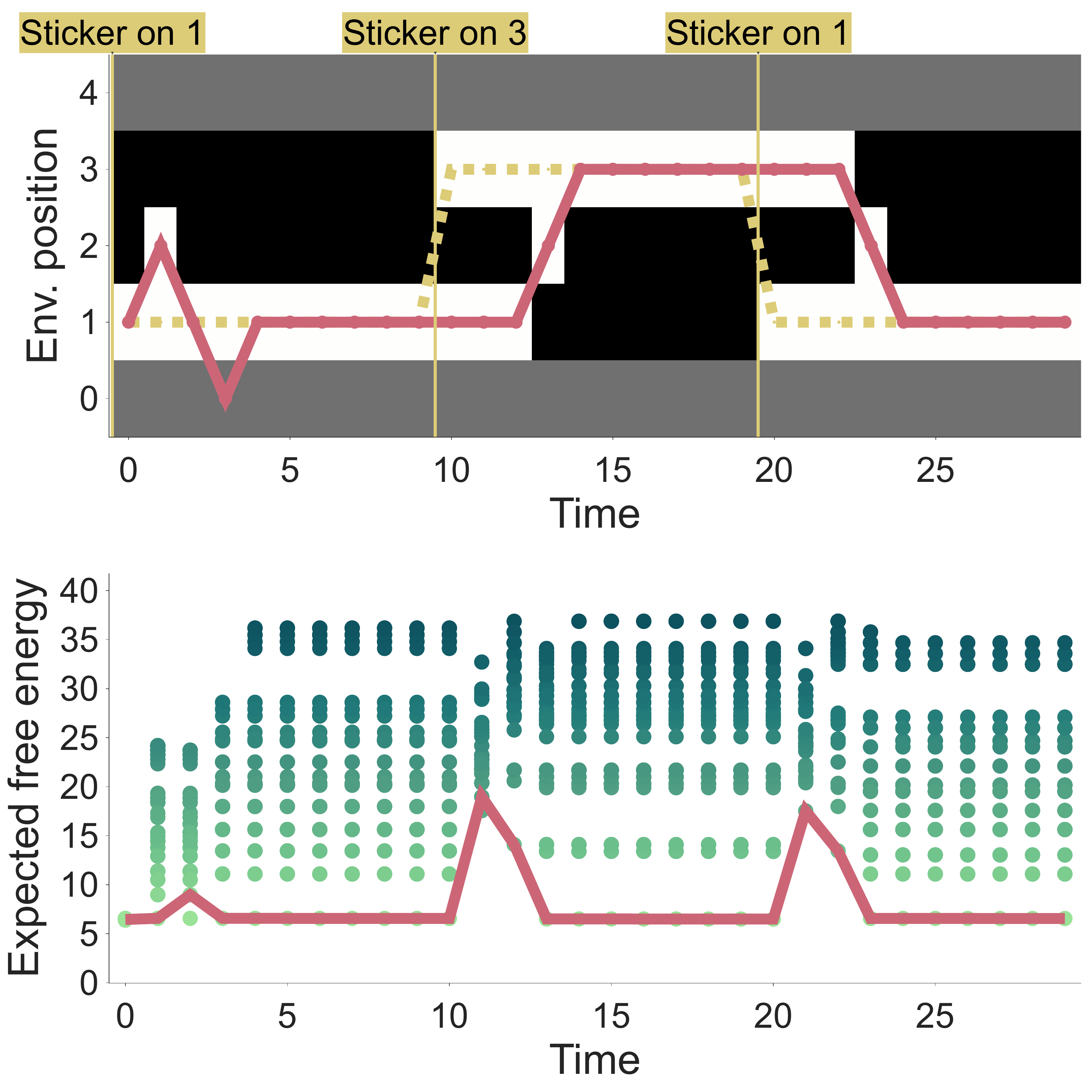}
      \caption{After acquiring the self-prior ($t$ = 10,000)}
      \label{fig:fig5b}
    \end{subfigure}
  \caption
  {
\ifshowkorean
    {\devsciscriptsize Self-prior 획득 전과 후의 에이전트의 행동 비교. 위쪽 그림은 시간에 따른 환경의 변화를 간략하게 나타낸 것이며, 빨간 선은 손의 위치, 노란 점선은 스티커의 위치를 나타낸다. 하얀 구간은 촉각이 발생한 곳이고, 검정 구간은 촉각이 발생하지 않은 곳이다. 회색 구간은 왼팔 밖이므로, 항상 촉각이 발생하지 않는다. 아랫쪽 그림은 시간에 따른 기대 자유 에너지를 나타낸 것이며, 각 초록색 점은 policy별 자유 에너지를 나타낸다. 자유 에너지가 낮은 policy일수록 선택될 확률이 높아지며, 빨간 선은 실제로 채택된 policy를 이은 것이다. (a) 처음에 Self-prior를 획득하기 전에는, 팔에 스티커가 붙어도 반응하지 않는다. (b) Self-prior를 획득한 후에는, 팔에 스티커가 붙으면, 해당 위치로 손을 움직이는 목표 지향적인 행동이 창발했다.}
\else
%TC:endignore
Comparison of the agent's behavior before and after acquiring the self-prior. The top panel illustrates environmental changes over time: the red line indicates the hand position, the yellow dashed line indicates the sticker position. White areas denote where tactile feedback occurred, black areas indicate no tactile feedback, and gray areas represent regions outside the arm where tactile feedback never occurs. The bottom panel shows expected free energy over time. Each green dot represents the free energy of a candidate policy, with lower free energy policies being more likely to be selected. The red line connects the actually selected policies. (a) Before acquiring the self-prior, the agent does not respond even when a sticker is attached to the arm. (b) After acquiring the self-prior, goal-directed behavior emerges: the agent moves its hand to the sticker’s location when it appears on the arm.
%TC:ignore
\fi
  }
\end{figure}

\ifshowkorean
{\devscismall
그 후, 에이전트를 충분히 긴 시간(10,000 스텝) 동안 제약 없이 motor babbling을 수행하게 하였다. 실제 유아의 경우, 이러한 경험은 외발적 선호를 충족시키는 다양한 일상 경험에 해당할 수 있지만, 본 실험에서는 단순화를 위해 에이전트의 행동 $a$를 무작위로 선택하는 motor babbling을 수행했다. 이 과정에서, “오른손이 왼팔 위에 있을 때 단일한 촉각 신호가 발생”하여, 그 결과 “오른손이 왼팔 밖에 있으면 촉각 신호가 발생하지 않음”이라는 패턴이 학습된다 (Left of Figure~\ref{fig:fig4}).

이 때, 보육자가 스티커를 왼팔의 position 1에 부착하면, 오른손이 위치한 position 4가 아닌 다른 위치에서 촉각 신호가 발생한다. 새롭게 학습된 self-prior에 따르면, 이는 낮은 확률의 상황이므로 자유 에너지가 증가하게 된다 (즉, 모델과의 불일치로 인해 주의가 집중됨). 이에 따라, 에이전트는 기대 자유 에너지를 감소시키기 위한 행동을 계획하기 시작하고, 스티커를 만지는 것이 self-prior와 가장 잘 일치한다고 결론짓는다. 그 결과, 에이전트는 position 1을 향해 손을 뻗는 목표 지향적 움직임을 보인다~(Figure~\ref{fig:fig5b}). 마찬가지로, 보육자가 스티커를 position 3으로 옮기면, 에이전트는 이 새로운 신호가 자신의 모델과 불일치함을 감지하고 즉시 position 3으로 팔을 뻗는다.
}
\else
%TC:endignore
Subsequently, we let the agent perform motor babbling without constraints for a sufficiently long period (10,000 steps). In the case of real infants, such experiences may correspond to a variety of everyday experiences that fulfill extrinsic preferences. However, for simplicity, we performed motor babbling in which the agent randomly selects its actions $a$. Through this process, the pattern that "a single tactile signal occurs when the right hand is over the left arm" is learned, and as a result, "no tactile signal occurs when the right hand is off the left arm" (Left of Figure~\ref{fig:fig4}).

At this point, when a caregiver attaches a sticker to position~1 on the left arm, a tactile signal occurs at a location different from the current hand position (position~4). According to the newly learned self-prior, this is a low-probability situation, resulting in an increase in free energy (i.e., attention is drawn due to the mismatch with the model). Accordingly, the agent begins planning actions to reduce expected free energy and concludes that touching the sticker aligns best with its self-prior. As a result, the agent shows goal-directed movement toward position~1~(Figure~\ref{fig:fig5b}). Similarly, when the caregiver moves the sticker to position~3, the agent detects this new signal as incongruent with its model and immediately reaches toward position~3.
%TC:ignore
\fi

\begin{figure}[ht]
  \centering
  \includegraphics[width=0.45\linewidth]{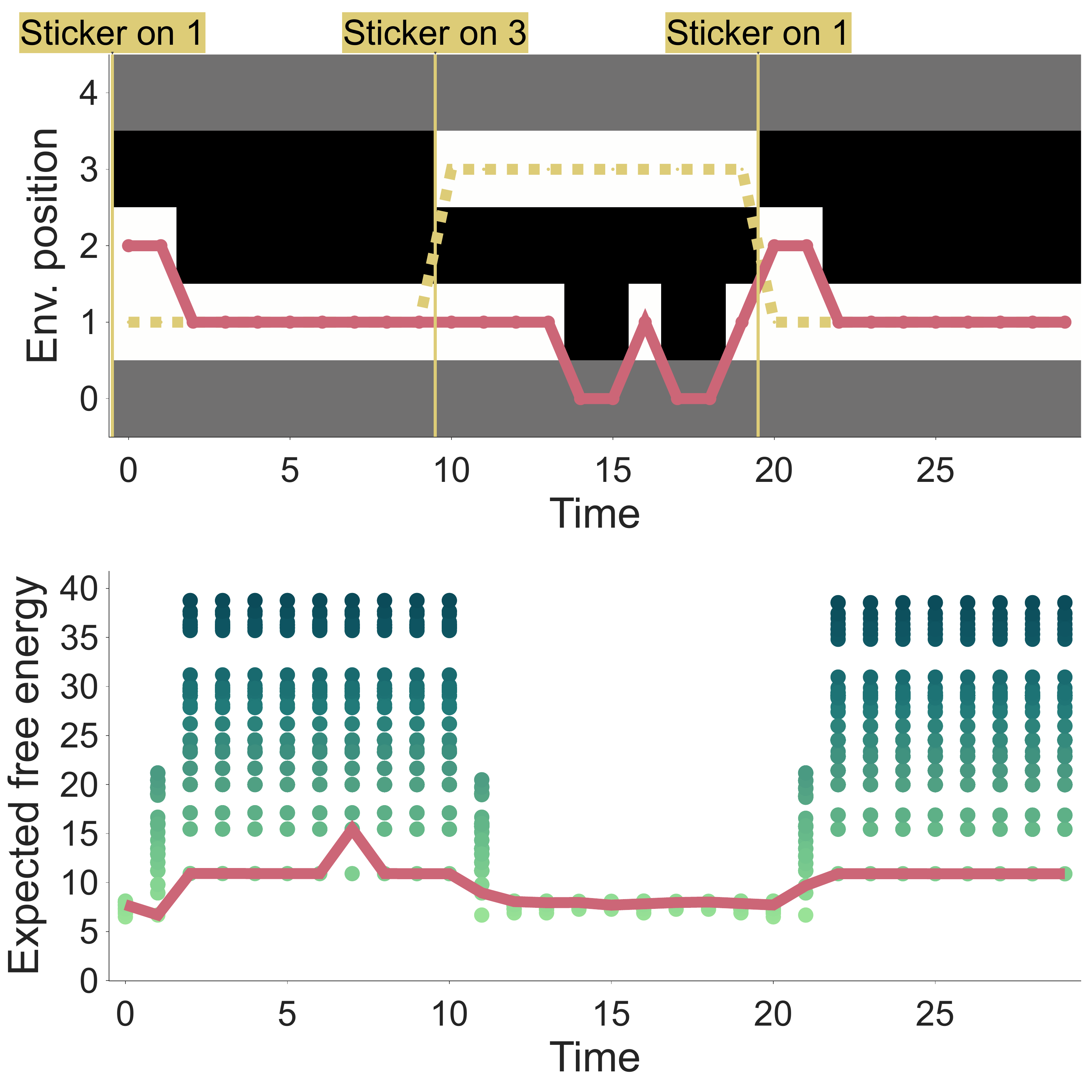}
  \caption{
\ifshowkorean
    {\devsciscriptsize 위치 3에 스티커가 계속 붙어있었던 에이전트의 행동 ($t = 30,000$). 다른 위치에 스티커가 붙으면 여전히 해당 위치로 손을 뻗지만, 위치 3에 스티커를 붙이면, 더 이상 관심을 가지지 않는다.}
\else
%TC:endignore
    Agent's behavior when a sticker has persistently been attached to position~3 ($t = 30{,}000$). When the sticker is placed on other positions, the agent still reaches toward them; however, it no longer shows interest when the sticker is placed at position~3.
%TC:ignore
\fi
  }
  \label{fig:fig6}
\end{figure}

\ifshowkorean
{\devscismall
다음으로, 스티커를 position 3에 붙여 둔 상태에서 에이전트를 다시 20,000 스텝 동안 motor babbling하도록 하였다. 이 과정에서 self-prior는 점차 갱신되며, 스티커가 없는 경우보다 position 3에서 촉각 신호가 발생하는 상태의 확률이 더 높아진다. 즉, 에이전트는 “오른손이 어디에 있더라도 position 3에서는 촉각이 발생한다”는 기대를 형성하게 된다 (Right of Figure~\ref{fig:fig4}).

이 확장된 babbling 뒤에, 스티커가 position 1로 옮겨지면, 여전히 에이전트는 스티커의 위치 변화로 인한 불일치를 감지하고 손을 뻗는다. 이는 position 1에서 촉각 신호가 발생하는 것이 self-prior와 일치하지 않기 때문이다. 반면, 스티커를 다시 position 3으로 옮기면, 에이전트는 이전과 같은 일관된 목표 지향적 행동을 보이지 않는다~(Figure~\ref{fig:fig6}). 이는 self-prior가 이미 position 3에서의 촉각 신호를 예상하도록 적응했기 때문에, 더 이상 불일치가 발생하지 않기 때문이다. 그 결과, 스티커에 접근하는 것과 그렇지 않은 것 사이의 기대 자유 에너지 차이가 미미해지며, 에이전트는 자유 에너지를 줄이는 명확한 방법을 찾지 못하고 랜덤하게 행동한다.
}
\else
%TC:endignore
Next, we let the agent undergo another round of motor babbling (20,000 steps) while the sticker remained at position~3. Over time, the self-prior was gradually updated so that the probability of tactile signal at position~3 increased relative to when the sticker was absent. In other words, the agent came to expect tactile input at position~3 regardless of its hand position~(Right of Figure~\ref{fig:fig4}). 

After this extended babbling, if the sticker was moved to position~1, the agent still noticed the mismatch (as a tactile signal at position~1 is inconsistent with the self-prior) and reached out to touch it. However, if the sticker was placed back at position~3, the agent no longer exhibited consistent goal-directed behavior~(Figure~\ref{fig:fig6}). Because the self-prior had already adapted to position~3, no additional mismatch arose. Consequently, the difference in expected free energy between reaching or not was negligible, so the agent simply behaved randomly.
%TC:ignore
\fi

\ifshowkorean
{\devscismall
self-prior에 의해 동기가 유발되지 않는 이 상황을 통해, 기대 자유 에너지 프레임워크에서 전통적인 정보 이득 요소가 내발적 동기에 미치는 영향을 분리하여 살펴볼 수 있다. 본 실험에서는 파라미터 $\mathbf{A}$ 및 $\mathbf{B}$가 이미 완벽한 상태로 고정되어 있으므로, 관찰을 통해 새롭게 얻을 수 있는 정보가 없다. 그 결과, 정보 탐색적(epistemic) 동기가 발생하지 않으므로, 모든 정책이 거의 동등한 자유 에너지를 갖게 되어, 에이전트는 랜덤한 행동을 하게 된다.
}
\else
%TC:endignore
This situation, in which the self-prior does not induce any motivation, allows us to isolate the influence of the traditional information gain component of intrinsic motivation in the expected free energy framework. In this experiment, since the parameters $\mathbf{A}$ and $\mathbf{B}$ are already fixed in a fully accurate state, there is no new information to be gained from observations. As a result, no epistemic (information-seeking) drive emerges, and all policies yield nearly identical expected free energy, causing the agent to behave randomly.
%TC:ignore
\fi

\ifshowkorean
{\devscismall
대조적으로, 우리는 self-prior에 의해 유발된 행동이 추가적인 정보 이득이 없는 상황에서도 생성될 수 있음을 확인했다. 즉, self-prior의 파라미터 $\mathbf{C}$가 경험을 통해 지속적으로 변화했기 때문에, 경험에 근거한 친숙함과 관련된 내발적 동기가 유도될 수 있었다. 그 결과, 에이전트는 새롭게 갱신된 self-prior와 관찰을 일치시키기 위한 행동을 수행하게 되었다.
}
\else
%TC:endignore
In contrast, we observed that actions arising from the self-prior could be generated even when no additional information gain was possible. In other words, because the parameters of the self-prior, $\mathbf{C}$, continuously evolved with experience, intrinsic motivation related to familiarity based on experience was still induced. As a result, the agent was driven to act in ways that would align the newly updated self-prior with the observations.
%TC:ignore
\fi

\subsection{Simulation in Continuous Environment}
\begin{figure}[t]
  \centering
  \includegraphics[width=0.45\linewidth]{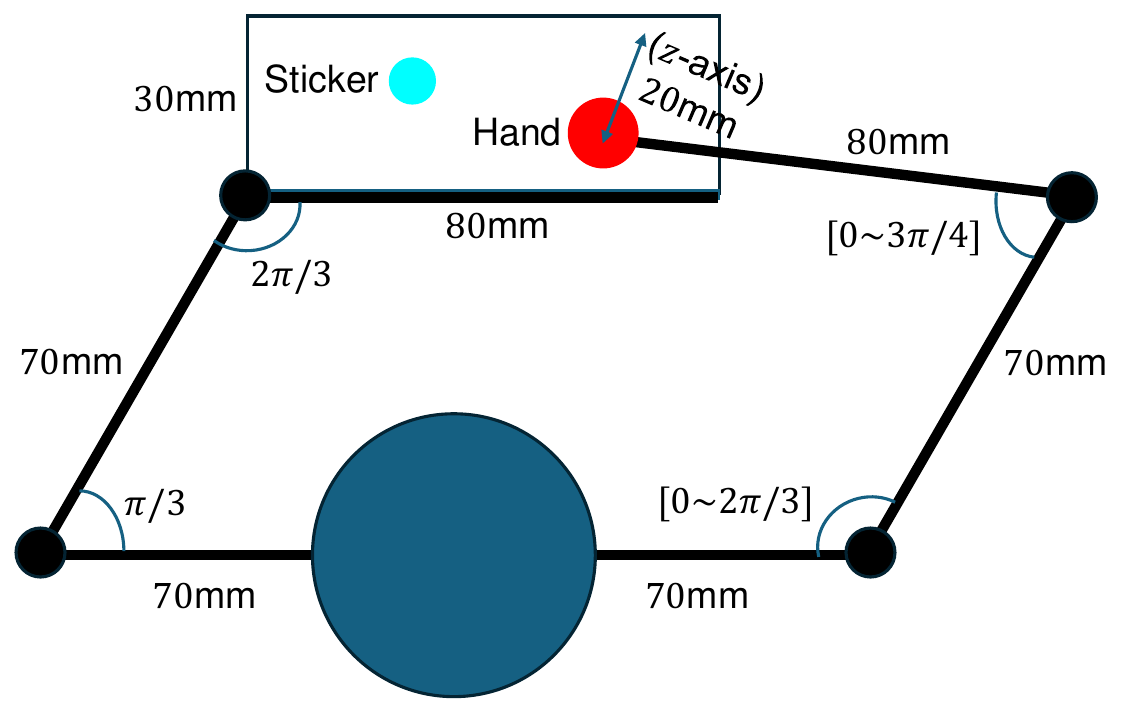}
  \caption{
\ifshowkorean
    {\devsciscriptsize 연속 환경의 개요. 이산 환경에서와 마찬가지로 오른손은 왼팔과 왼팔 주변을 움직일 수 있고, 오른손이 있는 곳이나 스티커가 있는 곳에서 촉각이 발생한다.}
\else
%TC:endignore
    Overview of the continuous environment. As in the discrete environment, the right hand can move over and around the left arm, and tactile sensations are generated where the hand or the sticker is located.
%TC:ignore
\fi
  }
  \label{fig:fig7}
\end{figure}

\ifshowkorean
{\devscismall
지금까지 우리는 간단한 이산 설정을 통해 self-prior에 의해 목표 지향적 행위가 창발할 수 있다는 핵심 아이디어를 소개했다. 이제 이를 \citep{marcel_learning_2022}의 환경 설정에서 영감을 받은 연속 환경으로 확장한다~(Figure~\ref{fig:fig7}).
}
\else
%TC:endignore
Up to this point, we have introduced the core idea that goal-directed behavior can emerge from the self-prior using a simple discrete setup. We now extend this to an environment with continuous-valued variables, inspired by the environment setup of \citep{marcel_learning_2022}~(Figure~\ref{fig:fig7}).
%TC:ignore
\fi

\ifshowkorean
{\devscismall
초기에 에이전트의 양팔은 $xy$-평면에 놓여 있다. Left forearm, left arm, torso, right arm, right forearm의 길이는 각각 $80$, $70$, $140$, $70$, $80\,\text{mm}$이다. Left forearm에는 너비 $30\,\text{mm}$의 2차원 촉각 센서 배열이 있으며, forearm을 기준으로 바깥(lateral) 방향에 위치한다. Right forearm의 끝에는 반지름 $6\,\text{mm}$의 원형 “right hand”가 있으며, 이는 left forearm에 촉각 입력을 생성할 수 있다.

모든 관절 각도는 radian 단위로 표현된다. Left elbow와 left shoulder의 각도는 각각 $\pi/3$ 및 $2\pi/3$으로 고정되어 있으며, right shoulder와 right elbow의 가동 범위는 각각 $[0,\,2\pi/3]$ 및 $[0,\,3\pi/4]$이다. 추가적으로, right hand는 의사적(pseudo) $z$축(높이) 방향으로 $[0,\,20]\,\text{mm}$ 범위에서 이동할 수 있다.
}
\else
%TC:endignore
Initially, the agent's two arms lie on the $xy$-plane. The lengths of the left forearm, left arm, torso, right arm, and right forearm are $80$, $70$, $140$, $70$, and $80\,\text{mm}$, respectively. The left forearm features a $30\,\text{mm}$-wide two-dimensional tactile sensor array, placed laterally relative to the forearm itself. At the endpoint of the right forearm is a circular "right hand" of radius $6\,\text{mm}$, which can generate tactile input on the left forearm.

All joint angles are expressed in radians. The left elbow and left shoulder angles are fixed at $\pi/3$ and $2\pi/3$, while the right shoulder and right elbow angles have respective ranges of $[0,\,2\pi/3]$ and $[0,\,3\pi/4]$. Additionally, the right hand can move along a pseudo-$z$ (height) axis within the interval $[0,\,20]\text{mm}$.
%TC:ignore
\fi

\ifshowkorean
{\devscismall
이산 환경 실험에서와 마찬가지로, 에이전트는 고유 감각과 촉각을 관찰하지만, 연속 환경에서는 이 값들이 모두 실수로 표현된다. 고유 감각 입력은 $(\text{shoulder},\text{elbow},\text{hand})$로 구성된 3차원 실수 벡터이며, 촉각 입력은 $80\times30$ 크기의 실수 행렬로, 값이 $[0,\,1]$ 범위로 정규화되어 있다. Right hand가 left forearm 위에 있고 높이가 $10\,\text{mm}$ 이하일 경우, left forearm의 촉각 센서가 활성화되며, 촉각 강도는 right hand의 높이에 반비례한다.
}
\else
%TC:endignore
Similar to the discrete case, the agent perceives both proprioception and touch; however, these quantities are now continuous. The proprioceptive input is a three-dimensional real-valued vector $(\text{shoulder},\text{elbow},\text{hand})$, while tactile input is represented by an $80\times30$ real-valued matrix normalized to $[0,\,1]$. Whenever the right hand is above the left forearm at a height of $\leq 10\,\text{mm}$, the tactile sensor on the left forearm is activated with intensity inversely proportional to the hand’s height.
%TC:ignore
\fi

\ifshowkorean
{\devscismall
연속 환경을 위한 모델에서 촉각과 고유 감각을 단일한 $o_t$ 표현으로 나타내기 위해, 각 모달리티를 동일한 차원을 갖는 embedding으로 변환한 후 통합한다.
촉각 행렬은 CNN을 사용하여 embedding으로 변환하고, 고유 감각 벡터는 MLP를 사용하여 embedding으로 변환한다.
두 embedding을 요소별 덧셈하여 최종적으로 통합된 관찰 embedding $o_t$를 생성한다.
% rev: 부록 추가
Embedding으로부터 원래의 감각으로 복원하기 위해 역변환 모듈을 사용한다 
\uline{(상세한 하이퍼파라미터는 부록} \ref{subsubsec:embedding_transformation_architecture_continuous}\uline{ 참조).}
}
\else
%TC:endignore
To represent both tactile and proprioceptive modalities as a unified observation $o_t$ in the model for the continuous environment, we convert each modality into an embedding of the same dimensionality and integrate them.
The tactile matrix is transformed into an embedding using a CNN, and the proprioceptive vector is transformed into an embedding using a MLP.
We combine the two embeddings via element-wise addition to yield the final integrated observation embedding $o_t$.
To decode from the embedding back to the original sensations, we use inverse transformation modules 
\rev{(see Appendix} \ref{subsubsec:embedding_transformation_architecture_continuous}\rev{ for detailed hyperparameters).}
%TC:ignore
\fi

\ifshowkorean
{\devscismall
Caregiver는 반지름 $4\,\text{mm}$의 스티커를 left forearm에 부착할 수 있으며, 스티커는 고정된 높이 $0\,\text{mm}$에 위치한다.
% rev: 환경이 간단한 이유 설명
\uline{Self-prior 메커니즘의 핵심을 증명하기 위한 단순화된 환경 규칙으로서,}
에이전트는 right hand의 중심이 스티커 중심으로부터 hand와 sticker 반지름의 합 이하의 거리 내에 $10$ 타임 스텝 동안 연속으로 유지되면 스티커를 제거할 수 있다.
% rev: 환경이 간단한 이유 설명
\uline{이는 잡기(grasping)나 밀기(pushing)와 같은 복잡한 조작 행동을 모델링하지 않고, 손과 스티커의 지속적인 접촉이 물체 제거로 이어지는 단순화된 가정이다.}
}
\else
%TC:endignore
A caregiver can attach a sticker of radius $4\,\text{mm}$ to the left forearm at a fixed height of $0\,\text{mm}$.
\rev{As a simplified environmental rule to demonstrate the core self-prior mechanism,}
the agent can remove the sticker if the center of its right hand remains within the sum of the hand's and sticker's radii for $10$ consecutive time steps.
\rev{This represents a simplified assumption wherein sustained contact between the hand and sticker leads to object removal, without modeling complex manipulation actions such as grasping or pushing.}
%TC:ignore  
\fi

\ifshowkorean
{\devscismall
Right shoulder와 right elbow는 각 시간 단계 $t$마다 $-0.05$에서 $+0.05\,\text{rad}$ 사이로 회전할 수 있으며, right hand는 높이를 $-0.5$에서 $+0.5\,\text{mm}$ 범위 내에서 이동할 수 있다. 추가적으로, right hand의 중심이 left forearm의 표면으로부터 $15\,\text{mm}$ 이내에 유지되도록 하는 제약을 부여하였다. 만약 이동이 이 범위를 벗어나면, 해당 시간 단계에서의 행동은 무시되며 이전 위치가 유지된다.
}
\else
%TC:endignore
At every time step, the right shoulder and right elbow angles can each rotate between $-0.05$ and $+0.05\,\text{rad}$, and the hand’s height can shift by $-0.5$ to $+0.5\,\text{mm}$. To keep the hand near the forearm, we impose an additional constraint that the hand’s center must remain within $15\,\text{mm}$ of the forearm’s surface. If a proposed movement would place the hand outside this region, the agent’s action is disregarded and its previous position is retained.
%TC:ignore
\fi

\ifshowkorean
{\devscismall
연속 환경 실험은 Ubuntu~22.04.5~LTS~64-bit (Linux~5.15.0-1066) 운영 체제에서 Intel Xeon E5-2698~v4 CPU와 NVIDIA Tesla V100-SXM2 GPU를 사용하여 실행하였으며, Python~3.11.10과 PyTorch~2.5.1을 이용하였다. Self-prior의 normalizing flow 기반 모델 학습에는 zuko 라이브러리(version~1.3.0)를 사용하였다. 알고리즘 구현은 Dreamer \citep{hafner_dreamer_2020} 및 Contrastive Active Inference \citep{mazzaglia_contrastive_2021}의 코드를 참고하였다.
}
\else
%TC:endignore
We conducted the continuous environment experiments on a machine running Ubuntu~22.04.5~LTS~64-bit (Linux~5.15.0-1066) with an Intel Xeon E5-2698~v4 CPU and an NVIDIA Tesla V100-SXM2 GPU, using Python~3.11.10 and PyTorch~2.5.1. For our normalizing flows based self-prior, we employed the “zuko” library (version~1.3.0), and we based much of our overall algorithmic implementation on Dreamer \citep{hafner_dreamer_2020} and Contrastive Active Inference \citep{mazzaglia_contrastive_2021}.
%TC:ignore
\fi

\ifshowkorean
{\devscismall
본 실험에서는 심층 모델의 학습 및 평가를 위해 replay buffer에 에피소드를 저장하고, 모델 학습 시 replay buffer에서 무작위로 샘플링하여 사용하였다. 에피소드 수집 시, 다양한 경험을 반영하기 위해 각 에피소드에서 무작위 행동을 사용할지 또는 능동적 추론의 정책을 사용할지를 동일한 확률(50\%)로 결정하였다. 또한, 각 에피소드에서 팔의 랜덤한 위치에 스티커가 붙어 있는 경우와 붙어 있지 않은 경우를 동일한 확률(50\%)로 결정하였다.
}
\else
%TC:endignore
To train and evaluate our deep model, we stored episodes in a replay buffer and randomly sampled them during model training. Each episode was collected either by using random actions or by following the agent's policy from active inference, chosen with equal probability (50\%). Likewise, for half of the episodes, a sticker was placed at a random location on the arm; for the other half, no sticker was used.
%TC:ignore
\fi

\ifshowkorean
{\devscismall
초기 100개의 에피소드를 수집한 후, 각 에피소드가 종료될 때마다 모델을 100 epochs씩 학습하였다. 각 epoch에서는 replay buffer에서 길이 $L=50$인 경로(trajectory) $B=50$개를 샘플링하였으며, 정책 학습을 위해 미래 지평선 $H=15$까지의 상상된 rollouts을 생성하였다. 변분 자유 에너지 최소화를 위한 trajectory 샘플링은 스티커의 존재 여부나 행동이 랜덤인지 정책에 의한 것인지에 관계없이 모든 데이터에서 이루어졌다. 이는 다양한 경험을 통해 올바른 world model을 학습하기 위한 것으로, 이는 이산 환경 실험에서 파라미터 $\mathbf{A}$ 및 $\mathbf{B}$가 이미 학습되어 있다고 가정한 것과 유사하다. 마찬가지로, 기대 자유 에너지 최소화를 위한 학습에서도 모든 데이터를 사용하여, 다양한 상황에서 어떤 행동이 자유 에너지를 최소화하는지 학습할 수 있도록 하였다.
}
\else
%TC:endignore
After collecting an initial batch of 100 episodes, we trained the model for 100 epochs at the end of each episode. In each epoch, we sampled $B=50$ trajectories of length $L=50$ from the replay buffer, and generated imagined rollouts with a planning horizon $H=15$ for policy learning. For variational free energy minimization, trajectories were sampled from all available data, regardless of the presence of a sticker or whether the behavior was random or policy-driven. This was intended to ensure the acquisition of an accurate world model, similar to the assumption in the discrete environment that $\mathbf{A}$ and $\mathbf{B}$ were already known. Likewise, all data were used for expected free energy minimization, so that the agent could learn which actions reduce free energy effectively across various conditions.
%TC:ignore
\fi

\ifshowkorean
{\devscismall
Self-prior 학습에 사용한 데이터는, 스티커가 부착된 에피소드는 약 5\%만 포함하도록 하고, 나머지는 스티커가 붙어 있지 않도록 샘플하여 구성하였다. 따라서 이산 환경 실험과 마찬가지로, self-prior는 “오른손이 왼팔 위에 있을 때 단일한 촉각이 발생”하고, “오른손이 왼팔 밖에 있으면 촉각이 발생하지 않음”이라는 관찰 확률을 높게 학습하게 된다.
}
\else
%TC:endignore
For self-prior learning, we constructed the dataset such that only about 5\% of the episodes included a sticker, while the rest were sampled without any sticker present. Consequently, as in the discrete environment experiment, the self-prior came to assign high probability to observations like “a single-point touch occurs when the right hand is above the left arm” and “no touch occurs when the right hand is away”.
%TC:ignore
\fi

\begin{figure}[t]
  \centering
      \begin{subfigure}[bp]{0.35\textwidth}
        \centering
      \includegraphics[width=0.52\linewidth]{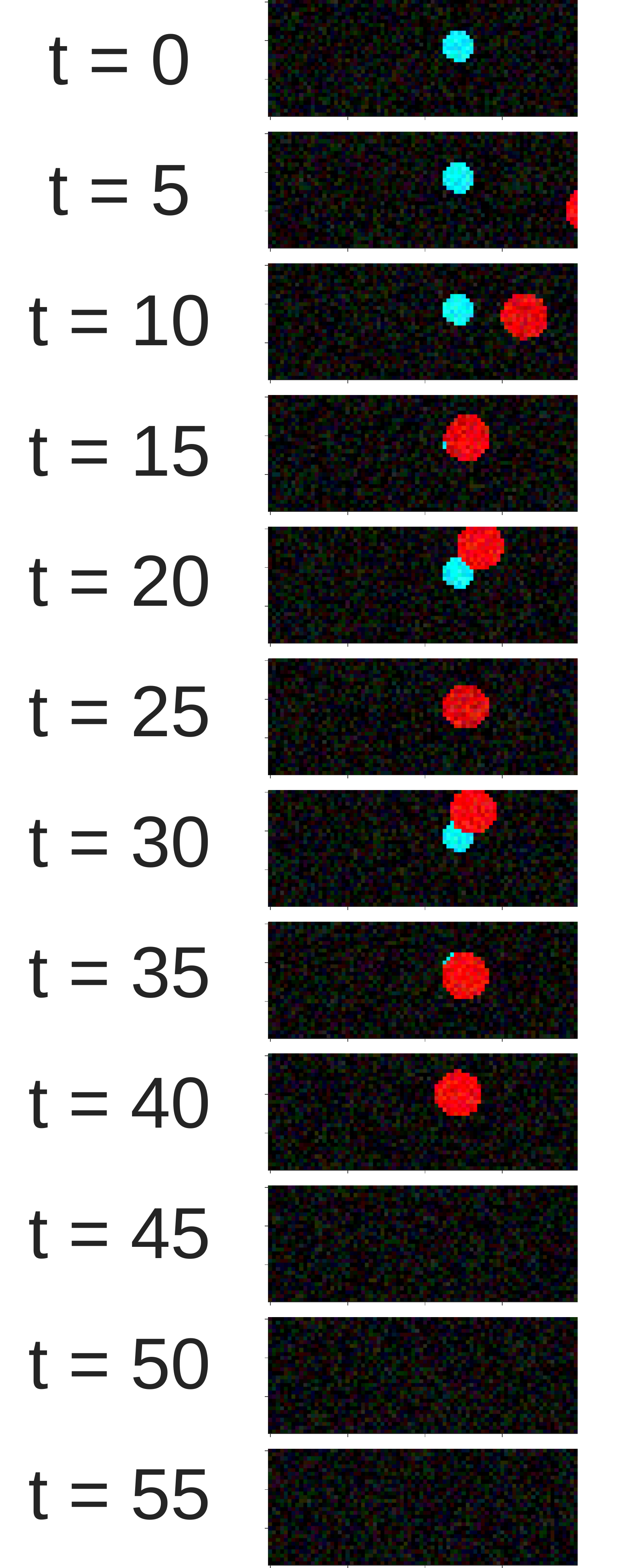}
      \caption{
      \ifshowkorean{
      \devsciscriptsize 리칭 행위의 시계열
      }
      \else
        %TC:endignore
        Time series of reaching behavior
        %TC:ignore
      \fi
      \label{fig:fig8a}
      }
    \end{subfigure}
    \hspace{0.05\textwidth}
    \begin{subfigure}[bp]{0.45\textwidth}
      \includegraphics[width=1.00\linewidth]{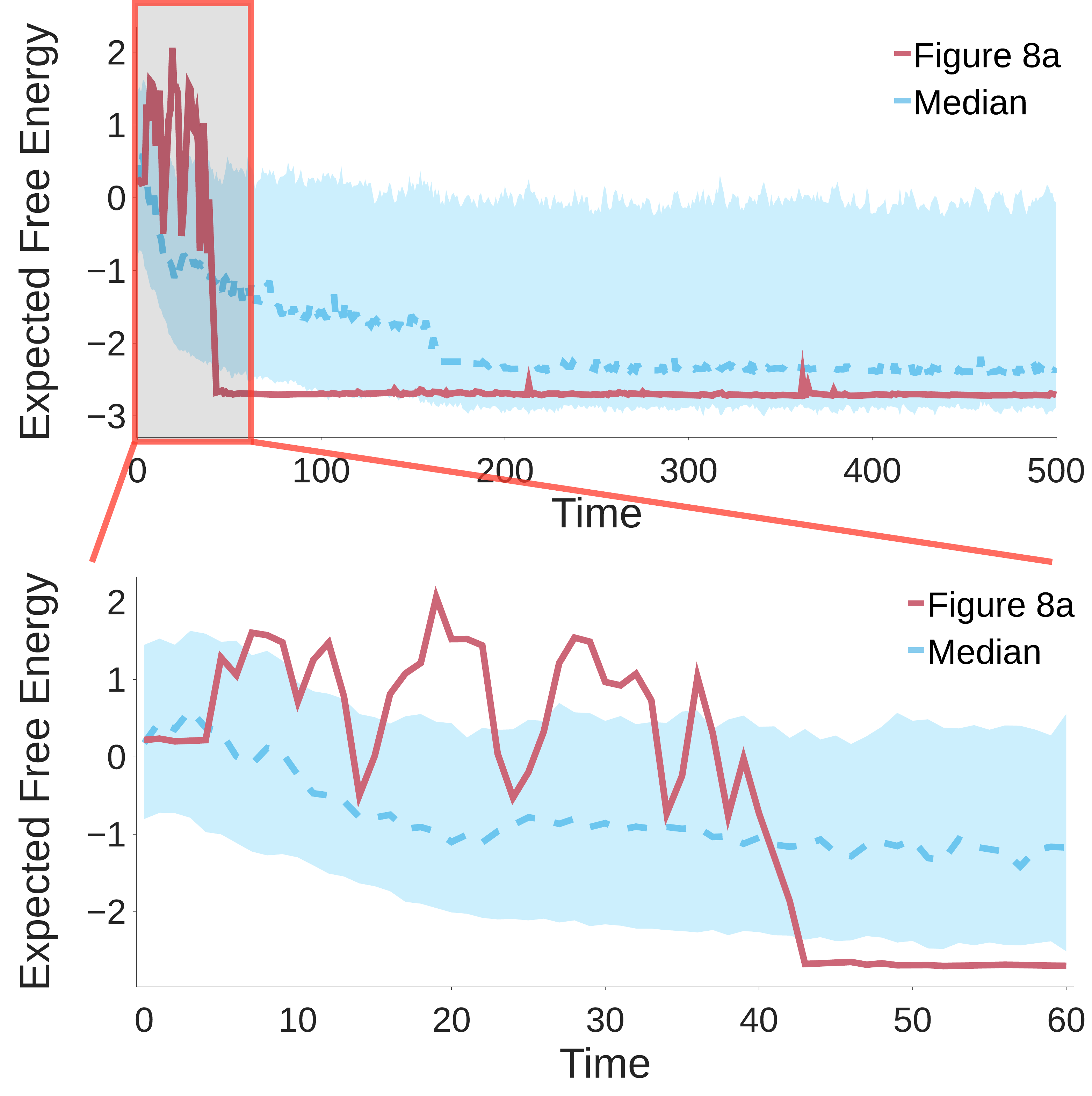}
      \caption{
      \ifshowkorean{
      \devsciscriptsize 기대 자유 에너지의 변화
      }
      \else
        %TC:endignore
        Change in expected free energy
        %TC:ignore
      \fi
      \label{fig:fig8b}
      }
    \end{subfigure}
  \caption{
\ifshowkorean
    {\devsciscriptsize 연속 환경에서 스티커가 붙었을 때의 에이전트의 행동. (a) 파란 원으로 표시된 스티커가 붙으면, 해당 위치로 빨간 원으로 표시된 손이 움직이는 목표 지향적 행동이 그려져 있다. 그림은 에이전트의 촉각 행렬을 가시화 한 것으로, 명확성을 위해 색상이 없는 grayscale 촉각 데이터를 바탕으로 손과 스티커에 해당하는 색상을 입혀서 나타내었다. (b) 스티커에 리칭하는 것으로 기대 자유 에너지를 감소시킬 수 있고, 스티커를 떼어내면 기대 자유 에너지가 최소화된다. 음영 부분은 8개의 model training seed (0\char`~7)에 대해 8개의 environment seed (0\char`~7)를 각각 테스트한 64개의 실험의 표준 편차 영역을 나타낸다.
    % rev: seed 갱신
    이 Figure는 리칭 행동을 빠르고 명확하게 보여줄 수 있는 seed를 골라 작성되었다 \uline{(seed = 4)}.}
\else
%TC:endignore
 Agent behavior when a sticker is placed in the continuous environment. (a) When a sticker (blue circle) is attached, the hand (red circle) moves toward it, illustrating goal-directed reaching behavior. The figure visualizes the agent’s tactile matrix, where grayscale tactile data is overlaid with colored markers for the hand and sticker for clarity. (b) Reaching for the sticker reduces expected free energy, and removing the sticker leads to its minimization. Shaded areas represent the standard deviation across 64 experiments using 8 model training seeds (0\char`~7) tested on 8 environment seeds (0\char`~7). The figure uses \rev{seed 4}, selected for clearly demonstrating the reaching behavior.
%TC:ignore
\fi
  }
  \label{fig:continuous_environment_result}
\end{figure}

\ifshowkorean
{\devscismall
2,000개의 에피소드를 수집하고 학습한 후, 에이전트의 왼팔에 스티커를 붙였다. 그 결과, 에이전트는 스티커를 향해 손을 뻗는 목표 지향적인 행동을 보였다 (Figure~\ref{fig:fig8a}). 이는 스티커를 만지는 것이 self-prior와 일치하여 자유 에너지가 최소화되었기 때문이다 (Figure~\ref{fig:fig8b}). 또한, 에이전트는 스티커에 도달한 후에도 지속적으로 만지면서 결국 스티커를 제거했는데, 이는 단순히 접촉한 후 흥미를 잃는 것이 아니라, 불일치를 줄이려는 지속적인 동기가 있음을 나타낸다.
}
\else
%TC:endignore
After collecting and training for 2,000 episodes, we attached a sticker to the agent's left arm.
As a result, the agent exhibited goal-directed behavior of reaching toward the sticker (Figure~\ref{fig:fig8a}).
This occurred because touching the sticker aligned with the self-prior, minimizing free energy (Figure~\ref{fig:fig8b}).
Moreover, after reaching the sticker, the agent continued touching it and eventually removed it, indicating that it had a persistent motivation to reduce the mismatch rather than simply losing interest after contact.
%TC:ignore
\fi

\ifshowkorean
{\devscismall
흥미롭게도, 에이전트는 초기 상태에서 오른손이 팔을 만지고 있지 않은 경우에도 스티커를 향해 리칭하는 행동을 보였다. 만약 self-prior가 스티커의 존재 여부를 촉각 정보에만 의존하여 학습되었다면, (\textit{i}) 왼팔을 오른손으로 만지는 경우와 (\textit{ii}) 스티커가 붙어 있는 경우는 구별되지 않을 수도 있다. 왜냐하면 두 상황 모두 팔 위에 단일한 촉각 자극이 존재하기 때문이다. 이 경우, 스티커를 향한 리칭 행동이 발생하지 않을 것이다. 그러나 실제로 에이전트는 스티커를 향해 손을 뻗었으며, 이는 self-prior가 촉각과 고유감각을 모두 통합하여 학습되었음을 시사한다. 즉, 멀티모달 감각에 대해 학습된 self-prior를 활용하여 리칭하는 우리의 전체 시스템은, \citep{gallagher_body_1986, hoffmann_body_2021} 등에서 언급된, 행동 계획과 제어를 위한 몸의 표현인 body schema와 흡사하다.
}
\else
%TC:endignore
Interestingly, the agent exhibited reaching behavior toward the sticker even when the right hand was not initially touching the arm. If the self-prior had been learned based solely on tactile information, (\textit{i}) the case where the left arm is touched by the right hand and (\textit{ii}) the case where a sticker is attached might not be distinguishable, since both yield a single tactile point on the arm. In that case, no reaching behavior toward the sticker would emerge. However, the agent did reach toward the sticker, suggesting that the self-prior was learned by integrating both tactile and proprioceptive inputs. In this sense, our system, which utilizes a self-prior learned from multimodal sensory inputs to generate reaching behavior, resembles the concept of a body schema mentioned in \citep{gallagher_body_1986, hoffmann_body_2021}, which supports action planning and control.
%TC:ignore
\fi

\ifshowkorean
{\devscismall
또한, 경로 중간에 오른손에 의해 추가적인 촉각 자극이 발생하여 self-prior의 예상과 일시적으로 불일치하더라도, 에이전트는 최종적으로 자유 에너지를 최소화하기 위해 이러한 단기적 불일치를 감수하며 행동하였다. 이를 통해, 우리의 에이전트가 중간 과정의 노이즈에 영향을 받지 않고, 일관된 목표를 유지하며 행동을 생성했음을 확인할 수 있다.
}
\else
%TC:endignore
Additionally, even if extra touches arose mid-trajectory (\emph{e.g.}, from the right hand itself) thus momentarily increasing mismatch relative to a single-point self-prior, the agent accepted this short-term deviation to achieve the ultimate goal of minimizing overall free energy. This robust, noise-tolerant sequence of goal-directed action demonstrates that our agent maintained a coherent intention throughout its interactions.
%TC:ignore
\fi
%TC:endignore

%TC:ignore
\section{Discussion}
\ifshowkorean
{\devscismall
우리는 에이전트의 감각 경험으로부터 \emph{스스로} 형성되는 고유한 선호인 self-prior를 제안했다. 그리고 우리는 이를 능동적 추론 틀 안에서 시뮬레이션된 아기 에이전트에 적용함으로서, 스티커로의 리칭이라는 목표 지향적인 행위의 창발을 확인했다. 이 장에서는, 우리의 작업이 발달 과학에 어떻게 기여할 수 있는지 명확하게 하기 위해, 내발적 동기에 의한 자발적인 행위의 생성을 시도했던 기존 계산론적 연구들과의 차별점을 설명하고, 초기 의도적 행동의 기원으로서의 self-prior의 역할을 제안하며, 마지막으로 이 논문의 한계점과 향후 연구 방향을 논의한다.
}
\else
%TC:endignore
We proposed the self-prior as an intrinsic preference that is \emph{autonomously} formed from an agent's sensory experiences.
We then applied it to a simulated infant agent within the active inference framework and confirmed the emergence of goal-directed behavior, specifically reaching for a sticker.
In this section, to clarify the contributions of our work to developmental science, we explain how our approach differs from previous studies that attempted to generate spontaneous behaviors, propose the role of the self-prior as a potential origin of early intentional behavior, and finally discuss the limitations of this work and future research directions.
%TC:ignore
\fi

% \subsection{Intrinsic Motivation for Spontaneous Behavior Generation}
% \subsection{Computation Models of Intrinsic Motivation for Spontaneous Behavior}
\subsection{Computation Models of Intrinsic Motivation}

\ifshowkorean
{\devscismall
% external vs [internal] (goal in external)
% [RL] vs FEP
% [extrinsic] vs intrinsic
% heterostatic vs [homeostatic]
% reward analysis vs [action analysis]
내발적 동기를 계산론적으로 다루고자 하는 접근은 주로 강화 학습 분야에서 시도되고 있는데, 이는 고품질의 보상 함수를 설계하는 것이 어렵기 때문이다. 결과적으로, 에이전트가 보상 함수를 더욱 잘 추론하기 위해 환경을 좀 더 효율적으로 탐색하게 하는 내발적 동기와 같은 방법이 제안되어 왔다. 하지만 근본적인 질문은 남아있다: 목표 자체가 주어지지 않은 환경에서 에이전트가 어떻게 스스로 동기를 형성하고 행동을 생성할 수 있는지, 그리고 그런 환경에서 에이전트의 행동을 유도하는 \emph{value}는 무엇인가~\citep{juechems_where_2019}?
}
\else
%TC:endignore
Approaches that aim to computationally address intrinsic motivation have primarily been explored in reinforcement learning, where designing high-quality reward functions remains a significant challenge. As a result, intrinsic motivation has been proposed as a means for agents to more efficiently explore the environment and better infer reward functions. However, a fundamental question persists: how can an agent autonomously form motivation and generate behavior in environments where no goals are given, and what kind of \emph{value} guides the agent’s actions under such conditions~\citep{juechems_where_2019}?
%TC:ignore
\fi

\ifshowkorean
{\devscismall
% external vs [internal]
% [RL] vs FEP
% [extrinsic] vs intrinsic
% heterostatic vs [homeostatic]
% reward analysis vs [action analysis]
\paragraph{외발적 항상성 강화 학습} 한 가지 대안적 접근법은 항상성 강화 학습으로, 환경에서 목표를 부여하지 않더라도 특정 감각 채널을 일정한 기준으로 유지하는 것만으로도 다양한 행동이 생성될 수 있다는 개념이다 \citep{keramati_reinforcement_2011}. 최근 연구에서는 심층 신경망을 활용하여 복잡한 환경에서 내부 생리적 상태에 따라 행동이 어떻게 변화하는지를 조사했다 \citep{yoshida_emergence_2024}. 유사하게, 자유 에너지에 기반한 능동적 추론에서도 에이전트는 내부 기준(선호 사전, preferred prior)에 맞추어 감각 채널을 조정하려고 자유 에너지를 최소화함으로써, 목표를 향한 리칭(reaching) 같은 단일 행동 \citep{oliver_empirical_2022} 또는 반복적인 행동을 포함한 목표 지향적 행위 \citep{matsumoto_goal-directed_2022}를 생성할 수 있다.

이러한 접근법들은 외부 보상이 없이 행동이 발생하므로 내발적인 것처럼 보이지만, 대부분 실험자가 기준값(예: 기준선 또는 선호 사전)을 사전에 \emph{고정}해 두었다는 점에서 한계가 있다. 즉, 에이전트가 보상을 내부적으로 계산한다고 해도, 그 기준 자체는 외부에서 부여된 것이므로 궁극적으로 외발적(extrinsic)인 특성을 갖는다. 따라서 이러한 방식은 "Value가 어디에서 비롯되며, 에이전트가 이를 스스로 설정할 수 있는가?"라는 근본적인 문제를 완전히 해결하지는 못한다.
}
\else
%TC:endignore
\paragraph{Extrinsic Homeostatic Reinforcement Learning} An alternative approach is homeostatic reinforcement learning, in which the agent can generate a range of behaviors simply by maintaining certain sensory channels to specified thresholds, without environmental goal \citep{keramati_reinforcement_2011}. Recent studies have extended this concept to complex settings using deep neural networks, examining how internal physiological states drive behavior \citep{yoshida_emergence_2024}. Under active inference, a similar mechanism emerges: by minimizing free energy with respect to an internal setpoint (expressed as a preferred prior), an agent can generate a single goal-directed act like reaching for a target \citep{oliver_empirical_2022} or orchestrate repeated behaviors \citep{matsumoto_goal-directed_2022}.

However, although these approaches appear intrinsic because behaviors occur without external rewards, most have a limitation in that the experimenter \emph{fixes} the reference values (\emph{e.g.}, setpoints or preferred priors) in advance. That is, even though the agent calculates rewards internally, the reference itself is externally provided, so it ultimately has an extrinsic nature. Therefore, these approaches do not fully resolve the fundamental question of "Where does value come from, and can the agent establish it by itself?"
%TC:ignore
\fi

\ifshowkorean
{\devscismall
% external vs [internal]
% [RL] vs FEP
% extrinsic vs [intrinsic] - new!
% [heterostatic] vs homeostatic - back...
% [reward analysis] vs action analysis - back...
\paragraph{보상 보조적 내발적 동기} 에이전트가 스스로 가치(value) 함수를 설정하도록 만들려면, 외부 보상뿐만 아니라 내부 기준도 사전에 정해지지 않은 상태여야 한다. 그러나 이러한 평가 지표가 사라지면 학습의 진행 상황이나 행동의 유용성을 판단하기 어려워진다. 이로 인해 많은 내발적 동기 연구들은 여전히 외발적 과업을 위한 탐색을 촉진하는 역할로만 분석되는 경향이 있다 \citep{aubret_information-theoretic_2023}. 즉, 에이전트가 스스로 목표를 생성하고 유지하는 과정은 상대적으로 덜 조명되었다.
}
\else
%TC:endignore
\paragraph{Reward-auxiliary Intrinsic Motivation} To allow the agent to establish its own value function independently, neither external rewards nor internal setpoints can be predetermined.
But without these metrics, it is difficult to track learning progress or judge the utility of behaviors, which is why many studies on intrinsic motivation still tend to analyze it primarily as a means to facilitate exploration for extrinsic tasks \citep{aubret_information-theoretic_2023}.
That is, the process by which agents generate and maintain goals themselves has been relatively less illuminated.
%TC:ignore
\fi

\ifshowkorean
{\devscismall
% external vs [internal]
% [RL] vs FEP
% extrinsic vs [intrinsic]
% [heterostatic] vs homeostatic
% reward analysis vs [action analysis] - new!
\paragraph{Heterostatic 내발적 동기} 일부 연구는 내발적 동기에 의해 발생하는 행동 자체를 조사하고, 외부적으로 정의된 보상이 전혀 없는 환경에서 이를 정성적으로 평가하였다 \citep{eysenbach_diayn_2018, pathak_icm_2017}. 이러한 연구들은 에이전트가 다양한 기술이나 행동을 습득하는 과정을 강조하지만, 일반적으로 새로운 상태를 지속적으로 탐색하는 heterostatic 접근 방식을 따른다. 따라서 환경에 대한 불확실성이 충분히 감소하면 추가적인 탐색이 줄어들고, 새로운 행동이 더 이상 발생하지 않는 한계를 가진다.
}
\else
%TC:endignore
\paragraph{Heterostatic Intrinsic Motivation} Some works do investigate the behaviors arising purely from intrinsic motivation and qualitatively evaluate them in scenarios with no externally defined reward \citep{eysenbach_diayn_2018, pathak_icm_2017}. While these highlight the agent’s acquisition of diverse skills or behaviors, they typically involve a heterostatic approach, continuously seeking novel states. Consequently, once uncertainty is sufficiently reduced, further exploration tends to diminish, and fewer new behaviors emerge.
%TC:ignore
\fi

\ifshowkorean
{\devscismall
% external vs [internal]
% RL vs [FEP] - new!
% extrinsic vs [intrinsic]
% [heterostatic] vs homeostatic
% reward analysis vs [action analysis]
능동적 추론은 자유 에너지라는 단일한 정량적 지표를 사용하여 외발적 동기와 내발적 동기를 통합할 수 있는 이론적 프레임워크를 제공한다 \citep{parr_active_2022}. 더욱이, \citep{biehl_expanding_2018}는 기존의 다양한 내발적 동기 메커니즘이 능동적 추론 안에서 사후 확률을 개선하는 과정으로 해석될 수 있음을 보였으며, 이는 \citep{schmidhuber_formal_2010}에서 제안한 "더 나은 세계 모델"을 추구하는 내발적 동기 개념과 유사한 관점이다. 그러나 이러한 사후 확률을 개선하는 내발적 동기는 본질적으로 heterostatic한 특성을 가지며, 학습이 진행되어 불확실성이 해소되면 새로운 행동이 점차 줄어드는 경향이 있다.
}
\else
%TC:endignore
Active inference offers a unified theoretical framework that can integrate both extrinsic and intrinsic motivations under the single quantitative measure of free energy \citep{parr_active_2022}. Furthermore, \citep{biehl_expanding_2018} showed that many existing methods for intrinsic motivation can be cast within active inference as processes for improving the posterior, a perspective reminiscent of \citep{schmidhuber_formal_2010}, who defined intrinsic motivation in terms of seeking “better world models.” However, such posterior-improving motives are likewise heterostatic: they promote exploration until uncertainty is resolved, after which they elicit fewer notable actions.
%TC:ignore
\fi

\ifshowkorean
{\devscismall
% external vs [internal]
% [RL] vs FEP - back..
% extrinsic vs [intrinsic]
% heterostatic vs [homeostatic] - new!
% reward analysis vs [action analysis]
\paragraph{항상성 내발적 동기} 최근에는 에이전트가 스스로 기준점을 정하면서도, 계속해서 새로운 것을 탐색하는 것이 아니라, 이미 학습된 상태나 과거 경험을 유지하려는 homeostatic한 내발적 동기를 다룬 연구들이 등장하기 시작했다.
이러한 연구들은 외부 보상 없이도 목표 지향적 행동이 발생할 수 있음을 보여주었다는 점에서 중요한 기여를 한다.
예를 들어, \citep{marcel_learning_2022}는 self-touch 관찰의 잠재 표현을 기준점으로 사용하여 reaching 행동을 유도하는 과정을 제안했으며, \citep{takemura_neural_2018, kim_amam_2023}은 motor babbling으로부터 학습한 forward 및 inverse 모델을 통해 명확한 보상 없이도 목표 위치로의 reaching을 생성할 수 있음을 보였다.
우리의 접근과 더욱 비슷한 \citep{sajid_exploration_2021}는 자유 에너지 프레임워크 안에서 "학습된 선호" 개념을 도입하고, 관찰과 선호 간의 불일치가 새로운 행동을 유발하는 방식을 의미론적으로 분석했다.
% rev: 한계 언급 대상 한정
\uline{이러한 연구들은 homeostatic 내발적 동기의 가능성을 보여주었으나, 단일한 내발적 동기 메커니즘에 초점을 맞춘 독립적인 프레임워크 내에서 작동하며, 외발적 동기와의 통합은 향후 과제로 남아있다.}
}
\else
%TC:endignore
\paragraph{Homeostatic Intrinsic Motivation} Recently, studies have begun to emerge that address homeostatic intrinsic motivation, in which agents autonomously define their own setpoints but aim to preserve already learned states or past experiences, rather than continuously seeking novelty.
These studies have made important contributions by demonstrating that goal-directed behaviors can emerge without external rewards.
For example, \citep{marcel_learning_2022} proposed a process in which latent representations of self-touch observations are used as setpoints to induce reaching behavior, and \citep{takemura_neural_2018, kim_amam_2023} demonstrated that forward and inverse models learned from motor babbling can generate reaching toward target positions without explicit rewards.
Most similar to our approach, \citep{sajid_exploration_2021} introduced the concept of "learned preferences" within the free energy framework and semantically analyzed how a drive to reduce mismatches between observation and preference can lead to new behaviors.
\rev{While these studies have demonstrated the potential of homeostatic intrinsic motivation, they operate within independent frameworks focused on a single intrinsic motivation mechanism, and integration with extrinsic motivations remains a future challenge.}
%TC:ignore
\fi

\ifshowkorean
{\devscismall
% external vs [internal]
% RL vs [FEP] - new!
% extrinsic vs [intrinsic]
% heterostatic vs [homeostatic]
% reward analysis vs [action analysis]
% rev: 차별점 재정리
\paragraph{정리}
\uline{정리하면, 우리의 접근법은 기존 연구들과 다음과 같은 점에서 차별화된다:}
\begin{enumerate}
    \item \emph{내발적(intrinsic)}: 에이전트가 외부에서 고정된 기준이 아닌 자신의 경험을 바탕으로 스스로 기준을 설정한다.
    \item \emph{독립적 행동 생성}: 탐색을 촉진하는 보조적 역할이 아니라, 리칭과 같은 특정한 목표 지향적 행동을 독립적으로 생성한다.
    \item \emph{homeostatic} 특성: 정보 이득이 없는 상황이라도, 익숙한 상태와의 불일치가 있으면 행동이 멈추지 않고 지속된다.
    \item \emph{자유 에너지 원리 및 능동적 추론}: 능동적 추론 프레임워크 내에서 정의되어 기존의 외발적 동기 및 heterostatic 내발적 동기(정보 이득)와 이론적으로 통합될 수 있다.
\end{enumerate}
}
\else
%TC:endignore
\paragraph{Summary}
\rev{In summary, our approach is differentiated from existing research in the following ways:}
\begin{enumerate}
    \item \textbf{Intrinsic}: The agent autonomously establishes criteria based on its own experience rather than adhering to externally fixed criteria.
    \item \textbf{Independent behavior generation}: It independently generates specific goal-directed behaviors such as reaching, rather than serving an auxiliary role to facilitate exploration.
    \item \textbf{Homeostatic}: Even in situations without information gain, behaviors persist without stopping as long as there is a mismatch with familiar states.
    \item \textbf{FEP with active inference}: Defined within the active inference framework, it can be theoretically integrated with existing extrinsic motivations and heterostatic intrinsic motivations (information gain).
\end{enumerate}
%TC:ignore
\fi

% 내발적 동기 Self-prior의 자리매김

\subsection{Origin of Early Intentional Behaviors}
\ifshowkorean
{\devscismall
유아는 초기에는 반사 기제에 크게 의존하여 행동하지만, 시간이 지남에 따라 점진적으로 “의도적인 행동”을 발달시킨다. 이는 (\textit{i}) 자극을 어떻게 처리할지 판단하는 동기 부여 시스템과 (\textit{ii}) 실제로 목표 지향적 행동을 실행하는 능력으로 구성된다 \citep{mele_intentional_1994}. \citep{zaadnoordijk_origins_2020}에 따르면, 처음에는 stimulus-driven intention이 먼저 나타나고, 이후 경험이 축적됨에 따라 점차 endogenous intention이 발달한다. 예를 들어, 신생아는 Rooting reflex와 같은 반사를 통해 영양을 섭취하는 경험을 쌓고, 이후 젖병을 보면 자연스럽게 손을 뻗어 마시려 한다 (stimulus-driven intention). 시간이 지나 내부 상태인 배고픔을 인식하게 되면, 스스로 젖병을 찾아 마시게 된다 (endogenous intention).
}
\else
%TC:endignore
Infants initially depend heavily on reflex mechanisms for action. Over time, however, they gradually develop what \citep{mele_intentional_1994} describes as “intentional behavior,” comprising (\textit{i}) a motivational system that judges how to handle incoming stimuli and (\textit{ii}) the capability to execute genuinely goal-directed actions. According to \citep{zaadnoordijk_origins_2020}, stimulus-driven intentions arise first, and then, with accumulating experience, more endogenous intentions emerge. For instance, newborns rely on reflexes such as rooting to acquire feeding experience; later, merely seeing a bottle prompts them to reach out and drink (stimulus-driven intention). As they become aware of internal states like hunger, they begin actively searching for the bottle themselves (endogenous intention).
%TC:ignore
\fi

\ifshowkorean
{\devscismall
시간이 지나면서, 유아가 자신을 독립적인 존재로 인식하기 위해서는 멀티 모달 중복성과 시간적·공간적 인과성이 포함된 다양한 경험을 해야 한다 \citep{rochat_self-perception_1998}. 순차적으로 발달하는 의도적인 행동은 이러한 경험을 자발적으로 제공할 가능성이 있다. \citep{thelen_transition_1993} 역시 유아가 초기에는 reaching을 수행할 능력도, 이를 수행하려는 의도도 부족하다고 지적하며, 발달 과정에서 점차적으로 일관된 의도를 형성하고, 이러한 의도에 기반한 능동적 탐색이 reaching 학습을 촉진한다고 하였다.
}
\else
%TC:endignore
Over time, to recognize themselves as independent entities, infants must undergo abundant experiences featuring multimodal redundancy and temporal–spatial contingencies \citep{rochat_self-perception_1998}. Sequentially developing intentional actions may spontaneously provide opportunities for such experiences. \citep{thelen_transition_1993} similarly noted that infants initially lack both the ability and the intention to perform reaching, and that the formation of stable intentions, coupled with active exploration driven by those intentions, facilitates the learning of reaching.
%TC:ignore
\fi

\ifshowkorean
{\devscismall
그러나 기존 연구들은 stimulus-driven intention을 포함한 intention과 intentional한 행동이 어떤 구조에 의해 발달하는지에 대한 논의를 다루지 않았다.
% rev: 기존 능동적 추론 의도 논문 언급
\uline{능동적 추론 프레임워크 내에서 전환 가능한 의도를 구현한 }\citep{priorelli_flexible_2023}\uline{도, 의도들이 스스로 발달하는 것이 아니라, 이산적으로 미리 정의되어 있었다.}
한편, 신경생리학적 연구에 따르면, 전대상피질은 개별 선택에서의 보상 관련 오류를 처리하는 반면, 후대상피질은 자기 신체나 과거 경험과 같은 자서전적 정보와 관련된 불일치를 감지하고 새로운 행동을 유발하는 역할을 한다 \citep{pearson_posterior_2011, brewer_what_2013}.
}
\else
%TC:endignore
However, prior studies have not discussed how intentions, including stimulus-driven intentions, and intentional behaviors develop through what structure.
\rev{Even }\citep{priorelli_flexible_2023}\rev{, which implemented switchable intentions within the active inference framework, used intentions that were discretely predefined rather than self-developing.}
Meanwhile, neurophysiological evidence suggests that the anterior cingulate cortex primarily deals with reward-related errors in individual decisions, whereas the posterior cingulate cortex processes discrepancy signals related to autobiographical information such as one's own body or past experiences, and triggers new behaviors \citep{pearson_posterior_2011, brewer_what_2013}.
%TC:ignore
\fi

\ifshowkorean
{\devscismall
본 연구에서 제안한 self-prior는 이러한 “자기와 관련된 예측”과 부합하며, 이 예측을 충족하지 못하는 새로운 입력이 들어올 때 행동이 유발된다는 점에서 후대상피질과의 연관성을 시사한다. 즉, 특정한 목표가 외부에서 설정되지 않더라도, 과거 경험을 통해 형성된 self-prior를 유지하려는 내발적 동기가 작용하여, 유아가 새롭게 발견한 스티커를 지속적으로 만지는 행동과 유사한 양상이 나타날 수 있다. 이는 Piaget의 1차 순환 반응 개념과 유사하게 반복적인 시도를 통해 점차 정교한 의도가 발달하는 과정과 연결되며, 유아의 의도적 행동의 기원을 계산적으로 탐구하는 이론적 토대를 제공할 수 있다.
}
\else
%TC:endignore
Our proposed self-prior aligns with this notion of “self-relevant predictions,” indicating that when new input fails to satisfy these predictions, the system initiates action. In other words, even without any externally designed goal, an intrinsic drive to preserve a self-prior learned from past experiences can manifest behaviors akin to an infant persistently touching a newly discovered sticker on their arm. This mirrors Piaget’s concept of primary circular reactions, where repeated attempts can refine intentions further, offering a theoretical foundation for computational studies on the origins of infant intentional behavior.
%TC:ignore
\fi

\ifshowkorean
{\devscismall 
본 연구에서는 self-prior의 학습률을 고정한 상태로 실험했지만, 학습률이 지나치게 크면 사소한 변화에도 과도하게 반응하는 반면(강박적인 반응), 반대로 지나치게 작으면 새로운 자극에 무관심해지는(무감동) 극단적인 사례도 고려할 수 있다. 향후 연구에서는 보다 유연한 학습률을 모사하여, 자기 관련 예측 오류 처리가 다양한 행동 양식을 어떻게 형성하는지 탐구할 수 있을 것이다.
}
\else
%TC:endignore
In our experiments, we fixed the self-prior’s learning rate. However, an excessively high rate could lead to overreactions (compulsive-like responses) to minor changes, whereas an overly low rate might cause indifference (apathy) toward new stimuli. Future work could investigate how more flexible learning rates shape different behavioral styles via self-relevant prediction-error processing.
%TC:ignore
\fi

\subsection{Limitations and Future Directions}
\ifshowkorean
{\devscismall
% \paragraph{단순한 실험 설정}
% rev: self-prior의 확장 가능성
\uline{본 연구는 구성론적(constructivist) 접근을 통해, 최소한의 조건에서 self-prior 메커니즘이 어떻게 작동하는지 증명하는 데 초점을 맞추었다. 유아의 스티커 리칭 과제는 이러한 메커니즘을 명확히 보여주기 위한 하나의 예시이며, 촉각 및 고유감각에 기반한 신체 리칭이라는 특정 조건에서 어떤 행동이 창발하는지를 보였다. Self-prior는 원리적으로 모달리티 독립적이므로, 동일한 메커니즘이 다양한 감각 모달리티에 적용될 수 있다. 예를 들어, 시각 모달리티를 사용하면 에이전트의 신체 밖에 있는 드물게 본 외부 물체에 대한 리칭 행동이 나타날 것으로 예상된다.}
}
\else
%TC:endignore
\rev{Our study employs a constructivist approach, focusing on demonstrating how the self-prior mechanism operates under minimal conditions.
The infant sticker-reaching task serves as one example to clearly illustrate this mechanism, showing what behaviors emerge under the specific condition of tactile and proprioceptive body reaching.
Since the self-prior is modality-independent in principle, the same mechanism can be applied to various sensory modalities.
For example, using visual modality would be expected to produce reaching behavior toward rarely seen external objects outside the agent's body.}
%TC:ignore
\fi
\ifshowkorean
{
% rev: self-prior 적용 가능성 연구들 추가 언급
\uline{실제로 영아 발달 과정에서 관찰되는 여러 현상들 역시 self-prior 메커니즘으로 설명할 수 있는 가능성이 있다. 예를 들어, }\citep{eskandari_effect_2020}\uline{에 의하면 네스트로 감싸준 미숙아들은 불안정한 신전 움직임이 현저히 줄어들고 안정된 구부린 자세를 유지했는데, 이는 태아 시절 경험한 구부린 자세와 촉각적 경계 감각을 학습한 self-prior와의 불일치를 해소하기 위해, 미숙아가 자궁 유사 환경을 선호하고 안정된 행동 상태를 유지하는 것으로 재해석할 수 있다.}

\uline{또한, 단순한 감각-운동 수준을 넘어 주의나 사회적 상호작용과 같은 더 복합적인 인지적 맥락에서 나타나는 행동도 self-prior로 설명할 수 있는 가능성이 있다. }\citep{warneken_altruistic_2006}\uline{는 성인이 목표를 달성하는 데 어려움을 겪는 상황에서 영아들이 도움을 주는 task들을 제시했는데, 이 중에는 성인이 잡지(magazine)를 서랍장(closet)에 넣는 것에 어려워하면 영아가 대신 문을 열어주는 행동이 관측되었다는 결과를 포함한다. 이를 self-prior 방식에 맞게 재해석하면, 평소에 성인이 서랍장의 문을 잘 열고 잡지를 넣던 시각 정보를 학습한 self-prior와의 불일치를 해소하기 위해, 영아가 대신 문을 열어줬을 가능성이 있다. 이처럼 경험했던 관측들에 대한 확률밀도를 학습한다는 동일한 원리가 다양한 맥락에서의 행동을 설명할 수 있다.}
}
\else
%TC:endignore
\rev{Indeed, several phenomena observed in actual infant development may also be explained by the self-prior mechanism.
For instance, according to }\citep{eskandari_effect_2020}\rev{, preterm infants nested in boundaries showed a significant reduction in unstable extension movements and maintained stable flexed postures, which can be reinterpreted as the infants preferring uterine-like environments and maintaining stable behavioral states to resolve mismatches with a self-prior learned from the flexed posture and tactile boundary sensations experienced during fetal stages.}

\rev{Furthermore, behaviors that emerge in more complex cognitive contexts beyond simple sensorimotor-level adjustments, such as attention or social interactions, may also be explained by the self-prior.
}\citep{warneken_altruistic_2006}\rev{ presented tasks in which infants help adults who are having difficulty achieving goals, including results where infants opened a door for an adult who was having difficulty putting a magazine into a closet.
Reinterpreting this in terms of the self-prior approach, it is possible that the infant opened the door to resolve a mismatch with a self-prior learned from visual information that adults normally open the closet door and put magazines in.
Thus, the same principle of learning the probability density over experienced observations can explain behaviors across diverse contexts.}
%TC:ignore
\fi
\ifshowkorean
{
우리의 접근법은 이산 환경과 낮은 자유도(3-DoF)의 연속 환경에서만 검증되었는데, 실제 유아나 로봇의 감각운동 상호작용은 훨씬 더 복잡하고 다양하므로, 본 연구를 고차원적인 상태 및 행동 공간으로 확장할 필요가 있다. 특히, 태아기를 포함한 초기 인간의 감각운동 경험의 시뮬레이션을 활용하여~\citep{kim_simulating_2022}, 태내에서 제공된 풍부한 감각운동 입력이 self-prior와 목표 지향적 행동을 어떻게 형성하는지 탐구함으로써, 신체적 제약 속에서 어떤 행동이 발달할 수 있는지 밝히는 연구가 가능할 것이다~\citep{kuniyoshi_fusing_2019}.
}
\else
%TC:endignore
Our approach has only been validated in discrete environments and low-degree-of-freedom (3-DoF) continuous environments, but real infants or robots exhibit far more complex and varied sensorimotor interactions.
Therefore, it is necessary to extend this study to high-dimensional state and action environments.
In particular, by utilizing simulations of early human sensorimotor experiences including fetal stages \citep{kim_simulating_2022}, it would be possible to explore how rich sensorimotor inputs provided in utero shape the self-prior and goal-directed behaviors, thereby revealing what behaviors can develop under physical constraints \citep{kuniyoshi_fusing_2019}.
%TC:ignore
\fi

\ifshowkorean
{
%rev: 장기 시계열 말고도 일반화도 거대 모델로 가능할 거 같다고 언급
\uline{이렇게 복잡한 환경 속에서, 다양한 맥락에서 습득한 prior의 재사용과 같은 일반화를 달성하기 위해서는, 사용 가능한 모든 모달리티를 통합 활용하고, 장시간에 걸친 다양한 장면을 학습하며, 이 모두를 확률적으로 모델화할 수 있는 고차원의 self-prior 모델이 필요하다. 이는 계층적 능동적 추론 }\citep{friston_graphical_2017}\uline{, 또는 긴 시계열을 다루는 Transformer 같은 심층 생성 모델 }\citep{vaswani_transformer_2017, chen_transdreamer_2022})\uline{으로 달성할 수 있을 가능성이 있다. Self-prior는 사전 지식 없이 경험한 감각을 밀도 모델에 학습시키는 방식이므로, 생성 모델의 스케일 업을 통해 이러한 확장이 원리적으로 가능하다.}

\uline{예를 들어, 촉각으로 학습한 리칭 행동을 시각 자극에 대한 리칭으로 전이하거나, 특정 모양의 스티커에 대한 반응을 다른 모양으로 일반화하거나, 자신의 신체에 부착된 물체에 대한 행동을 외부 물체로 확장하는 것과 같은 일반화는 모두 다양한 감각 경험 간의 통계적 규칙성을 학습할 수 있는 대규모 생성 모델을 통해 가능할 것으로 예상된다. 이러한 일반화는 단순히 동일한 행동의 반복이 아니라, 서로 다른 맥락에서 형성된 self-prior들 간의 구조적 유사성을 인식하고 활용하는 과정으로 이해될 수 있다.}
}
\else
%TC:endignore
\rev{In such complex environments, achieving generalizations such as reusing priors acquired across diverse contexts would require integrating all available modalities, learning various scenes over extended periods, and employing high-dimensional self-prior models capable of probabilistically modeling all of these.
Such models could potentially be achieved through hierarchical active inference~}\citep{friston_graphical_2017}\rev{, or deep generative models like Transformers for handling long time series~}\citep{vaswani_transformer_2017, chen_transdreamer_2022}\rev{.
Since the self-prior learns experienced sensations in a density model without prior knowledge, such extensions are possible in principle through scaling up generative models.}

\rev{For example, generalizations such as transferring reaching behavior learned through touch to visual stimuli, generalizing responses to a specific shape of sticker to other shapes, or extending behaviors toward objects attached to one's own body to external objects are all expected to be possible through large-scale generative models capable of learning statistical regularities across diverse sensory experiences.
Such generalization can be understood not simply as repetition of the same action, but as a process of recognizing and utilizing structural similarities among self-priors formed in different contexts.}
%TC:ignore
\fi
\ifshowkorean
{
또한 우리는 외부에서 부여된 보상 기준을 배제함으로써 친숙함에 기반한 내발적 동기에 의해서만 유발된 행위를 보였지만, 실제 유아나 로봇은 에너지 항상성과 새로운 정보 탐색과 같은 다양한 동기를 균형 있게 유지해야 한다. 따라서, 이러한 동기들이 대규모 작업에서도 서로 보완하거나 충돌하면서 행동을 조직하는 방식을 체계적으로 평가하는 실험을 설계하는 것이 향후 연구에서 중요한 과제가 될 것이다.
}
\else
%TC:endignore
Additionally, we showed behaviors driven solely by familiarity based intrinsic motivation by excluding externally imposed reward criteria, but in reality, infants or robots must balance multiple motivations such as energy homeostasis and novel information exploration.
Therefore, designing experiments that systematically assess how these motivations complement or conflict and organize behavior in large-scale tasks will be an important task for future research.
%TC:ignore
\fi

\ifshowkorean
{\devscismall
% \paragraph{self-prior의 학습}
한편 우리의 연구는, 기대 자유 에너지를 사용하는 기존의 연구에서는 주로 고정되어 제공되던 preferred prior를 사용하는 extrinsic 항으로부터, 스스로 학습 가능한 self-prior를 사용하는 intrinsic homeostatic 항을 분리해 냄으로서, 미리 지정된 목표를 달성하기 위한 행동뿐만이 아니라, 스스로 목표 지향적인 행동을 창발할 수 있음을 보였다. 다만, 현 단계에서는 self-prior의 \emph{사용}은 기대 자유 에너지에 의한 정책 생성에 통합되어 있지만, self-prior의 \emph{학습}은 변분 자유 에너지의 학습과는 별도로 수행된다는 한계가 존재한다.
}
\else
%TC:endignore
Meanwhile, our study demonstrated that, by separating an intrinsic homeostatic term using a learnable self-prior from the extrinsic term using a fixed preferred prior that was commonly provided in prior studies using expected free energy, an agent can not only generate actions aimed at achieving predefined goals but also spontaneously generate goal-directed behaviors.
However, a current limitation is that while the \emph{use} of the self-prior is integrated into policy generation via expected free energy, the \emph{learning} of the self-prior is performed separately from the learning of variational free energy.
%TC:ignore
\fi

\ifshowkorean
{\devscismall
이러한 분리의 문제를 해결하기 위한 실마리 또한 계층적인 구조에 있을 수 있다. 즉, 장기적인 경험을 반영하는 상위 계층의 숨겨진 상태가 점진적으로 학습되며, 이로부터 추정된 가능도 분포가 하위 계층의 self-prior로 사용될 수 있다. 이는 기존의 계층적 추론에서 상위 계층이 단순히 느린 시간 척도에서 prior를 학습하고 행동을 결정하여 하위 계층에 영향을 미치는 것과 달리 (예를 들면, $P(s_t|s_{t-1},a_{t-1}) = \text{Cat}(\mathbf{B}_{a_{t-1}})$), 지속적으로 하위 계층으로부터 정보를 받아 prior를 학습하여 장시간에 걸친 self-prior 형성만을 수행한다는 점에서 다르다 (예를 들면, $P(s) = \text{Cat}(\mathbf{B})$).
% rev: 계층적 접근을 통한 내부 상태의 선호 설명
\uline{이러한 계층적 접근은 또한 내부 상태에 대한 선호를 구축하는 것과도 밀접하게 연결될 것으로 예상되는데, 현재 구현에서는 관측 공간에서 self-prior를 학습하지만, 계층적 접근을 통해 상위 계층의 내부 잠재 표현이 경험을 통해 형성될 가능성이 있다.}
}
\else
%TC:endignore
One potential clue to resolving this separation may also lie in a hierarchical architecture.
Specifically, a higher-level hidden state that reflects long-term experiences may be gradually learned, and the likelihood distribution inferred from it could serve as the self-prior for the lower level.
This differs from conventional hierarchical inference, where the higher level simply learns priors and determines actions on a slower timescale to influence the lower level (e.g., $P(s_t|s_{t-1},a_{t-1}) = \text{Cat}(\mathbf{B}_{a_{t-1}})$).
Instead, it continuously receives information from the lower level to learn a prior, serving only to form a self-prior over a long duration (e.g., $P(s) = \text{Cat}(\mathbf{B})$).
This hierarchical approach is also expected to be closely connected to building preferences over internal states.
\rev{While the current implementation learns the self-prior in observation space, through a hierarchical approach, there is a possibility that higher-level internal latent representations may be formed through experience.}
%TC:ignore
\fi

\ifshowkorean
{\devscismall
또 다른 가능성으로는, self-prior의 학습이 피질이 아닌, 해마와 같은 별도의 기억 시스템에 의해 담당되므로, 이로 인해 추론 시스템에 대한 가설인 자유 에너지 원리와는 구조적으로 분리되어야 할 필요성이 있다는 점이다. 이러한 열린 가설들에 대한 구체적인 검증은, 향후의 과제로 남겨둔다.
}
\else
%TC:endignore
Another possibility is that the learning of the self-prior is handled not by the cortex but by a separate memory system such as the hippocampus, and thus may require structural separation from the inference system proposed under the free energy principle. Concrete validation of these open hypotheses is left for future work.
%TC:ignore
\fi

\ifshowkorean
{\devscismall
% \paragraph{연속 환경에서 고정된 익숙한 관찰}
마지막으로, 이산 환경에서는, 스티커가 붙은 상태로 시간이 흐르면 self-prior가 점진적으로 변화하고, 이에 따라 행동도 변화하는 현상을 확인할 수 있었다. 반면, 연속 환경에서는 self-prior의 학습에 사용된 관찰에서 스티커가 포함될 확률을 5\%로 고정하여 실험을 진행했다. 이는 self-prior 모델이 과거 데이터를 replay buffer에 저장하고, 시간적 순서를 고려하지 않은 채 학습되었기 때문에, 관찰 빈도의 변화에 따른 self-prior의 점진적 변화를 조사하기 어려웠기 때문이다. 실제 유아나 로봇에서 stimulus-driven intention이 점차 내재적인 의도로 강화되는 과정을 재현하려면, 시간적 순서를 고려한 self-prior 학습 메커니즘이 필요하다. 이는 앞서 언급한 계층적 구조에서 상위 계층이 장기적 경험을 통합하는 방식으로 구현될 수 있을 것이다.
}
\else
%TC:endignore
Finally, in the discrete environment, we could observe that as time passed with a sticker attached, the self-prior gradually changed and, accordingly, behavior also changed.
By contrast, in the continuous environment, we conducted the experiment by fixing the probability of stickers being included in the observations used for self-prior learning at 5\%.
This was because the self-prior model stored past data in a replay buffer and learned without considering temporal order, making it difficult to investigate gradual changes in the self-prior according to changes in observation frequency.
To reproduce the process by which stimulus-driven intention gradually strengthens into a more endogenous form of intention in real infants or robots, a self-prior learning mechanism that considers temporal order is necessary.
This could potentially be implemented through the hierarchical architecture mentioned earlier, where higher levels integrate long-term experiences.
%TC:ignore
\fi
%TC:endignore

%TC:ignore
\section{Conclusion}
\ifshowkorean
{\devscismall
본 연구는 자유 에너지 원리와 능동적 추론 틀을 바탕으로, 에이전트가 body schema와 유사한 역할을 하는 self-prior를 학습하고 유지하려는 내발적 동기가, 외부에서 설정된 보상 없이도 목표 지향적 행동을 유도할 수 있음을 제시했다. 특히, 기존 능동적 추론 연구에서는 새로운 정보를 탐색하려는 heterostatic한 내발적 동기에 주로 초점을 맞춘 반면, 우리는 익숙한 감각 경험을 유지하려는 homeostatic한 내발적 동기의 중요성을 강조했다.
}
\else
%TC:endignore
Building on the free energy principle and the active inference framework, we have demonstrated that an agent's intrinsic drive to learn and maintain a "self-prior", which is akin to a body schema, can induce goal-directed behaviors even in the absence of externally specified rewards.
In particular, whereas the conventional active inference literature has emphasized heterostatic intrinsic motivation (i.e., seeking new information), our work highlights a homeostatic form of intrinsic motivation, one that actively strives to preserve familiar sensory experiences.
%TC:ignore
\fi

\ifshowkorean
{\devscismall
우리는 유아가 자신의 팔 위의 스티커를 만지고 확인하는 행동을 시뮬레이션 예시로 들어, 자기 모델과 입력 간의 불일치를 해소하려는 내부 동기가, 리칭이나 스티커 제거와 같은 행동으로 어떻게 나타날 수 있는지를 구체적으로 보였다. 이는 발달심리학에서 논의되는 stimulus-driven intentional 행동을 계산론적으로 해석할 수 있는 가능성을 제시하며, 나아가 자서전적 기억과 관련된 오류를 처리하는 후대상피질과의 연관성을 시사한다.
}
\else
%TC:endignore
Using a simulated infant touching and examining a sticker on its arm as an illustrative example, we showed how an internally derived drive to resolve mismatches between the self model and incoming data can manifest in behaviors such as reaching and sticker removal, all without any explicit reward criteria. This offers a computational interpretation of stimulus-driven intentional actions in developmental psychology and suggests potential relevance to the neurobiological literature on posterior cingulate cortex, which processes errors linked to autobiographical information.
%TC:ignore
\fi

\ifshowkorean
{\devscismall
향후 연구에서는 유아 전체 시뮬레이션이나 고자유도 로봇과 같은 복잡한 물리 환경으로 확장하여, 장기적인 의도 형성과 행동의 세분화 과정을 연구할 필요가 있다. 또한 친숙함 기반 동기뿐만 아니라 외발적 동기와 정보 탐색 동기를 함께 통합하여, 여러 동기가 단일한 자유 에너지 프레임워크 내에서 어떻게 상호작용하는지를 실험적으로 검증해야 한다. 이러한 연구는 의도적 행동의 자발적인 발달 궤적을 분석하는 데 중요한 역할을 할 것이다.
}
\else
%TC:endignore
Future work should extend our approach to complex physical environments such as full-scale infant simulations or high-DoF robots to investigate the long-term formation and refinement of intentions and behaviors.
Additionally, we must experimentally validate how multiple motivations interact within a single free energy framework, incorporating not only familiarity based motivation but also extrinsic and information-seeking motivations.
Such research will play an important role in analyzing the spontaneous developmental trajectory of intentional behavior.
%TC:ignore
\fi
%TC:endignore
%TC:ignore

\ifdevscianonymise
\else
    \acksection
    This work was supported by JST, PRESTO Grant Number JPMJPR23S4, Japan.
\fi

%%%%%%%%%%%%%%%%%%%%%%%%%%%%%%%%%%%%%%%%%%%%%%%%%%%%%%%%%%%%

% mycomment: for Biblatex
% \printbibliography

% mycomment: for bibtex
% \bibliography{main.bib}

% mycomment: for natbib
\bibliography{main.bib}

\begin{thebibliography}{}

\bibitem[Aubret et~al., 2023]{aubret_information-theoretic_2023}
Aubret, A., Matignon, L., \& Hassas, S. (2023).
\newblock An {Information}-{Theoretic} {Perspective} on {Intrinsic} {Motivation} in {Reinforcement} {Learning}: {A} {Survey}.
\newblock {\em Entropy}, 25(2), 327.

\bibitem[Biehl et~al., 2018]{biehl_expanding_2018}
Biehl, M., Guckelsberger, C., Salge, C., Smith, S.~C., \& Polani, D. (2018).
\newblock Expanding the {Active} {Inference} {Landscape}: {More} {Intrinsic} {Motivations} in the {Perception}-{Action} {Loop}.
\newblock {\em Frontiers in neurorobotics}.

\bibitem[Bigelow, 1986]{bigelow_development_1986}
Bigelow, A.~E. (1986).
\newblock The development of reaching in blind children.
\newblock {\em British Journal of Developmental Psychology}, 4(4), 355--366.

\bibitem[Brewer et~al., 2013]{brewer_what_2013}
Brewer, J., Garrison, K., \& Whitfield-Gabrieli, S. (2013).
\newblock What about the “{Self}” is {Processed} in the {Posterior} {Cingulate} {Cortex}?
\newblock {\em Frontiers in Human Neuroscience}, 7.

\bibitem[Chen et~al., 2022]{chen_transdreamer_2022}
Chen, C., Wu, Y.-F., Yoon, J., \& Ahn, S. (2022).
\newblock {Reinforcement} {Learning} with {Transformer} {World} {Models}.
\newblock {\em arXiv:2202.09481}.

\bibitem[Chung et~al., 2014]{chung_empirical_2014}
Chung, J., Gulcehre, C., Cho, K., \& Bengio, Y. (2014).
\newblock Empirical {Evaluation} of {Gated} {Recurrent} {Neural} {Networks} on {Sequence} {Modeling}.
\newblock {\em arXiv:1912.01603}.

\bibitem[Czikszentmihalyi, 1990]{czikszentmihalyi_flow_1990}
Czikszentmihalyi, M. (1990).
\newblock {\em Flow: {The} psychology of optimal experience}.
\newblock New York: Harper \& Row.

\bibitem[Di~Domenico \& Ryan, 2017]{di_domenico_emerging_2017}
Di~Domenico, S.~I. \& Ryan, R.~M. (2017).
\newblock The {Emerging} {Neuroscience} of {Intrinsic} {Motivation}: {A} {New} {Frontier} in {Self}-{Determination} {Research}.
\newblock {\em Frontiers in Human Neuroscience}, 11.

\bibitem[Durkan et~al., 2019]{durkan_neural_2019}
Durkan, C., Bekasov, A., Murray, I., \& Papamakarios, G. (2019).
\newblock Neural spline flows.
\newblock {\em Advances in neural information processing systems}, 32.

\bibitem[Eskandari et~al., 2020]{eskandari_effect_2020}
Eskandari, Z., Seyedfatemi, N., Haghani, H., Almasi-Hashiani, A., \& Mohagheghi, P. (2020).
\newblock Effect of nesting on extensor motor behaviors in preterm infants: A randomized clinical trial.
\newblock {\em Iranian Journal of Neonatology}, 11(3), 64--70.

\bibitem[Eysenbach et~al., 2018]{eysenbach_diayn_2018}
Eysenbach, B., Gupta, A., Ibarz, J., \& Levine, S. (2018).
\newblock {Diversity} is {All} {You} {Need}: {Learning} {Skills} without a {Reward} {Function123}.
\newblock {\em arXiv:1802.06070}.

\bibitem[Friston, 2010]{friston_free-energy_2010}
Friston, K. (2010).
\newblock The free-energy principle: a unified brain theory?
\newblock {\em Nature Reviews Neuroscience}, 11(2), 127--138.

\bibitem[Friston et~al., 2016]{friston_active_2016}
Friston, K., FitzGerald, T., Rigoli, F., Schwartenbeck, P., O'Doherty, J., \& Pezzulo, G. (2016).
\newblock Active inference and learning.
\newblock {\em Neuroscience \& Biobehavioral Reviews}, 68, 862--879.

\bibitem[Friston et~al., 2017]{friston_graphical_2017}
Friston, K.~J., Parr, T., \& de~Vries, B. (2017).
\newblock The graphical brain: {Belief} propagation and active inference.
\newblock {\em Network Neuroscience}, 1(4), 381--414.

\bibitem[Gallagher, 1986]{gallagher_body_1986}
Gallagher, S. (1986).
\newblock Body {Image} and {Body} {Schema}: {A} {Conceptual} {Clarification}.
\newblock {\em The Journal of Mind and Behavior}, 7(4), 541--554.

\bibitem[Gopnik, 2009]{gopnik_philosophical_2009}
Gopnik, A. (2009).
\newblock {\em The philosophical baby: {What} children's minds tell us about truth, love \& the meaning of life}.
\newblock Random House.

\bibitem[Haarnoja et~al., 2018]{haarnoja_sac_2018}
Haarnoja, T., Zhou, A., Abbeel, P., \& Levine, S. (2018).
\newblock Soft actor-critic: Off-policy maximum entropy deep reinforcement learning with a stochastic actor.
\newblock In {\em International conference on machine learning}  (pp.\ 1861--1870).: PMLR.

\bibitem[Hafner et~al., 2020]{hafner_dreamer_2020}
Hafner, D., Lillicrap, T., Ba, J., \& Norouzi, M. (2020).
\newblock {Dream} to {Control}: {Learning} {Behaviors} by {Latent} {Imagination}.
\newblock {\em arXiv:1912.01603}.

\bibitem[Hafner et~al., 2019]{hafner_planet_2019}
Hafner, D., Lillicrap, T., Fischer, I., Villegas, R., Ha, D., Lee, H., \& Davidson, J. (2019).
\newblock Learning latent dynamics for planning from pixels.
\newblock In {\em International conference on machine learning}  (pp.\ 2555--2565).: PMLR.

\bibitem[Hoffmann, 2021]{hoffmann_body_2021}
Hoffmann, M. (2021).
\newblock Body models in humans, animals, and robots: {Mechanisms} and plasticity.
\newblock In {\em Body schema and body image: {New} directions.}  (pp.\ 152--180). New York, NY, US: Oxford University Press.

\bibitem[Hoffmann et~al., 2017]{hoffmann_development_2017}
Hoffmann, M., Chinn, L.~K., Somogyi, E., Heed, T., Fagard, J., Lockman, J.~J., \& O'Regan, J.~K. (2017).
\newblock Development of reaching to the body in early infancy: {From} experiments to robotic models.
\newblock In {\em 2017 {Joint} {IEEE} {International} {Conference} on {Development} and {Learning} and {Epigenetic} {Robotics} ({ICDL}-{EpiRob})}  (pp.\ 112--119).

\bibitem[Juechems \& Summerfield, 2019]{juechems_where_2019}
Juechems, K. \& Summerfield, C. (2019).
\newblock Where {Does} {Value} {Come} {From}?
\newblock {\em Trends in Cognitive Sciences}, 23(10), 836--850.

\bibitem[Kanazawa et~al., 2023]{kanazawa_open-ended_2023}
Kanazawa, H., Yamada, Y., Tanaka, K., Kawai, M., Niwa, F., Iwanaga, K., \& Kuniyoshi, Y. (2023).
\newblock Open-ended movements structure sensorimotor information in early human development.
\newblock {\em Proceedings of the National Academy of Sciences}, 120(1), e2209953120.

\bibitem[Keramati \& Gutkin, 2011]{keramati_reinforcement_2011}
Keramati, M. \& Gutkin, B. (2011).
\newblock A {Reinforcement} {Learning} {Theory} for {Homeostatic} {Regulation}.
\newblock In {\em Advances in {Neural} {Information} {Processing} {Systems}}, volume~24: Curran Associates, Inc.

\bibitem[Kim et~al., 2022]{kim_simulating_2022}
Kim, D., Kanazawa, H., \& Kuniyoshi, Y. (2022).
\newblock Simulating a {Human} {Fetus} in {Soft} {Uterus}.
\newblock In {\em 2022 {IEEE} {International} {Conference} on {Development} and {Learning} ({ICDL})}  (pp.\ 135--141).

\bibitem[Kim et~al., 2023]{kim_amam_2023}
Kim, D., Kanazawa, H., \& Kuniyoshi, Y. (2023).
\newblock {Emergence} of {Reaching} using {Predictive} {Learning} as {Sensorimotor} {Development} in {Complex} {Dynamics}.
\newblock In {\em The 11th {International} {Symposium} on {Adaptive} {Motion} of {Animals} and {Machines} ({AMAM2023})}  (pp.\ 144--145).: Adaptive Motion of Animals and Machines Organizing Committee.

\bibitem[Kingma, 2013]{kingma2013auto}
Kingma, D.~P. (2013).
\newblock Auto-encoding variational bayes.
\newblock {\em arXiv:1312.6114}.

\bibitem[Kuniyoshi, 2019]{kuniyoshi_fusing_2019}
Kuniyoshi, Y. (2019).
\newblock Fusing autonomy and sociality via embodied emergence and development of behaviour and cognition from fetal period.
\newblock {\em Philosophical Transactions of the Royal Society B: Biological Sciences}, 374(1771), 20180031.

\bibitem[Marcel et~al., 2022]{marcel_learning_2022}
Marcel, V., O’Regan, J.~K., \& Hoffmann, M. (2022).
\newblock Learning to reach to own body from spontaneous self-touch using a generative model.
\newblock In {\em 2022 {IEEE} {International} {Conference} on {Development} and {Learning} ({ICDL})}  (pp.\ 328--335).

\bibitem[Matsumoto et~al., 2022]{matsumoto_goal-directed_2022}
Matsumoto, T., Ohata, W., Benureau, F. C.~Y., \& Tani, J. (2022).
\newblock Goal-{Directed} {Planning} and {Goal} {Understanding} by {Extended} {Active} {Inference}: {Evaluation} through {Simulated} and {Physical} {Robot} {Experiments}.
\newblock {\em Entropy}, 24(4), 469.

\bibitem[Mazzaglia et~al., 2021]{mazzaglia_contrastive_2021}
Mazzaglia, P., Verbelen, T., \& Dhoedt, B. (2021).
\newblock Contrastive {Active} {Inference}.
\newblock {\em Advances in neural information processing systems}, 34, 13870--13882.

\bibitem[Mele \& Moser, 1994]{mele_intentional_1994}
Mele, A.~R. \& Moser, P.~K. (1994).
\newblock Intentional {Action}.
\newblock {\em Noûs}, 28(1), 39--68.

\bibitem[Millidge, 2020]{millidge_deep_2020}
Millidge, B. (2020).
\newblock Deep active inference as variational policy gradients.
\newblock {\em Journal of Mathematical Psychology}, 96, 102348.

\bibitem[Millidge et~al., 2020]{millidge_relationship_2020}
Millidge, B., Tschantz, A., Seth, A.~K., \& Buckley, C.~L. (2020).
\newblock On the {Relationship} {Between} {Active} {Inference} and {Control} as {Inference}.
\newblock In T. Verbelen, P. Lanillos, C.~L. Buckley, \& C. De~Boom (Eds.), {\em Active {Inference}}, Communications in {Computer} and {Information} {Science}  (pp.\ 3--11).  Cham: Springer International Publishing.

\bibitem[Oliver et~al., 2022]{oliver_empirical_2022}
Oliver, G., Lanillos, P., \& Cheng, G. (2022).
\newblock An {Empirical} {Study} of {Active} {Inference} on a {Humanoid} {Robot}.
\newblock {\em IEEE Transactions on Cognitive and Developmental Systems}, 14(2), 462--471.

\bibitem[Oudeyer \& Kaplan, 2009]{oudeyer_what_2009}
Oudeyer, P.-Y. \& Kaplan, F. (2009).
\newblock What is intrinsic motivation? {A} typology of computational approaches.
\newblock {\em Frontiers in neurorobotics}, 1, 6.

\bibitem[Parr et~al., 2022]{parr_active_2022}
Parr, T., Pezzulo, G., \& Friston, K.~J. (2022).
\newblock {\em Active {Inference}: {The} {Free} {Energy} {Principle} in {Mind}, {Brain}, and {Behavior}}.
\newblock The MIT Press.

\bibitem[Pathak et~al., 2017]{pathak_icm_2017}
Pathak, D., Agrawal, P., Efros, A.~A., \& Darrell, T. (2017).
\newblock Curiosity-driven exploration by self-supervised prediction.
\newblock In {\em International conference on machine learning}  (pp.\ 2778--2787).: PMLR.

\bibitem[Pearson et~al., 2011]{pearson_posterior_2011}
Pearson, J.~M., Heilbronner, S.~R., Barack, D.~L., Hayden, B.~Y., \& Platt, M.~L. (2011).
\newblock Posterior cingulate cortex: adapting behavior to a changing world.
\newblock {\em Trends in Cognitive Sciences}, 15(4), 143--151.

\bibitem[Priorelli et~al., 2023]{priorelli_deep_2023}
Priorelli, M., Pezzulo, G., \& Stoianov, I.~P. (2023).
\newblock Deep kinematic inference affords efficient and scalable control of bodily movements.
\newblock {\em Proceedings of the National Academy of Sciences}, 120(51), e2309058120.

\bibitem[Priorelli \& Stoianov, 2023]{priorelli_flexible_2023}
Priorelli, M. \& Stoianov, I.~P. (2023).
\newblock Flexible intentions: An active inference theory.
\newblock {\em Frontiers in Computational Neuroscience}, Volume 17 - 2023.

\bibitem[Rochat, 1998]{rochat_self-perception_1998}
Rochat, P. (1998).
\newblock Self-perception and action in infancy.
\newblock {\em Experimental Brain Research}, 123(1), 102--109.

\bibitem[Ryan \& Deci, 2000]{ryan_intrinsic_2000}
Ryan, R.~M. \& Deci, E.~L. (2000).
\newblock Intrinsic and {Extrinsic} {Motivations}: {Classic} {Definitions} and {New} {Directions}.
\newblock {\em Contemporary Educational Psychology}, 25(1), 54--67.

\bibitem[Sajid et~al., 2021]{sajid_exploration_2021}
Sajid, N., Tigas, P., Zakharov, A., Fountas, Z., \& Friston, K. (2021).
\newblock Exploration and preference satisfaction trade-off in reward-free learning.
\newblock {\em arXiv:2106.04316}.

\bibitem[Sancaktar et~al., 2020]{sancaktar_end_2020}
Sancaktar, C., van Gerven, M. A.~J., \& Lanillos, P. (2020).
\newblock End-to-end pixel-based deep active inference for body perception and action.
\newblock In {\em 2020 Joint IEEE 10th International Conference on Development and Learning and Epigenetic Robotics (ICDL-EpiRob)}  (pp.\ 1--8).

\bibitem[Schmidhuber, 2010]{schmidhuber_formal_2010}
Schmidhuber, J. (2010).
\newblock Formal {Theory} of {Creativity}, {Fun}, and {Intrinsic} {Motivation} (1990–2010).
\newblock {\em IEEE Transactions on Autonomous Mental Development}, 2(3), 230--247.

\bibitem[Schulman et~al., 2015]{schulman_gae_2015}
Schulman, J., Moritz, P., Levine, S., Jordan, M., \& Abbeel, P. (2015).
\newblock {High}-dimensional continuous control using generalized advantage estimation.
\newblock {\em arXiv:1506.02438}.

\bibitem[Shultz, 2013]{shultz_computational_2013}
Shultz, T.~R. (2013).
\newblock Computational {Models} in {Developmental} {Psychology}.
\newblock In P.~D. Zelazo (Ed.), {\em The {Oxford} {Handbook} of {Developmental} {Psychology}, {Vol}. 1: {Body} and {Mind}}. Oxford University Press.

\bibitem[Takemura et~al., 2018]{takemura_neural_2018}
Takemura, N., Inui, T., \& Fukui, T. (2018).
\newblock A neural network model for development of reaching and pointing based on the interaction of forward and inverse transformations.
\newblock {\em Developmental Science}, 21(3), e12565.

\bibitem[Thelen et~al., 1993]{thelen_transition_1993}
Thelen, E., Corbetta, D., Kamm, K., Spencer, J.~P., Schneider, K., \& Zernicke, R.~F. (1993).
\newblock The {Transition} to {Reaching}: {Mapping} {Intention} and {Intrinsic} {Dynamics}.
\newblock {\em Child Development}, 64(4), 1058--1098.

\bibitem[Vaswani et~al., 2017]{vaswani_transformer_2017}
Vaswani, A., Shazeer, N., Parmar, N., Uszkoreit, J., Jones, L., Gomez, A.~N., Kaiser, L., \& Polosukhin, I. (2017).
\newblock {Attention} {Is} {All} {You} {Need}.
\newblock {\em arXiv:1706.03762}.

\bibitem[Warneken \& Tomasello, 2006]{warneken_altruistic_2006}
Warneken, F. \& Tomasello, M. (2006).
\newblock Altruistic helping in human infants and young chimpanzees.
\newblock {\em Science}, 311(5765), 1301--1303.

\bibitem[White, 1959]{white_motivation_1959}
White, R.~W. (1959).
\newblock Motivation reconsidered: the concept of competence.
\newblock {\em Psychological review}, 66(5), 297.

\bibitem[Yoshida et~al., 2024]{yoshida_emergence_2024}
Yoshida, N., Daikoku, T., Nagai, Y., \& Kuniyoshi, Y. (2024).
\newblock Emergence of integrated behaviors through direct optimization for homeostasis.
\newblock {\em Neural Networks}, 177, 106379.

\bibitem[Zaadnoordijk \& Bayne, 2020]{zaadnoordijk_origins_2020}
Zaadnoordijk, L. \& Bayne, T. (2020).
\newblock The {Origins} of {Intentional} {Agency}.
\newblock {\em psyArXiv:wa8gb}.

\bibitem[Zaadnoordijk et~al., 2022]{zaadnoordijk_lessons_2022}
Zaadnoordijk, L., Besold, T.~R., \& Cusack, R. (2022).
\newblock Lessons from infant learning for unsupervised machine learning.
\newblock {\em Nature Machine Intelligence}, 4(6), 510--520.

\bibitem[Çatal et~al., 2020]{catal_learning_2020}
Çatal, O., Wauthier, S., De~Boom, C., Verbelen, T., \& Dhoedt, B. (2020).
\newblock Learning {Generative} {State} {Space} {Models} for {Active} {Inference}.
\newblock {\em Frontiers in Computational Neuroscience}, 14.

\end{thebibliography}

%%%%%%%%%%%%%%%%%%%%%%%%%%%%%%%%%%%%%%%%%%%%%%%%%%%%%%%%%%%%
\clearpage
\appendix

\section{Appendix}

\ifshowkorean
{
\subsection{이산 환경을 위한 모델의 상세}
}
\else

\subsection{Detailed Description of the Model for the Discrete Environment}

\fi

\subsubsection{Variational Free Energy Derivation}
\label{subsubsec:variational_free_energy_derivation}

\ifshowkorean
{\devscismall
이산 환경을 위한 모델에서, 관찰 $o_t$와 숨겨진 상태 $s_t$는 각각 가능한 경우의 수 $N_o$와 $N_s$개의 행을 갖는 one-hot 열벡터이다. 카테고리 분포의 파라미터는 행렬 및 벡터로 구현된다: $\mathbf{A}$는 $N_o \times N_s$ 행렬이며, 각각의 $\mathbf{B}_{a}$ 행렬은 $N_s \times N_s$ 정방 행렬이다. 예를 들어, 가능한 행동이 \{LEFT, STOP, RIGHT\}인 경우, $\mathbf{B}_{\text{LEFT}}, \mathbf{B}_{\text{STOP}}, \mathbf{B}_{\text{RIGHT}}$이 각각 존재한다. $\phi_t$는 $N_s$ 차원의 열벡터이다.
}
\else

In the discrete model, observation $o_t$ and hidden state $s_t$ are one-hot column vectors with $N_o$ and $N_s$ rows, respectively. Categorical distribution parameters are implemented as matrices: $\mathbf{A}$ is an $N_o \times N_s$ matrix, and each $\mathbf{B}_{a}$ is an $N_s \times N_s$ square matrix. For instance, if actions are \{LEFT, STOP, RIGHT\}, separate matrices $\mathbf{B}_{\text{LEFT}}, \mathbf{B}_{\text{STOP}}, \mathbf{B}_{\text{RIGHT}}$ exist.

\fi

\ifshowkorean
{\devscismall
변분 자유 에너지 $\mathcal{F}$는 다음과 같이 전개된다:
\begin{equation}
    \begin{aligned}
\mathcal{F}
&= \mathbb{E}_{Q(s_{t})}[\log Q(s_{t}) - \log P(o_{t},s_{t})] \\
&= \mathbb{E}_{Q(s_t)} [\log{Q(s_t)} - \log{P(s_t)}]
 - \mathbb{E}_{Q(s_t)}[\log P(o_t \mid s_t)] \\
&= \phi_t \cdot(\log{\phi_t} - \log{(\mathbf{B}_{a_{t-1}} \phi_{t-1})}) - \phi_t \cdot (\log{(\mathbf{A} \cdot o_t)} )
    \end{aligned}
\end{equation}
자유 에너지를 최소화하는 $\phi$는 $\mathcal{F}$의 $\phi$에 대한 미분으로부터 도출된다:
\begin{equation}
    \begin{aligned}
\frac{\partial \mathcal{F}} {\partial\phi_t}
&= \log{\phi_t} - \log{(\mathbf{B}_{a_{t-1}} \phi_{t-1})} + 1  - \log{(\mathbf{A} \cdot o_t)}   = 0 \\
\therefore \log{\phi_t} &\approx \log{(\mathbf{A} \cdot o_t)}  + \log{(\mathbf{B}_{a_{t-1}}\phi_{t-1})} \\
\phi_t &\approx \sigma(\log{(\mathbf{A} \cdot o_t)}  + \log{(\mathbf{B}_{a_{t-1}}\phi_{t-1})})
    \end{aligned}
\end{equation}
이것이 본문의 Eq.~\eqref{eq:fe_discrete_post}에 해당한다.
}
\else

The variational free energy $\mathcal{F}$ expands as:
\begin{equation}
    \begin{aligned}
\mathcal{F}
&= \mathbb{E}_{Q(s_{t})}[\log Q(s_{t}) - \log P(o_{t},s_{t})] \\
&= \mathbb{E}_{Q(s_t)} [\log{Q(s_t)} - \log{P(s_t)}]
 - \mathbb{E}_{Q(s_t)}[\log P(o_t \mid s_t)] \\
&= \phi_t \cdot(\log{\phi_t} - \log{(\mathbf{B}_{a_{t-1}} \phi_{t-1})}) - \phi_t \cdot (\log{(\mathbf{A} \cdot o_t)} )
    \end{aligned}
\end{equation}
The parameter $\phi$ that minimizes free energy is derived by differentiation:
\begin{equation}
    \begin{aligned}
\frac{\partial \mathcal{F}} {\partial\phi_t}
&= \log{\phi_t} - \log{(\mathbf{B}_{a_{t-1}} \phi_{t-1})} + 1  - \log{(\mathbf{A} \cdot o_t)}   = 0 \\
\therefore \log{\phi_t} &\approx \log{(\mathbf{A} \cdot o_t)}  + \log{(\mathbf{B}_{a_{t-1}}\phi_{t-1})} \\
\phi_t &\approx \sigma(\log{(\mathbf{A} \cdot o_t)}  + \log{(\mathbf{B}_{a_{t-1}}\phi_{t-1})})
    \end{aligned}
\end{equation}
This corresponds to Eq.~\eqref{eq:fe_discrete_post} in the main text.

\fi

\subsubsection{Expected Free Energy Derivation}
\label{subsubsec:expected_free_energy_derivation}

\ifshowkorean
{\devscismall
기대 자유 에너지의 계산 가능한 형태로의 전개는 다음과 같은 단계적 근사를 수행한다:
}
\else
The expected free energy can be derived through step-by-step approximations into a computationally tractable form:
\fi

\begin{equation}
    \begin{aligned}
    \mathcal{G} &= \mathbb{E}_{Q(s_{t+1},o_{t+1}\mid a_{t})}[\log Q(s_{t+1}\mid a_{t}) - \log \tilde{P}(o_{t+1},s_{t+1}\mid a_{t})] \\
    &= \mathbb{E}_{Q(s_{t+1},o_{t+1}\mid a_{t})}[\log Q(s_{t+1}\mid a_{t}) - \log P(s_{t+1}\mid o_{t+1}, a_{t}) - \log \tilde{P}(o_{t+1})] \\
    &\approx \mathbb{E}_{Q(s_{t+1},o_{t+1} \mid a_{t})}[\log Q(s_{t+1}\mid a_{t}) - \log Q(s_{t+1}\mid o_{t+1}, a_{t}) - \log \tilde{P}(o_{t+1})] \\
    &= \mathbb{E}_{Q(s_{t+1},o_{t+1}\mid a_{t})}[\log Q(o_{t+1}\mid a_{t}) - \log Q(o_{t+1}\mid s_{t+1}, a_{t}) - \log \tilde{P}(o_{t+1})]
    \end{aligned}
    \label{eq:efe_discompose}
\end{equation}

\ifshowkorean
{\devscismall
세 번째 줄의 근사는 true posterior $P(s_{t+1}\mid o_{t+1}, a_{t})$를 approximate posterior $Q(s_{t+1}\mid o_{t+1}, a_{t})$로 대체한 것이다. 네 번째 줄은 베이즈 정리 $Q(s,o) = Q(o|s)Q(s) = Q(s|o)Q(o)$를 적용하여 유도했다.
이를 실제 계산을 위해 재정리하면:
}
\else
The approximation in the third line replaces the true posterior $P(s_{t+1}\mid o_{t+1}, a_{t})$ with the approximate posterior $Q(s_{t+1}\mid o_{t+1}, a_{t})$.
The fourth line is derived by applying Bayes' rule: $Q(s,o) = Q(o|s)Q(s) = Q(s|o)Q(o)$.
Reformulating for computation:
\fi

\begin{equation}
    \begin{aligned}
    \mathcal{G} &\approx \mathbb{E}_{Q(s_{t+1}\mid a_{t})Q(o_{t+1}\mid s_{t+1},a_{t} )}[ - \log Q(o_{t+1}\mid s_{t+1}, a_{t})] \\
    &\phantom{=======}+ \mathbb{E}_{Q(o_{t+1}\mid a_{t})}[\log Q(o_{t+1}\mid a_{t}) - \log \tilde{P}(o_{t+1})] \\
    &= \mathbb{E}_{Q(s_{t+1}\mid a_{t})P(o_{t+1} \mid s_{t+1})}[ - \log P(o_{t+1}\mid s_{t+1})] \\
    &\phantom{=======}+ \mathbb{E}_{Q(o_{t+1}\mid a_{t})}[\log Q(o_{t+1}\mid a_{t}) - \log \tilde{P}(o_{t+1})] \\
    &= \mathbb{E}_{Q(s_{t+1}\mid a_{t})}[\mathcal{H}[P(o_{t+1} \mid s_{t+1})]] + D_{\mathrm{KL}}[Q(o_{t+1} \mid a_{t}) \| \tilde{P}(o_{t+1})] \\
    &= (\mathbf{B}_{a_{t}}\phi_{t}) \cdot \mathcal{H}[\mathbf{A}] + D_{\mathrm{KL}}[\mathbf{A} \mathbf{B}_{a_{t}}\phi_{t} \| \mathbf{C}]
    \end{aligned}
\end{equation}

\ifshowkorean
{\devscismall
이것이 본문의 Eq.~\eqref{eq:efe_discrete}에 해당한다. 우리의 설정에서는 $\mathbf{A}$와 $\mathbf{B}$가 이미 정확하게 알려져 있다고 가정하고 있으므로, 파라미터의 정보 이득을 위한 action entropy 항은 이산 환경 실험에서 생략되었다 (위의 식에 대한 더 자세한 설명은 \citep{parr_active_2022}를 참고하라).
}
\else
This corresponds to Eq.~\eqref{eq:efe_discrete} in the main text.
In our setup, since we assume that $\mathbf{A}$ and $\mathbf{B}$ are already known accurately, the action entropy term for information gain about parameters is omitted in the discrete environment experiment (see \citep{parr_active_2022} for more details on the above equations).
\fi

\ifshowkorean
{
\subsection{연속 환경을 위한 모델의 상세}
}
\else

\subsection{Detailed Description of the Model for the Continuous Environment}

\fi

\ifshowkorean
{
\subsubsection{연속 환경 모델의 Embedding 변환 구조}
}
\else

\subsubsection{Embedding Transformation Architecture for Continuous Environment Model}

\fi
\label{subsubsec:embedding_transformation_architecture_continuous}

\ifshowkorean
{\devscismall
연속 환경을 위한 모델에서 촉각과 고유 감각을 단일한 $o_t$ 표현으로 나타내기 위해, 각 모달리티를 동일한 차원을 갖는 embedding으로 변환한다.
촉각 행렬($80\times30$)은 Conv2D 모듈을 사용하여 embedding으로 변환하고, 고유 감각 벡터(3차원)는 MLP를 사용하여 embedding으로 변환한다.
두 embedding을 요소별 덧셈하여 최종적으로 통합된 관찰 embedding $o_t$를 생성한다.
Embedding으로부터 원래의 감각으로 복원하기 위해, 촉각은 ConvTranspose2D를 사용하고, 고유 감각은 MLP를 사용한다.
}
\else

To represent both tactile and proprioceptive modalities as a unified observation $o_t$ in the model for the continuous environment, we convert each modality into an embedding of the same dimensionality.
The tactile matrix ($80\times30$) is transformed into an embedding using a Conv2D module, and the proprioceptive vector (3-dimensional) is transformed into an embedding using a MLP.
We combine the two embeddings via element-wise addition to yield the final integrated observation embedding $o_t$.
To decode from the embedding back to the original sensations, we use a ConvTranspose2D for tactile and a MLP for proprioception.

\fi

\subsubsection{RSSM Computation}
\label{subsubsec:rssm_computation}

\ifshowkorean
{\devscismall
결정론적 상태 $h_t$는 다음과 같이 계산된다:
\begin{enumerate}
\item 이전 시간 단계의 확률적 상태 $s_{t-1}$과 행동 $a_{t-1}$을 선형 레이어를 통해 단일 벡터로 변환
\item 변환된 벡터를 이전 결정론적 상태 $h_{t-1}$와 함께 GRU cell~\citep{chung_empirical_2014}에 입력
\item GRU cell의 출력이 새로운 결정론적 상태 $h_t$가 됨
\end{enumerate}

확률적 사전 분포 $p_{\phi}(\hat{s}_t \mid h_t)$는 $h_t$를 MLP에 통과시켜 다변량 가우시안 분포의 평균과 분산 파라미터를 출력하며, 이 분포에서 확률적 상태를 샘플링한다.

확률적 사후 분포 $q_{\phi}(s_t \mid h_t, o_t)$는 결정론적 상태 $h_t$와 embedding된 관찰 $o_t$를 연결(concatenate)한 벡터를 입력으로 받는다. 연결된 벡터를 MLP에 통과시켜 다변량 가우시안 분포의 평균과 분산 파라미터를 출력하며, 이 분포에서 확률적 상태를 샘플링한다.

Likelihood $p_{\phi}(o_t \mid h_t, s_t)$는 결정론적 상태 $h_t$와 확률적 상태 $s_t$를 연결한 벡터를 선형 레이어를 통해 embedding된 $o_t$으로 변환한다. 이후 embedding된 $o_t$를 각 modal별 관찰로 복원(decode)한다.

이러한 RSSM 구조를 구성하는 모든 파라미터 $\phi$는 아래와 같은 변분 자유 에너지를 최소화하도록 경사 하강법에 의해 학습된다:
}
\else

The deterministic state $h_t$ is computed as follows:
\begin{enumerate}
\item Encode previous stochastic state $s_{t-1}$ and action $a_{t-1}$ into a single vector via a linear layer
\item Feed the encoded vector together with previous deterministic state $h_{t-1}$ into a GRU cell~\citep{chung_empirical_2014}
\item The GRU cell output becomes the new deterministic state $h_t$
\end{enumerate}

The stochastic prior $p_{\phi}(\hat{s}_t \mid h_t)$ passes $h_t$ through a MLP to output mean and variance parameters of a multivariate Gaussian, from which the stochastic state is sampled.

The stochastic posterior $q_{\phi}(s_t \mid h_t, o_t)$ receives a concatenated vector of deterministic state $h_t$ and embedded observation $o_t$. The concatenated vector is passed through a MLP to output mean and variance parameters of a multivariate Gaussian, from which the stochastic state is sampled.

The likelihood $p_{\phi}(o_t \mid h_t, s_t)$ transforms the concatenated $h_t$ and $s_t$ into an embedded $o_t$ through a linear layer, which is then decoded into individual modality-specific observations.

All parameters $\phi$ constituting this RSSM structure are optimized via gradient descent to minimize the following variational free energy:

\fi

\begin{equation}
    \begin{aligned}
\mathcal{F}
&= \mathbb{E}_{q(s_{t})}[\log q(s_{t}) - \log p(o_{t},s_{t})] \\
&= \mathbb{E}_{q(s_t)} [\log{q(s_t)} - \log{p(s_t)}]
 - \mathbb{E}_{q(s_t)}[\log p(o_t \mid s_t)] \\
&= D_{\mathrm{KL}}[q(s_t)\|p(s_t)]
 - \mathbb{E}_{q(s_t)}[\log p(o_t \mid s_t)] \\
\therefore \argmin_\phi \mathcal{F}&= \argmin_\phi \big[ D_{\mathrm{KL}}[q_\phi(s_t \mid h_t, o_t)\|p_\phi(s_t \mid h_t)]
 - \mathbb{E}_{q_\phi(s_t \mid h_t, o_t)}[\log p_\phi(o_t \mid h_t, s_t)] \big]
    \end{aligned}
    \label{eq:continuous_free_energy}
\end{equation}

\subsubsection{Policy and Value Network Training}
\label{subsubsec:policy_and_value_network_training}

\ifshowkorean
{\devscismall
Policy network $q_\theta(a_t \mid s_t)$와 utility (value) network $g_\psi(s_t)$는 모두 확률적 상태 $s_t$를 입력으로 받아 MLP를 사용하여 다변량 가우시안 분포의 파라미터를 출력하고, 이 분포에서 결과를 샘플링한다.

학습 목적 함수는 다음과 같다:
\begin{equation}
    \begin{aligned}
\argmin_\theta \mathcal{L}_\text{policy} &= \argmin_\theta \sum\nolimits_t G^\lambda_t \\
\argmin_\psi \mathcal{L}_\text{utility} &= \argmin_\psi \sum\nolimits_t (g_\psi(s_t) - G^\lambda_t) \\
G^\lambda_t &= \mathcal{G}(s_t) + \gamma_t 
    \begin{cases}
      (1 - \lambda) g_\psi(s_{t+1}) + \lambda G^\lambda_{t+1}, & \text{if}\ t<H \\
      g_\psi(s_H), & \text{if}\ t=H
    \end{cases} \\
    \mathcal{G} \approx
    -\mathbb{E}_{q(o^I_t)}[\log \tilde{p}(o^I_t)]
    &-\mathbb{E}_{q(o_t)}[D_{\mathrm{KL}}[q(s_t \mid o_t) \| q(s_t \mid s_{t-1}, a_{t-1})]]
    -\mathbb{E}_{q(s_t)}[\mathcal{H}(q(a_t \mid s_t))]
    \end{aligned}
\end{equation}

여기서 $G^\lambda_t$는 GAE($\lambda$) 추정을 사용한 기대 효용의 근사값이다. $\gamma_t$는 할인 계수이며, $H$는 시뮬레이션 지평선이다.
}
\else

Both the policy network $q_\theta(a_t \mid s_t)$ and utility (value) network $g_\psi(s_t)$ take the stochastic state $s_t$ as input, use a MLP to output multivariate Gaussian parameters, and sample from this distribution.

The training objectives are:
\begin{equation}
    \begin{aligned}
\argmin_\theta \mathcal{L}_\text{policy} &= \argmin_\theta \sum\nolimits_t G^\lambda_t \\
\argmin_\psi \mathcal{L}_\text{utility} &= \argmin_\psi \sum\nolimits_t (g_\psi(s_t) - G^\lambda_t) \\
G^\lambda_t &= \mathcal{G}(s_t) + \gamma_t 
    \begin{cases}
      (1 - \lambda) g_\psi(s_{t+1}) + \lambda G^\lambda_{t+1}, & \text{if}\ t<H \\
      g_\psi(s_H), & \text{if}\ t=H
    \end{cases}
    \end{aligned}
\end{equation}

Here, $G^\lambda_t$ is the GAE($\lambda$)-estimated expected utility approximation. $\gamma_t$ is the discount factor and $H$ is the simulation horizon.
\fi

\ifshowkorean
{
\subsection{연속 환경 실험의 하이퍼파라미터}
}
\else
\subsection{Hyperparameters for Continuous Environment Experiments}
\fi

\ifshowkorean
{\devscismall
표~\ref{table:hyperparams_comprehensive}는 연속 환경 실험에 사용된 모든 하이퍼파라미터를 정리한 것이다.
}
\else
Table~\ref{table:hyperparams_comprehensive} presents all hyperparameters used in the continuous environment experiments.
\fi

\begin{table}[ht]
    \centering
    \caption{
    \ifshowkorean
    {\devsciscriptsize 연속 환경 실험의 하이퍼파라미터}
    \else
    Hyperparameters for continuous environment experiments
    \fi
    }
    \label{table:hyperparams_comprehensive}
    \begin{tabular}{lll}
    \toprule
    Parameter & Value & Description \\
    \midrule
    \multicolumn{3}{l}{
    \ifshowkorean
    {\devsciscriptsize \textbf{환경 및 데이터 수집}}
    \else
    
    \textbf{Environment \& Data Collection}
    
    \fi
    } \\
    \ifshowkorean
    {\devsciscriptsize 에피소드 길이 ($L$)}
    \else
    
    Episode length ($L$)
    
    \fi
    & $1000$ & 
    \ifshowkorean
    {\devsciscriptsize 각 에피소드의 시간 스텝 수}
    \else
    
    Time steps per episode
    
    \fi
    \\
    \ifshowkorean
    {\devsciscriptsize 초기 랜덤 에피소드}
    \else
    
    Initial random episodes
    
    \fi
    & $100$ & 
    \ifshowkorean
    {\devsciscriptsize 학습 전 수집하는 랜덤 에피소드 수}
    \else
    
    Random episodes before training
    
    \fi
    \\
    \ifshowkorean
    {\devsciscriptsize 탐색 노이즈}
    \else
    
    Exploration noise
    
    \fi
    & $0.3$ & 
    \ifshowkorean
    {\devsciscriptsize 훈련시 정책에 추가되는 탐색 노이즈}
    \else
    
    Noise added to policy during training 
    
    \fi
    \\
    \midrule
    \multicolumn{3}{l}{
    \ifshowkorean
    {\devsciscriptsize \textbf{학습 스케줄}}
    \else
    
    \textbf{Training Schedule}
    
    \fi
    } \\
    \ifshowkorean
    {\devsciscriptsize Replay buffer 크기}
    \else
    
    Replay buffer size
    
    \fi
    & $450$ & 
    \ifshowkorean
    {\devsciscriptsize 메모리에 저장되는 최대 에피소드 수}
    \else
    
    Max episodes stored in memory
    
    \fi
    \\
    \ifshowkorean
    {\devsciscriptsize 에피소드당 학습 epoch}
    \else
    
    Training epochs per episode
    
    \fi
    & $100$ & 
    \ifshowkorean
    {\devsciscriptsize 각 에피소드 후 학습 반복 횟수}
    \else
    
    Training iterations after each episode
    
    \fi
    \\
    \ifshowkorean
    {\devsciscriptsize 배치 크기 ($B$)}
    \else
    
    Batch size ($B$)
    
    \fi
    & $50$ & 
    \ifshowkorean
    {\devsciscriptsize 각 epoch에서 샘플링하는 경로 수}
    \else
    
    Number of trajectories sampled per epoch
    
    \fi
    \\
    \ifshowkorean
    {\devsciscriptsize 경로 길이 ($L$)}
    \else
    
    Trajectory length ($L$)
    
    \fi
    & $50$ & 
    \ifshowkorean
    {\devsciscriptsize 각 샘플 경로의 시간 스텝 수}
    \else
    
    Time steps per sampled trajectory
    
    \fi
    \\
    \ifshowkorean
    {\devsciscriptsize 계획 지평선 ($H$)}
    \else
    
    Planning horizon ($H$)
    
    \fi
    & $15$ & 
    \ifshowkorean
    {\devsciscriptsize 정책 학습을 위한 미래 예측 범위}
    \else
    
    Future prediction horizon for policy learning
    
    \fi
    \\
    
    \ifshowkorean
    {\devsciscriptsize 경사 클리핑}
    \else
    
    Gradient clipping
    
    \fi
    & $100.0$ & 
    \ifshowkorean
    {\devsciscriptsize 경사의 최대 노름}
    \else
    
    Maximum gradient norm
    
    \fi
    \\
    
    \ifshowkorean
    {\devsciscriptsize 네트워크 최적화기}
    \else
    Network optimizer
    \fi
    & AdamW & 
    \ifshowkorean
    {\devsciscriptsize 학습을 위한 최적화기}
    \else
    Optimizer for training
    \fi
    \\
    
    \midrule
    \multicolumn{3}{l}{
    \ifshowkorean
    {\devsciscriptsize \textbf{Observation Embedding}}
    \else
    
    \textbf{Observation Embedding}
    
    \fi
    } \\
    \ifshowkorean
    {\devsciscriptsize CNN Encoder 채널}
    \else
    
    CNN encoder channels
    
    \fi
    & $4, 8, 16, 16$ & 
    \ifshowkorean
    {\devsciscriptsize Encoder 각 레이어의 입력/출력 채널 수}
    \else
    
    Output channels for each encoder layer
    
    \fi
    \\
    \ifshowkorean
    {\devsciscriptsize CNN Decoder 채널}
    \else
    
    CNN decoder channels
    
    \fi
    & $16, 8, 4, 3$ & 
    \ifshowkorean
    {\devsciscriptsize Decoder 각 레이어의 입력/출력 채널 수}
    \else
    
    Output channels for each decoder layer
    
    \fi
    \\
    \ifshowkorean
    {\devsciscriptsize CNN Encoder kernel 크기}
    \else
    
    CNN encoder kernel sizes
    
    \fi
    & $4, 4, 4, 2$ & 
    \ifshowkorean
    {\devsciscriptsize Encoder 각 레이어의 kernel 크기}
    \else
    
    Kernel size for each encoder layer
    
    \fi
    \\
    \ifshowkorean
    {\devsciscriptsize CNN Decoder kernel 크기}
    \else
    
    CNN decoder kernel sizes
    
    \fi
    & $2, 4, 4, 4$ & 
    \ifshowkorean
    {\devsciscriptsize Decoder 각 레이어의 kernel 크기}
    \else
    
    Kernel size for each decoder layer
    
    \fi
    \\
    \ifshowkorean
    {\devsciscriptsize CNN stride}
    \else
    
    CNN stride
    
    \fi
    & $2$ & 
    \ifshowkorean
    {\devsciscriptsize 모든 레이어의 stride 값}
    \else
    
    Stride value for all layers
    
    \fi
    \\
    \ifshowkorean
    {\devsciscriptsize MLP 은닉층 크기}
    \else
    
    MLP hidden layer size
    
    \fi
    & $32$ & 
    \ifshowkorean
    {\devsciscriptsize MLP 은닉층 유닛 수}
    \else
    
    Units in MLP hidden layer
    
    \fi
    \\
    \ifshowkorean
    {\devsciscriptsize MLP 은닉층 수}
    \else
    
    MLP hidden layers
    
    \fi
    & $2$ & 
    \ifshowkorean
    {\devsciscriptsize MLP 은닉층 수}
    \else
    
    Number of hidden layers in MLP
    
    \fi
    \\
    \ifshowkorean
    {\devsciscriptsize 통합 Embedding 크기}
    \else
    
    Integrated embedding size
    
    \fi
    & $64$ & 
    \ifshowkorean
    {\devsciscriptsize 최종 embedding 차원}
    \else
    
    Final embedding dimension
    
    \fi
    \\
    \ifshowkorean
    {\devsciscriptsize 활성화 함수}
    \else
    
    Activation layer
    
    \fi
    & ELU & 
    \ifshowkorean
    {\devsciscriptsize 은닉층의 활성화 함수}
    \else
    
    Activation layer
    
    \fi
    \\
    \midrule
    \multicolumn{3}{l}{
    \ifshowkorean
    {\devsciscriptsize \textbf{World Model (RSSM)}}
    \else
    
    \textbf{World Model (RSSM)}
    
    \fi
    } \\
    \ifshowkorean
    {\devsciscriptsize 결정론적 상태 차원 ($\dim(h_t)$)}
    \else
    
    Deterministic state size ($\dim(h_t)$)
    
    \fi
    & $200$ & 
    \ifshowkorean
    {\devsciscriptsize GRU 은닉 상태의 차원}
    \else
    
    Dimension of GRU hidden state
    
    \fi
    \\
    \ifshowkorean
    {\devsciscriptsize 확률론적 상태 차원 ($\dim(s_t)$)}
    \else
    
    Stochastic state size ($\dim(s_t)$)
    
    \fi
    & $30$ & 
    \ifshowkorean
    {\devsciscriptsize 확률론적 잠재 변수의 차원}
    \else
    
    Dimension of stochastic latent variable
    
    \fi
    \\
    \ifshowkorean
    {\devsciscriptsize 은닉층 크기}
    \else
    
    Hidden layer size
    
    \fi
    & $200$ & 
    \ifshowkorean
    {\devsciscriptsize 은닉층의 유닛 수}
    \else
    
    Units in hidden layers
    
    \fi
    \\
    \ifshowkorean
    {\devsciscriptsize 은닉층 수}
    \else
    
    Number of hidden layers
    
    \fi
    & $1$ & 
    \ifshowkorean
    {\devsciscriptsize 은닉층 수}
    \else
    
    Number of hidden layers
    
    \fi
    \\
    \ifshowkorean
    {\devsciscriptsize 활성화 함수}
    \else
    
    Activation layer
    
    \fi
    & ELU & 
    \ifshowkorean
    {\devsciscriptsize 은닉층의 활성화 함수}
    \else
    
    Activation layer
    
    \fi
    \\
    \ifshowkorean
    {\devsciscriptsize 학습률 ($\alpha_\phi$)}
    \else
    
    Learning rate ($\alpha_\phi$)
    
    \fi
    & $10^{-3}$ & 
    \ifshowkorean
    {\devsciscriptsize World model 파라미터의 학습률}
    \else
    
    Learning rate for world model parameters
    
    \fi
    \\
    \ifshowkorean
    {\devsciscriptsize Free nats}
    \else
    
    Free nats
    
    \fi
    & $3.0$ & 
    \ifshowkorean
    {\devsciscriptsize KL divergence의 최소 임계값}
    \else
    
    Minimum threshold for KL divergence
    
    \fi
    \\
    \ifshowkorean
    {\devsciscriptsize KL 스케일 ($\beta$)}
    \else
    
    KL scale ($\beta$)
    
    \fi
    & $2.0$ & 
    \ifshowkorean
    {\devsciscriptsize KL divergence 가중치 ($\beta$-VAE)}
    \else
    
    KL divergence weight ($\beta$-VAE)
    
    \fi
    \\
    \midrule
    \multicolumn{3}{l}{
    \ifshowkorean
    {\devsciscriptsize \textbf{Policy 및 Value Network}}
    \else
    
    \textbf{Policy \& Value Networks}
    
    \fi
    } \\
    \ifshowkorean
    {\devsciscriptsize 은닉층 크기}
    \else
    
    Hidden layer size
    
    \fi
    & $200$ & 
    \ifshowkorean
    {\devsciscriptsize 은닉층 유닛 수}
    \else
    
    Units in hidden layers
    
    \fi
    \\
    \ifshowkorean
    {\devsciscriptsize 은닉층 수}
    \else
    
    Number of hidden layers
    
    \fi
    & $3$ & 
    \ifshowkorean
    {\devsciscriptsize 은닉층 수}
    \else
    
    Number of hidden layers
    
    \fi
    \\
    \ifshowkorean
    {\devsciscriptsize 활성화 함수}
    \else
    
    Activation layer
    
    \fi
    & ELU & 
    \ifshowkorean
    {\devsciscriptsize 은닉층의 활성화 함수}
    \else
    
    Activation layer
    
    \fi
    \\
    \ifshowkorean
    {\devsciscriptsize 학습률 ($\alpha_\theta, \alpha_\psi$)}
    \else
    
    Learning rate ($\alpha_\theta, \alpha_\psi$)
    
    \fi
    & $10^{-3}$ & 
    \ifshowkorean
    {\devsciscriptsize 파라미터의 학습률}
    \else
    
    Learning rate for parameters
    
    \fi
    \\
    \ifshowkorean
    {\devsciscriptsize Entropy 온도}
    \else
    
    Entropy temperature
    
    \fi
    & $10^{-4}$ & 
    \ifshowkorean
    {\devsciscriptsize Entropy 정규화 계수}
    \else
    
    Entropy regularization coefficient
    
    \fi
    \\
    \ifshowkorean
    {\devsciscriptsize 할인 계수 ($\gamma$)}
    \else
    
    Discount factor ($\gamma$)
    
    \fi
    & $0.99$ & 
    \ifshowkorean
    {\devsciscriptsize 미래 보상의 할인율}
    \else
    
    Discount rate for future rewards
    
    \fi
    \\
    \ifshowkorean
    {\devsciscriptsize GAE lambda ($\lambda$)}
    \else
    
    GAE lambda ($\lambda$)
    
    \fi
    & $0.95$ & 
    \ifshowkorean
    {\devsciscriptsize GAE 추정의 평활화 파라미터}
    \else
    
    Smoothing parameter for GAE estimation
    
    \fi
    \\
    \midrule
    \multicolumn{3}{l}{
    \ifshowkorean
    {\devsciscriptsize \textbf{Self-Prior (Neural Spline Flow)}}
    \else
    
    \textbf{Self-Prior (Neural Spline Flow)}
    
    \fi
    } \\
    \ifshowkorean
    {\devsciscriptsize Spline 변환 수}
    \else
    
    Number of spline transforms
    
    \fi
    & $3$ & 
    \ifshowkorean
    {\devsciscriptsize Flow에서 결합되는 변환 층 수}
    \else
    
    Number of transform layers in the flow
    
    \fi
    \\
    \ifshowkorean
    {\devsciscriptsize 은닉층 크기}
    \else
    
    Hidden layer size
    
    \fi
    & $64$ & 
    \ifshowkorean
    {\devsciscriptsize Spline network의 은닉층 유닛 수}
    \else
    
    Units in spline network hidden layers
    
    \fi
    \\
    \ifshowkorean
    {\devsciscriptsize 활성화 함수}
    \else
    
    Activation layer
    
    \fi
    & ReLU & 
    \ifshowkorean
    {\devsciscriptsize 은닉층의 활성화 함수}
    \else
    
    Activation layer
    
    \fi
    \\
    \ifshowkorean
    {\devsciscriptsize 학습률 ($\alpha_\xi$)}
    \else
    
    Learning rate ($\alpha_\xi$)
    
    \fi
    & $10^{-3}$ & 
    \ifshowkorean
    {\devsciscriptsize Self-prior 파라미터의 학습률}
    \else
    
    Learning rate for self-prior parameters
    
    \fi
    \\
    \bottomrule
    \end{tabular}
    \end{table}

\ifdevscianonymise
    \clearpage
    \section*{List of Figure Legends}
    -- Figure 1: Emergence of reaching behavior via the self-prior and active inference. (a) When a sticker is placed on the left arm of the simulated agent, it detects a mismatch with its prior experience of not having a sticker, and reaches toward the sticker with its right hand to minimize the discrepancy. (b) Development of the self-prior through experience: as sensory experiences are collected, the probability distribution over sensory patterns gradually develops. (c) The active inference process in which the agent plans future actions to minimize expected free energy by aligning sensory inputs with the learned self-prior. As a result, the agent performs a reaching action toward the sticker. A full-body infant illustration is used for clarity, the actual experiment was conducted in a pseudo-3D environment.
    
    -- Figure 2: Graphical model of active inference using deep neural networks that minimize variational free energy $\mathcal{F}$ and expected free energy $\mathcal{G}$. The self-prior $\tilde{p}(o^I_t)$ is trained to maximize the log-likelihood of observations $o^I_t$. In the expected free energy calculation (highlighted in blue), the learned self-prior serves as the behavioral setpoint alongside the fixed preferred prior. Although the preferred prior and self-prior can theoretically be applied simultaneously, we use only the self-prior in this study for clarity of exposition; thus, the preferred prior is shown faded in the figure.
    
    -- Figure 3: Overview of the discrete environment. The right hand can move left or right either above the left arm or outside of it, and tactile input occurs either where the right hand is located or where the sticker is attached.
    
    -- Figure 4: Change in self-prior over time. Before the sticker is attached, the probability increases for situations where no sticker is present on the arm ($t<10,000$). After the sticker is attached, the agent gradually adapts to the new situation where the sticker is present ($t\ge10,000$).
    
    -- Figure 5: Comparison of the agent's behavior before and after acquiring the self-prior. The top panel illustrates environmental changes over time: the red line indicates the hand position, the yellow dashed line indicates the sticker position. White areas denote where tactile feedback occurred, black areas indicate no tactile feedback, and gray areas represent regions outside the arm where tactile feedback never occurs. The bottom panel shows expected free energy over time. Each green dot represents the free energy of a candidate policy, with lower free energy policies being more likely to be selected. The red line connects the actually selected policies. (a) Before acquiring the self-prior, the agent does not respond even when a sticker is attached to the arm. (b) After acquiring the self-prior, goal-directed behavior emerges: the agent moves its hand to the sticker’s location when it appears on the arm.
    
    -- Figure 5a: Before acquiring the self-prior ($t$ = 0)
    
    -- Figure 5b: After acquiring the self-prior ($t$ = 10,000)
    
    -- Figure 6: Agent's behavior when a sticker has persistently been attached to position~3 ($t = 30{,}000$). When the sticker is placed on other positions, the agent still reaches toward them; however, it no longer shows interest when the sticker is placed at position~3.
    
    -- Figure 7: Overview of the continuous environment. As in the discrete environment, the right hand can move over and around the left arm, and tactile sensations are generated where the hand or the sticker is located.
    
    -- Figure 8: Agent behavior when a sticker is placed in the continuous environment. (a) When a sticker (blue circle) is attached, the hand (red circle) moves toward it, illustrating goal-directed reaching behavior. The figure visualizes the agent’s tactile matrix, where grayscale tactile data is overlaid with colored markers for the hand and sticker for clarity. (b) Reaching for the sticker reduces expected free energy, and removing the sticker leads to its minimization. Shaded areas represent the standard deviation across 64 experiments using 8 model training seeds (0\char`~7) tested on 8 environment seeds (0\char`~7). The figure uses seed 4, selected for clearly demonstrating the reaching behavior.
    
    -- Figure 8a: Time series of reaching behavior
    
    -- Figure 8b: Change in expected free energy
\else
    
\fi

\end{document}